\documentclass{article} 
\usepackage{iclr2027_conference,times}

\usepackage{amsmath,amsfonts,bm}

\def\eqref#1{equation~\ref{#1}}

\def\1{\bm{1}}

\DeclareMathAlphabet{\mathsfit}{\encodingdefault}{\sfdefault}{m}{sl}
\SetMathAlphabet{\mathsfit}{bold}{\encodingdefault}{\sfdefault}{bx}{n}

\usepackage{booktabs}
\usepackage{tabularx}
\usepackage[most]{tcolorbox}
\usepackage{float}
\usepackage{placeins}

\definecolor{tabgrey}{RGB}{242,242,242}

\definecolor{thmbg}{RGB}{248,245,253}   

\newtcolorbox{tablebox}[1][]{%
  enhanced,
  breakable,
  colback=thmbg,
  colframe=thmbg,
  boxrule=0pt,
  arc=2pt,
  left=8pt,
  right=8pt,
  top=7pt,
  bottom=7pt,
  #1
}
\usepackage{url}            
\usepackage{booktabs}       
\usepackage{amsfonts}       
\usepackage{nicefrac}       
\usepackage{microtype}      
\usepackage{xcolor}         
\usepackage{graphicx}
\usepackage[most]{tcolorbox}
\usepackage{enumitem} 
\usepackage{amsthm,amssymb}
\usepackage{hyperref}
\usepackage{cleveref}   
\usepackage{titletoc}

\newtheoremstyle{boxedplain}{0pt}{0pt}{\itshape}{}{\bfseries}{.}{5pt plus 1pt minus 1pt}{}
\theoremstyle{boxedplain}
\newtheorem{definition}{Definition}[section]
\newtheorem{proposition}[definition]{Proposition}

\newtheorem{lemma}[definition]{Lemma}

\newtheoremstyle{boxedplain}
  {0pt}
  {0pt}
  {\itshape}
  {}
  {\bfseries}
  {.}
  {5pt plus 1pt minus 1pt}
  {}

\theoremstyle{boxedplain}

\tcbset{
  thmbox/.style={
    breakable,
    colback=thmbg,
    colframe=thmbg,
    boxrule=0pt,
    arc=2pt,
    boxsep=0pt,
    left=4pt,
    right=4pt,
    top=1.5pt,
    bottom=2pt,
    before skip=\topsep,
    after skip=\topsep
  }
}

\tcolorboxenvironment{definition}{thmbox}
\tcolorboxenvironment{proposition}{thmbox}
\tcolorboxenvironment{corollary}{thmbox}
\tcolorboxenvironment{lemma}{thmbox}

\newtcolorbox{procedurebox}[1]{
  colback=white,          
  colframe=gray!30,       
  coltitle=black,         
  fonttitle=\bfseries,
  title={#1},
  boxrule=0.8pt,
  arc=0pt,                
  left=15pt,              
  right=10pt,
  top=5pt,
  bottom=5pt,
  enhanced,
  borderline west={2pt}{0pt}{gray!80}
}

\newlist{steps}{enumerate}{1}
\setlist[steps]{
  label=\textbf{\arabic*.}, 
  leftmargin=*,
  labelsep=10pt,
  nosep 
}
\newtcolorbox{takeawaybox}[1]{
  enhanced,
  breakable,
  colback=white,
  colframe=black,
  coltitle=black,
  boxrule=0.45pt,
  sharp corners,
  left=7pt,
  right=7pt,
  top=2pt,
  bottom=4pt,
  before skip=6pt,
  after skip=6pt,
  fonttitle=\bfseries,
  title={#1},
  attach boxed title to top left={
    xshift=8pt,
    yshift*=-\tcboxedtitleheight/2
  },
  boxed title style={
    colback=white,
    colframe=white,
    boxrule=0pt,
    sharp corners,
    left=1.5pt,
    right=1.5pt,
    top=0pt,
    bottom=0pt
  }
}
\definecolor{freq}{HTML}{246A73} 
\definecolor{bayes}{HTML}{A31621} 
\definecolor{resample}{HTML}{5E503F}
\definecolor{post-risk}{HTML}{560BAD}
\definecolor{blrcolor}{HTML}{55A868}
\definecolor{treecolor}{HTML}{DD8452}
\definecolor{dklcolor}{HTML}{C44E52}
\definecolor{nncolor}{HTML}{8172B2}
\definecolor{noauc}{HTML}{D9D9D9}

\usepackage{hyperref}
\usepackage{url}

\title{A Ground-Truth Framework for Uncertainty Disentanglement with Posterior Risk}

\author{\textbf{Frieder Wizgall}$^{*,\,1}$
\quad \textbf{Georg Tirpitz}$^{*,\,1}$\
\quad \textbf{Moritz Seiler}$^{2,\,5}$\\[3pt]
\quad \textbf{Kerstin Ritter}$^{2,\,3,\,4,\,5}$
\quad \textbf{B\'{a}lint Mucs\'{a}nyi}$^{1}$\\[3pt]
$^{1}$University of T\"ubingen, T\"ubingen, Germany,\\
$^{2}$Hertie Institute for AI in Brain Health, University of T\"ubingen, T\"ubingen, Germany,\\
$^{3}$T\"ubingen AI Center,University of T\"ubingen, T\"ubingen, Germany,\\
$^{4}$Charit\'e–Universit\"atsmedizin Berlin, Department of Psychiatry and Psychotherapy, Berlin, Germany,\\
$^{5}$Bernstein Center for Computational Neuroscience Berlin, Berlin, Germany
}

\begin{document}

\maketitle
\renewcommand{\thefootnote}{}
\footnotetext{* Equal contribution (shared first authorship).}
\renewcommand{\thefootnote}{\arabic{footnote}}

\begin{abstract}
Reliable uncertainty estimates are critical in safety-sensitive applications. For such estimates to be useful in practice, it is crucial to understand the sources underlying a model’s uncertainty, motivating the disentanglement of total uncertainty into epistemic and aleatoric uncertainty. Existing notions of uncertainty differ in the sources they capture and, consequently, in their definitions of aleatoric and epistemic uncertainty, with no universally accepted definition. We define uncertainty through sample-conditional pointwise posterior risk, which is the expected loss of a predictor under the distribution of plausible ground-truth functions given the observed sample. This definition unifies probabilistic and risk-based concepts of uncertainty. To assess state-of-the-art uncertainty disentanglement methods, we develop a framework that directly compares their estimates against ground-truth uncertainty defined primarily by posterior risk, alongside commonly used alternative uncertainty definitions. We find that Spectral-normalized Neural Gaussian Processes and Variational Latent Gaussian Processes most closely recover the ground-truth uncertainty, while most methods track posterior variance more closely than posterior risk, missing the predictors' bias. Beyond method rankings, we investigate how strongly estimated aleatoric and epistemic uncertainty are entangled and how sensitive uncertainty quality is to modeling choices, yielding practical guidance for uncertainty disentanglement. To support further method development and validation, we release $13$ semi-synthetic UCI/OpenML datasets with known posteriors, enabling the computation of ground-truth uncertainty. Code and data will are made publicly available at \url{https://github.com/WizgallF/udbench-benchmark}.

\end{abstract}

\section{Introduction}
\label{sec:introduction}

AI systems are increasingly deployed in safety-critical domains where uncertainty is essential for safety and interpretability. In such settings, uncertainty estimates can guide referral to human experts, additional testing, data acquisition, and decisions about when a model's output can be trusted \citep{ovadia2019trust,kompa_second_2021,kendall2017}. Crucially, \emph{why} a model is uncertain matters as much as \emph{how much}: uncertainty can be irreducible due to intrinsic task randomness (aleatoric uncertainty/AU), or reducible because the model is poorly optimized, lacks coverage, or is poorly specified (epistemic uncertainty/EU) \citep{hullermeier2021aleatoric}. This distinction is useful, for example, in applied clinical machine learning,
where EU is used to flag unreliable cases for review  \citep{faghani2023radiology,huang2024medimguq}. 
Proposed uncertainty concepts differ in the sources they cover and, consequently, in how they define and distinguish aleatoric and epistemic uncertainty \citep{kotelevskii2025, lahlou2023, jimenezjuergens2025}. Yet a universally accepted definition remains lacking.
Lacking this definition risks underestimating uncertainty because the employed definition itself may omit certain sources.

\paragraph{Probabilistic and risk-based concepts of uncertainty.}
Whilst probabilistic and risk-based concepts mostly agree on the definition of AU, they differ substantially on EU.
Probabilistic approaches commonly quantify EU using the predictor-induced posterior over functions. We distinguish the \emph{predictors' posterior} from a \emph{reference posterior} induced by the data-generating process and the observed sample. The two coincide only under exact Bayesian inference using the data-generating prior and likelihood. Otherwise, the spread of the predictors' posterior does not faithfully represent the distribution over plausible ground-truth functions. Moreover, a misspecified prior and/or approximate inference can introduce systematic bias into the model's predictions, in ways this notion of EU does not capture.
Conversely, risk-based approaches directly address the bias of a predictor by fixing one ground truth and measuring EU as excess risk of the predictor relative to it. As a definition, this acknowledges model misspecification but misses uncertainty induced by the limited sample. As a result, EU can be arbitrarily small in data-sparse regions whenever the predictor happens to match the ground-truth function, despite substantial uncertainty from the learner. This is unintuitive from the learner’s perspective, where limited data should generally imply greater uncertainty because the learner cannot know that it matched the ground-truth function.

\paragraph{Sample-conditional posterior risk.}
We propose a notion of ground-truth uncertainty that keeps the reference posterior separate from the estimator while allowing for estimator misspecification, optimization error, and EU due to a lack of data coverage to contribute to uncertainty. This notion is the \emph{sample-conditional posterior risk}: the expected loss of the deployed predictor under the reference posterior over plausible ground-truth functions. Intuitively, it asks how reliable the deployed predictor is across the ground-truth functions that remain plausible given what the learner has observed. Importantly, posterior risk should not serve as a new definition for an estimator but rather as a reliable, actionable ground-truth notion of what uncertainty should be from the learner's perspective. Defining a ground-truth enables, to our knowledge, the first direct evaluation of these different sources of EU together with AU and, by extension, uncertainty disentanglement.
Most existing benchmarks instead evaluate EU through proxy tasks such as out-of-distribution (OOD) detection or correlations between estimated uncertainty components \citep{mucsanyi2024}. Such evaluations are appropriate when performance on the proxy task itself is the objective: if EU is intended specifically for OOD detection, its usefulness for OOD detection is directly relevant. However, proxy tasks are less informative when the goal is to assess whether an estimator faithfully recovers EU as a general-purpose uncertainty quantity. In this case, success on a downstream task does not establish that the estimated EU is correct, and even perfect EU need not solve the proxy task \citep{li2025oodwrong,paplham2026evaluatingepistemicuncertaintyood}. Without a ground-truth EU target, the proxy can therefore become an implicit definition of what EU should be, rather than a test of whether EU has been estimated faithfully. Our benchmark makes this distinction testable by directly comparing uncertainty estimates with ground truth and examining how this direct evaluation relates to proxy-task performance.

\paragraph{Semi-synthetic evaluation with reference posteriors.}
To make sample-conditional posterior risk operational, we require a setting in which the reference posterior over ground-truth functions is explicit and samplable. We therefore instantiate the theory in a semi-synthetic tabular regression setting with real covariates and synthetic targets. To cover different priors, we instantiate the reference posterior through three distinct prior families over latent functions: Gaussian Processes (GPs), Heteroscedastic Bayesian Additive Trees (HBART), and Bayesian Neural Networks (BNNs). Importantly, the reference distribution is determined by the data-generating process and the observed sample, rather than by the estimator being evaluated. With this construction, the posterior is correctly specified with respect to the actual data-generating process. This enables direct computation of ground-truth uncertainty and therefore more fine-grained analysis than proxy evaluations. To summarize, we make the following contributions:

\begin{itemize}
\item \textbf{A sample-conditional posterior-risk formulation of uncertainty.}
We define uncertainty as the risk of the deployed predictor under a method-independent reference posterior conditioned on the observed sample, and locate posterior-spread and fixed-truth excess-risk notions as boundary cases.

\item \textbf{A suite of semi-synthetic tasks for uncertainty disentanglement.}
We release a large-scale suite of semi-synthetic tasks for uncertainty disentanglement evaluation, spanning $13$ datasets, three prior families, four aleatoric noise modes, and two data-coverage scenarios with the corresponding code. The methods' robustness to a change of ground-truth prior is also assessed. Each task includes a correctly specified reference posterior, enabling direct computation of ground-truth uncertainty and fine-grained method evaluation.

\item \textbf{A direct evaluation of $\mathbf{17}$ state-of-the-art uncertainty disentanglement methods.}
We benchmark $17$ state-of-the-art uncertainty disentanglement methods directly against posterior risk, alongside the most prevalent uncertainty definitions. We also quantify excess uncertainty entanglement and report the most robust and reliable methods for uncertainty disentanglement identified by our framework.
\end{itemize}

\section{Uncertainty and pointwise risk}





\label{sec:theory}

\begin{table}[t]
\caption{Notation used throughout the paper}
\label{tab:notation}
\centering

\begin{tablebox}[unbreakable]
\small
\setlength{\tabcolsep}{4pt}
\renewcommand{\arraystretch}{1.12}

\begin{tabularx}{\linewidth}{@{}lX@{}}
\toprule
\textbf{Symbol} & \textbf{Meaning} \\
\midrule

\(X,Y;\ x,y\)
&
Random input $X$ and outcome $Y$, and their realizations $x,y$. Their respective
spaces are \(\mathcal X\subseteq\mathbb R^d\) and \(\mathcal Y = \mathbb R\) for regression or \(\mathcal Y=\{0,\dots,K\}\) for  classification.
\\

\(f\)
&
Deterministic generative function \(f\in\mathcal F\), where
\(\mathcal F\) denotes the space of possible generative functions and where $F$ denotes a random variable over $\mathcal{F}$.
\\

\(P(\,\cdot\,)\)
&
Joint distribution over
\((X,Y)\in\mathcal X\times\mathcal Y\).
\\

\(P(\,\cdot\mid x, f)\)
&
Pointwise first-order distribution around the ground-truth function \(f\) with
$Y \mid (X=x, F=f) \sim P(\,\cdot\mid x,f)$, with density  $p(y\mid x,f)$ distributed w.r.t. the aleatoric noise model.
\\

\(\mathcal S\)
&
Random sample \(\mathcal S:=\{(x_i,y_i)\}_{i=1}^N\) with
\((x_i,y_i)\stackrel{\mathrm{i.i.d.}}{\sim}P(\,\cdot\mid x,f)\).
\\

\(a,\ \hat a_{\mathcal S}\)
&
Action \(a\in\mathcal A\), where \(\mathcal A\subseteq\mathcal Y\)
for a point prediction and \(\mathcal A=\mathcal P(\mathcal Y)\)
for a probabilistic prediction.
The predictor \(\hat a_{\mathcal S}:\mathcal X\to\mathcal A\) trained on \(\mathcal S\) and returns the action
\(\hat a_{\mathcal S}(x)\) at input \(x\).
\\

\(\ell\)
&
Loss \(\ell\colon\mathcal A\times\mathcal Y\to\mathbb R\), covering
pointwise losses and proper scoring rules.
\\

\(\Pi\)
&
Distribution over generative functions \(f\).
\\

\bottomrule
\end{tabularx}
\end{tablebox}
\end{table}
\paragraph{The Bayesian world view.}
We assume that observations arise from a real-world process described by a function $f$, which, together with observation noise, induces the conditional outcome distribution $P(\,\cdot\mid x,f)$.
We further assume a function-generating prior $\Pi_0$ describing which functions are plausible in the domain and their probabilities before observing any data. Conditioning this prior on some sample $\mathcal S$ using the corresponding observation model yields the correctly specified \emph{reference posterior} $\Pi_\mathrm{ref}(\,\cdot\mid\mathcal S)$ over plausible functions.

\subsection{Uncertainty as sample-conditional posterior risk}
\label{sec:posterior-risk}

Following the pointwise risk notion of uncertainty
\citep{lahlou2023,kotelevskii2022,kotelevskii2025,fishkov2025}, we take
pointwise risk as our starting point to later on combine the Bayesian world view with the risk-based notion of uncertainty.

\begin{definition}[First-order risk-based notion]
For a fixed input \(x\), the \emph{pointwise risk} of \(\hat a\) is
\begin{equation}
R(\hat a;x)
:=
\mathbb E_{Y\sim P(\,\cdot\mid x, f)}\!\left[\ell\bigl(\hat a(x),Y\bigr)\right]
=
\int_{\mathcal Y} \ell\bigl(\hat a(x),y\bigr)\,P(dy \mid x, f).
\label{eq:pointwise-risk}
\end{equation}
If \(\ell\) is a
strictly proper scoring rule, the Bayes act is
uniquely \(P(\,\cdot\mid x, f)\), where $f$ is a single deterministic ground-truth function $f^\star$.
\end{definition}

Our contribution is to expand $P(\,\cdot\mid x, f)$ by a construction that matches a learner's perspective. We argue that the Bayesian world-view is the only view that matches the learners' perspective in terms of quantifying uncertainty, as the learner does not observe the full \(P(\,\cdot\mid x, f^\star)\), only
the finite sample \(\mathcal S\) with limited coverage. Therefore, the distribution \(P(\,\cdot\mid x, f)\) is not clearly identified by \(\mathcal S\), inducing a second-order distribution over plausible outcome distributions around functions $f$. According to the Bayesian world view $f$ is distributed according to the reference posterior $\Pi_\mathrm{ref}(\,\cdot\mid\mathcal S)$. This induces an expanded notion of risk-based uncertainty, defined as the risk of a predictor after seeing $\mathcal S$ over the distribution of plausible real-world processes $f$ according to $\Pi_\mathrm{ref}(\,\cdot\mid\mathcal S)$. 


\begin{definition}[Sample-conditional pointwise posterior risk]
For a fixed input \(x\) and observed sample \(\mathcal S\), the
sample-conditional pointwise posterior risk of a deployed predictor
\(\hat a_{\mathcal S}\) is
\begin{equation}
\begin{aligned}
R_{\mathcal S}(\hat a_{\mathcal S};x)
&:=
\mathbb E_{F\sim\Pi_\mathrm{ref}(\,\cdot\mid\mathcal S)}
\mathbb E_{Y\sim P(\,\cdot\mid x, F)}
\!\left[\ell\bigl(\hat a_{\mathcal S}(x),Y\bigr)\right]
\\
&=
\int_{\mathcal F}\int_{\mathcal Y}
\ell\bigl(\hat a_{\mathcal S}(x),y\bigr)
P(dy\mid x, f)\,\Pi_\mathrm{ref}(df\mid\mathcal S)
\\
&=
\mathbb E_{Y\sim P(\,\cdot\mid x,\mathcal S)}
\!\left[\ell\bigl(\hat a_{\mathcal S}(x),Y\bigr)\right]\end{aligned}
\label{eq:gtruth-risk}
\end{equation}

The final line uses the posterior predictive mixture $P(\,\cdot\mid x,\mathcal S)=\int_{\mathcal F}P(\,\cdot\mid x, f)\,\Pi_\mathrm{ref}(df\mid\mathcal S).$

\end{definition}

We follow \citet{kendall2017, lahlou2023} and define total uncertainty as the sum of aleatoric and epistemic uncertainty. Additionally, we define sample-conditional pointwise posterior risk as the total uncertainty. 

\begin{definition}[Disentanglement of the pointwise posterior risk]
For each generative function $f$, define its conditional Bayes risk as
$R_f^\star(x)
:=
\inf_{a\in\mathcal A}
\mathbb E_{Y\sim P(\,\cdot\mid x, f)}[\ell(a,Y)].$
Aleatoric uncertainty is the posterior expectation of this risk,
and epistemic uncertainty is the posterior expected excess risk:
\begin{equation}
\begin{aligned}
\mathrm{AU}_{\mathcal S}(x)
&:=
\mathbb E_{F\sim\Pi_\mathrm{ref}(\,\cdot\mid\mathcal S)}
\!\left[R_F^\star(x)\right],
\\
\mathrm{EU}_{\mathcal S}(\hat a_{\mathcal S};x)
&:=
\mathbb E_{F\sim\Pi_\mathrm{ref}(\,\cdot\mid\mathcal S)}
\!\left[
\mathbb E_{Y\sim P(\,\cdot\mid x, F)}
[\ell(\hat a_{\mathcal S}(x),Y)]
-
R_F^\star(x)
\right]
\\
&=
R_{\mathcal S}(\hat a_{\mathcal S};x)
-
\mathrm{AU}_{\mathcal S}(x).
\end{aligned}
\label{eq:au-eu}
\end{equation}
Thus, AU is the expected posterior Bayes risk that would remain even if the
generative function were known, while EU measures the deployed
predictor's expected posterior excess risk over predicting the generative function.
Their sum equals the pointwise posterior risk, which quantifies
total uncertainty.
\end{definition}

Following \citet{adlam_understanding_2022}, we use the Bregman divergence framework and the generalized bias--variance decomposition of \citet{pfau2025}. The Bregman form provides a general AU--EU decomposition across all proper losses considered here. We can thus define the split once and obtain each loss-specific instantiation by substituting the corresponding divergence, recovering known quantities, for example, the Mutual Information as part of EU \citep{depeweg2018decomposition, smith2018}, as shown in Appendix~\ref{app:proofs}. The abstraction therefore provides generality over the most prevalent proper losses \citep{bülte2026uncertainty}.

\begin{proposition}[Bregman decomposition of posterior risk]
Let \(\mathcal D\) be a convex representation of the action space. With a slight
abuse of notation, throughout the Bregman analysis we identify \(Y\) with its
canonical representation in \(\mathcal D\): the observed value under squared
loss, its one-hot encoding under multiclass log loss, and the function
\(t\mapsto\mathbf 1\{Y\leq t\}\) under CRPS. Let
\(\phi:\mathcal D\to\mathbb R\) be strictly convex and differentiable, with
Bregman divergence
\(D_\phi(u\Vert v):=\phi(u)-\phi(v)
-\langle\nabla\phi(v),u-v\rangle\).
Assume that \(\ell(a,Y)=D_\phi(Y\Vert a)\) and that all quantities below are
well-defined. Write \(\mathbb E_F\) for expectation under
\(\Pi_{\mathrm{ref}}(\,\cdot\mid\mathcal S)\) and
\(\mathbb E_{Y\mid F}\) for expectation under \(P(\,\cdot\mid x,F)\).
Then \(a_f(x):=\mathbb E_{Y\sim P(\,\cdot\mid x,f)}[Y]\) is the conditional
Bayes act, and
\(\bar a_{\mathrm{ref}}(x):=\mathbb E_F[a_F(x)]\) is the posterior Bregman
centroid. The pointwise posterior risk decomposes as
\begin{equation}
\begin{aligned}
R_{\mathcal S}(\hat a_{\mathcal S};x)
&=
\underbrace{
\mathbb E_F\!\left[
\mathbb E_{Y\mid F}\!\left[
D_\phi\!\left(
Y\,\middle\Vert\,a_F(x)
\right)
\right]
\right]
}_{\textbf{AU}}
+
\underbrace{
\textcolor{post-risk}{
\mathbb E_F\!\left[
D_\phi\!\left(
a_F(x)\,\middle\Vert\,\hat a_{\mathcal S}(x)
\right)
\right]}
}_{\textbf{EU}}
\\
&=
\underbrace{
\mathbb E_F\!\left[R_F^\star(x)\right]
}_{\text{Bayes error}}
+
\underbrace{
\textcolor{bayes}{
\mathbb E_F\!\left[
D_\phi\!\left(
a_F(x)\,\middle\Vert\,\bar a_{\mathrm{ref}}(x)
\right)
\right]}
}_{\textcolor{bayes}{\text{generalized variance}}}
+
\underbrace{
\textcolor{freq}{
D_\phi\!\left(
\bar a_{\mathrm{ref}}(x)\,\middle\Vert\,
\hat a_{\mathcal S}(x)
\right)}
}_{\textcolor{freq}{\text{deployment gap}}}.
\end{aligned}
\label{eq:bregman}
\end{equation}
We call the final term the \emph{deployment gap} to distinguish it from
frequentist bias averaged over repeated samples.
\end{proposition}

The Bregman decomposition identifies AU as the posterior expected
Bayes risk: the irreducible loss that remains even when the underlying
generative function is known, independently of the deployed predictor.
EU combines two terms: a predictor-independent generalized variance,
which captures the spread of plausible ground-truth functions given $\mathcal S$, and a deployment
gap, which measures the Bregman divergence between the corresponding
centroid and the deployed prediction.
Under squared loss, these terms reduce to the posterior variance and
the squared deviation of the deployed prediction from the posterior mean.

\begin{figure*}[!htbp]
    \centering
    \includegraphics[width=\textwidth]{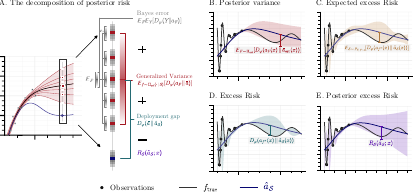}
    \caption{\textbf{Decomposition of posterior risk and pathologies of alternative uncertainty notions.}
    A: Visual decomposition of \Cref{eq:bregman}. B–E: Different uncertainty notions retain only parts of this decomposition and can therefore behave pathologically. Posterior variance can remain small for a misspecified or underfit predictor (B), while excess-risk notions can assign low uncertainty when the predictor happens to be close to the realized ground truth despite limited data (C, D). Posterior excess risk retains both uncertainty over plausible functions and error of the deployed predictor (E).}
    \label{fig:freq-vs-bayes}
\end{figure*}

\subsection{The relationship of posterior risk to concurrent notions of uncertainty}
\label{sec:gp-oof}

\paragraph{\textcolor{bayes}{Bayesian Posterior-spread notion.}}
A common Bayesian view identifies EU with the spread of an estimator posterior
\(\Pi_{\mathrm{est}}(\,\cdot\mid\mathcal S)\) over functions or parameters. In
the Bregman setting of \Cref{eq:bregman}, let
\(\bar a_{\mathrm{est}}(x)\) denote its corresponding posterior centroid. At
this posterior-optimal action, total uncertainty decomposes as
\[
\widehat{\mathrm{TU}}_{\mathrm{Bayes}}(x)
=
\underbrace{
\mathbb E_{F\sim\Pi_{\mathrm{est}}}
\mathbb E_{Y\mid F}
\left[
D_\phi\!\left(
Y\,\middle\Vert\,a_F(x)
\right)
\right]
}_{\widehat{\mathrm{AU}}_{\mathrm{Bayes}}(x)}
+
\underbrace{
\textcolor{bayes}{
\mathbb E_{F\sim\Pi_{\mathrm{est}}}
\left[
D_\phi\!\left(
a_F(x)\,\middle\Vert\,\bar a_{\mathrm{est}}(x)
\right)
\right]}
}_{\widehat{\mathrm{EU}}_{\mathrm{Bayes}}(x)}.
\]
Bayesian EU is therefore the generalized spread of the estimator posterior. It
equals posterior-risk EU only when the estimator posterior matches the
reference posterior, and its centroid is correct. Inference within the
estimator's model does not guarantee this match. Differences in prior,
likelihood, hypothesis class, or approximate inference can make the posteriors
arbitrarily different in center and spread, and the estimator centroid need not
be reference-optimal. In \Cref{fig:freq-vs-bayes} Panel B, insufficient model
capacity accordingly produces an inaccurate, underfit prediction with low
Bayesian uncertainty. Under correct specification, exact inference, and
centroid deployment, the deployment gap vanishes by construction, leaving
posterior spread as the only EU term.

Posterior-risk EU instead evaluates the deployed predictor under the reference
posterior:
\[
\textcolor{post-risk}{\mathrm{EU}_{\mathcal S}(\hat a_{\mathcal S};x)}
=
\underbrace{
\textcolor{bayes}{
\mathbb E_{F\sim\Pi_{\mathrm{ref}}(\,\cdot\mid\mathcal S)}
\left[
D_\phi\!\left(
a_F(x)\,\middle\Vert\,\bar a_{\mathrm{ref}}(x)
\right)
\right]}
}_{\text{generalized variance}}
+
\underbrace{
\textcolor{freq}{
D_\phi\!\left(
\bar a_{\mathrm{ref}}(x)
\,\middle\Vert\,
\hat a_{\mathcal S}(x)
\right)}
}_{\text{deployment gap}}.
\]
The estimator spread need not equal the first term, while the second measures
the deviation of the deployed predictor from the reference-posterior centroid.
Posterior-spread EU alone therefore does not account for misspecification and
can assign low uncertainty to arbitrarily inaccurate predictions. It is not a
reliable general ground-truth target, although it may approximate posterior-risk
EU when posterior mismatch and the deployment gap are small. Posterior spread
can nevertheless serve two roles. Under a probabilistic view, it estimates the
reference-posterior spread that defines EU. Under a risk-based view, fixed-truth
excess risk remains the target, while Bayesian machinery provides a practical
estimator of that target \citep{kotelevskii2025,fishkov2025}. A Bayesian
estimator therefore need not imply a probabilistic definition of EU.

\paragraph{\textcolor{freq}{Risk-based notions.}}
First-order risk-based notions instead fix one ground truth \(f^\star\) and
define EU by deviation from its Bayes act, which is equivalent to setting
\(\Pi_{\mathrm{ref}}(\,\cdot\mid\mathcal S)=\delta_{f^\star}\). With
\(a_{f^\star}(x):=
\mathbb E_{Y\sim P(\,\cdot\mid x,f^\star)}[Y]\), \Cref{eq:bregman} gives
\[
\mathrm{TU}_{f^\star}(\hat a_{\mathcal S};x)
=
\underbrace{
\mathbb E_{Y\sim P(\,\cdot\mid x,f^\star)}
\left[
D_\phi\!\left(
Y\,\middle\Vert\,a_{f^\star}(x)
\right)
\right]
}_{\mathrm{AU}_{f^\star}(x)}
+
\underbrace{
\textcolor{freq}{
D_\phi\!\left(
a_{f^\star}(x)\,\middle\Vert\,\hat a_{\mathcal S}(x)
\right)}
}_{\mathrm{EU}_{f^\star}(\hat a_{\mathcal S};x)}.
\]
Collapsing the reference posterior removes the generalized variance and
measures the deployment gap relative to the fixed Bayes act
\(a_{f^\star}(x)\), rather than the reference-posterior centroid
\(\bar a_{\mathrm{ref}}(x)\). Pointwise excess risk therefore measures realized
estimator error, but not uncertainty from the incomplete identification of
\(f^\star\) by \(\mathcal S\). It can vanish in data-sparse regions whenever
the deployed predictor happens to agree with the realized Bayes act, as shown
in \Cref{fig:freq-vs-bayes} Panel C, although the sample cannot establish that
this agreement has occurred. It can thus assign low EU precisely where
uncertainty induced by limited data remains substantial.

Second-order risk-based notions instead introduce a distribution over repeated
training samples \(\mathcal S_i\), rather than plausible ground-truth
functions, and measure expected error across repeated experiments
\citep{jürgens2024epistemicuncertaintyfaithfullyrepresented,jimenezjuergens2025}.
They provide a frequentist second-order analogue of posterior risk, but average
over hypothetical samples instead of conditioning on the realized
\(\mathcal S\). They therefore do not represent uncertainty induced by the
coverage of that realized sample.

\section{Related Work}

Beyond the notions in \Cref{sec:gp-oof}, concurrent work by
\citet{alamri2026subjectiverisk} decomposes subjective risk under the predictive distribution, recovering several existing AU and EU measures under strictly proper losses. This is orthogonal to our goal of fixing a method-independent reference posterior to define ground-truth targets for estimator evaluation rather than unifying estimator-side measures.

On the evaluation side of uncertainty literature, most works utilize proxy tasks. These
studies report that predictive entropy can conflate ambiguous in-distribution
and OOD inputs \citep{mukhoti2021pitfalls}, deterministic uncertainty methods
can combine competitive OOD detection with poor calibration under distribution shift \citep{postels22a}, and regression uncertainty estimators
can become strongly overconfident under real-world shifts
\citep{gustafsson2023regshift}. Other indirect diagnostics examine interactions
between AU and EU estimates \citep{valdenegro2022deeper} or use their rank
correlation as one disentanglement criterion \citep{mucsanyi2024}. These
findings identify limitations of particular estimators or scores, but do not
distinguish inaccurate EU estimation from a mismatch between EU and the proxy
objective. \citet{li2025oodwrong} argue that a supervised classifier's
predictive uncertainty, including conventional model-posterior EU, does not
generally determine OOD membership.
\citet{paplham2026evaluatingepistemicuncertaintyood} similarly show that
Bayes-optimal strategies for OOD detection and active learning do not generally
coincide with scores used to quantify EU. Proxy performance therefore depends
on both estimator quality and objective alignment.

More direct evaluations use controlled simulations of physical systems with
quantified measurement uncertainty \citep{caldeira2021deeplyuncertain} or
compare approximate methods against HMC-derived EU on synthetic problems
\citep{schweighofer2023quam}. Concurrent work by
\citet{paplham2026evaluatingepistemicuncertaintyood} also derives an equivalent
decomposition of Bayesian expected risk into posterior-expected conditional
Bayes risk and posterior-expected regret. Their benchmark evaluates the
resulting AU and EU estimates through risk--regret--coverage trade-offs on
datasets with densely repeated human annotations. Their empirical targets are
estimated from annotator-derived conditional label distributions, whereas ours
are analytic pointwise targets derived from a declared generative process and
its sample-conditional reference posterior and further separate EU into
posterior spread and the deployment gap.

\section{Evaluation protocol}
\label{sec:protocol}

We define posterior risk as the ground-truth target for uncertainty estimates. To test how closely state-of-the-art methods align with this notion of uncertainty, and to identify the ground-truth notion they most closely approximate, we construct a set of 13 semi-synthetic datasets with reference posteriors that allow us to calculate ground-truth uncertainty measures. We instantiate each dataset using three different prior families (GP, HBART, and BNN; Appendix~\ref{app:data:priors}), four aleatoric noise models (Appendix~\ref{app:data:noise}), and two training-sample coverage scenarios (Appendix~\ref{app:data:sample}). This results in 312 tasks, each repeated over 10 seeds. In each task, we rank-correlate the models' uncertainty estimates with the ground-truth uncertainty using Spearman's $\rho$ to directly assess the reliability of their uncertainty approximations.

\paragraph{Semi-synthetic tasks and ground-truth targets.}
Ground-truth uncertainty targets require both the aleatoric noise model and the reference posterior $\Pi_{\mathrm{ref}}(\,\cdot \mid \mathcal{S})$, neither of which is available for real data.
We therefore retain the covariates of 13 real tabular datasets and generate the targets from a known prior over functions with known noise (Appendix~\ref{app:data}).
Conditioning the same prior on the training sample yields a reference posterior that is correctly specified by construction with respect to the data-generating process.
As ground-truth AU, we use the specified noise model's variance. For EU, we define posterior excess risk, instantiated under squared loss, as the primary target. To assess alternative notions of EU, we also use reference-posterior variance and squared deviation from the ground-truth function as secondary targets.

\paragraph{Metrics.}
Because the absolute scale of risk depends on the evaluation loss, which is not specified to the methods, and many downstream tasks require only a reliable ordering of inputs, we use Spearman's $\rho$ as our primary metric. We average $\rho$ over all task instances of a dataset and report the min--max seed range across repeated experiments. To determine the stability of the method ranking across configurations, we report Kendall's $W$. For each axis (dataset, noise mode, prior family, and covariate scenario), we fix that axis while averaging over all others, thereby measuring its influence on the model rankings. Correlation with both ground-truth targets does not establish disentanglement, because an estimate of total uncertainty may correlate with both.
We therefore measure excess entanglement,
\begin{equation}
\Delta_{\mathrm{ent}}
=
\rho(\widehat{\mathrm{EU}}, \widehat{\mathrm{AU}})
-
\rho(\mathrm{EU}^\star, \mathrm{AU}^\star),
\label{eq:excess-disentanglement}
\end{equation}

where hats denote a method's estimates and stars denote the corresponding ground-truth values.
Positive values indicate that the estimates depend on one another more strongly than the ground-truth quantities do. To test whether OOD detection serves as a good proxy for EU estimation, we additionally perform an OOD detection experiment using the White Wine UCI dataset as in-distribution data and the Red Wine dataset as OOD data. We then compare the ranking of models according to their AUROC in OOD detection, based on their EU estimates, with the ranking obtained from our direct Spearman-correlation-based benchmark.

\paragraph{Methods.}
We evaluate 17 methods from four families: Bayesian linear regression (BLR), GP-based models, tree ensembles and Neural Networks (NNs).
Each method returns a prediction, an AU estimate, and an EU estimate, and all methods are trained using standard architecture and training settings for each task instance (Appendix~\ref{app:benchmarked-models}).

\section{Evaluation}
\label{sec:evaluations}
We begin by asking how closely the methods’ uncertainty estimates track ground-truth EU and AU, which EU targets they capture most faithfully, and whether their relative performance is stable across tasks. Figure 2 summarizes performance across the 13 datasets using the mean Spearman rank correlation with the oracle EU and AU targets. Across methods, correlations range from \(\rho=0.16\)
to \(\rho=0.69\) for EU and from \(\rho=0.42\) to \(\rho=0.86\) for AU. VLGP and SNGP-Het achieve the
highest agreement with both targets, whereas NN-DEUP and NN-EDL perform the worst, which, for NN-EDL,
is in line with the failure modes identified by \citet{jürgens2024epistemicuncertaintyfaithfullyrepresented}. For every method, EU
estimates agree more closely with reference-posterior variance than with posterior excess risk and
show weak agreement with pointwise excess risk.
Despite these differences in absolute performance, the method ranking is stable across the benchmark
axes of prior family, noise mode, and covariate scenario. Kendall's \(W\), computed separately along
each axis, is lowest across prior families (\(W=0.81\) for the EU and \(0.93\) for the AU ranking, \Cref{tab:kendall_w}). Part of the variation across
prior families is a matched-prior advantage: methods rank better when the ground truth is
drawn from a prior like their own. The best and worst methods, however, are the same under every oracle (\Cref{fig:home_ranks} in Appendix
~\ref{app:stability_rankings}). Additionally rankings were stable across datasets (EU Kendall's $W=0.78$, AU $W=0.89$ see \Cref{fig:abalone_ranking}, \ref{fig:airfoil_ranking}). Strong performance therefore does not require an exactly specified
prior, and several methods remain effective across all generative settings considered.

\begin{figure}
    \centering
    \includegraphics[width=1\linewidth]{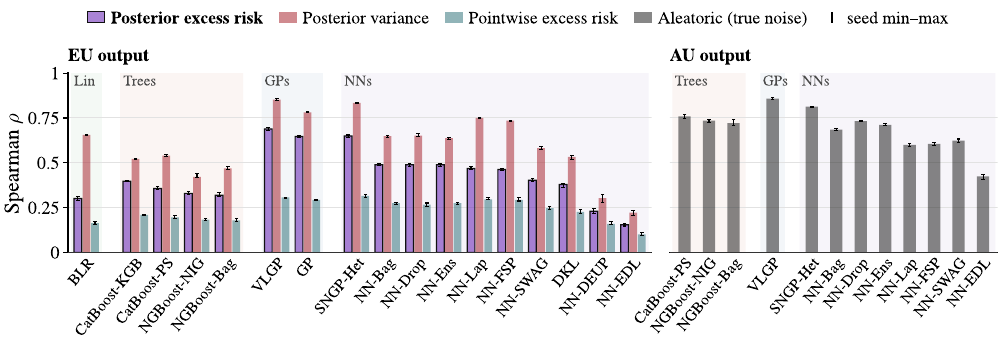}
    \caption{\textbf{Main benchmark results across the 13 datasets.} Bars show the mean
    Spearman rank correlation between estimated and oracle epistemic or
    aleatoric uncertainty. Error bars show the minimum-to-maximum range across
    repeated experiments with 10 different random seeds. SNGP-Het and VLGP
    achieve the highest rank agreement for both EU and AU. Across methods, EU
    estimates track posterior variance more closely than posterior risk and
    track pointwise excess risk poorly.}
    \label{fig:placeholder}
\end{figure}

\begin{figure}
    \centering
    \includegraphics[width=1\linewidth]{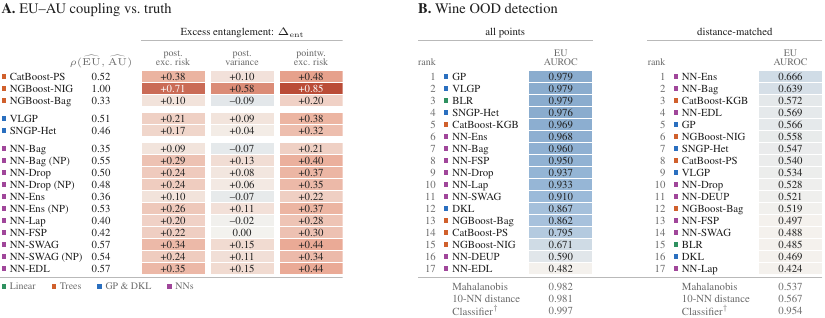}
    \caption{\textbf{A. EU--AU coupling.} $\rho(\widehat{\mathrm{EU}}, \widehat{\mathrm{AU}})$: Spearman
        correlation between a method's EU and AU outputs, averaged over all conditions. $\Delta_{\mathrm{ent}}$:
        excess over the ground-truth coupling for three EU definitions; $\Delta_{\mathrm{ent}} > 0$ means the
        outputs are more coupled than the truth. (NP): natural-parameter head.
        \textbf{B. Wine OOD detection.} AUROC of the EU output for red vs.\ held-out white wine (models trained
        on white wine), on all points and after matching Mahalanobis distance to the training data, where
        distance alone scores $0.5$. $^\dagger$Supervised red-vs-white classifier, an upper bound on separability.}
    \label{fig:disentanglement_ood}
\end{figure}

\paragraph{Are methods AU-EU estimates overly entangled?}
\Cref{fig:disentanglement_ood}A shows the excess entanglement (\Cref{eq:excess-disentanglement}) of each method and demonstrates that its magnitude depends on the definition of EU. When EU is defined as posterior excess risk, methods generally exhibit stronger EU--AU coupling than the corresponding ground truth. This discrepancy is smaller when EU is defined as posterior variance. Across the benchmark, ground-truth posterior variance and AU have a mean Spearman correlation of \(\rho=0.42\), which reflects the true degree of the ground-truth correlations. Because most evaluated methods aim to estimate posterior variance, this target provides an interesting comparison. Relative to posterior variance, excess coupling is largest for NGBoost-NIG, followed by NN-EDL and NN-SWAG. Motivated by the improved training stability reported by \citet{immer2023NPLoss}, we also evaluated natural-parameter heads but found that they increase excess entanglement for some methods. Appendix~\ref{app:disentanglement} discusses possible reasons for the excess entanglement of EDL methods and natural-parameter heads. Accounting for the ground-truth EU--AU correlation suggests that excess entanglement is, at least w.r.t.~posterior variance, less severe than previously reported \citep{mucsanyi2024}. Moreover, method rankings by excess entanglement differ from those by agreement with ground-truth EU and AU. This discrepancy demonstrates the need for direct targets because low excess entanglement alone does not imply faithful uncertainty estimates in our experiments.

\paragraph{Is OOD detection a good proxy for EU?}
We compare our benchmark ranking with a red- vs.\ white-wine OOD task
\citep{openml_red_wine_44972, openml_white_wine_44971}. Methods are trained on white wine and use their
EU to separate red from held-out white wine (Figure~\ref{fig:disentanglement_ood}B). Whether the two
rankings agree depends on the type of shift. Red wine as a whole lies far from white wine, so that
Mahalanobis distance alone detects it (AUROC $0.98$). For this distance-type shift, the OOD detection ranking
closely matches our benchmark (Kendall's $W = 0.89$, where $1$ means identical rankings). We then match red- and white wine evaluation points in distance to the training data. The red wines still differ in feature combinations and have higher GP-oracle uncertainty compared to white wines (Appendix~\ref{app:OOD}). But the OOD method rankings no longer agree ($W = 0.49$--$0.64$, where $0.5$ means no association). The ensembles now detect red wine best, because
the disagreement between their members grows on red wine. The EU of VLGP and SNGP-Het, the best methods
in our benchmark, hardly changes, although they track the ground-truth EU within the white-wine data
far more closely than the ensembles. This has two consequences. First, a method that detects one type
of shift well is not guaranteed to detect others. Second, OOD detection cannot directly evaluate how
well EU is estimated, because which methods it favours depends on the chosen OOD task.

\paragraph{Does Spectral Normalization help NNs in estimating uncertainty?}
SNGP-Het tracks the posterior excess risk far better than DKL ($\rho = 0.65$ vs.\ $0.38$). Both place
a GP on learned features, but SNGP spectrally normalizes its backbone and uses a differently constructed and approximated GP output layer. Without such a constraint, feature extractors
are known to collapse: inputs from unrelated regions are mapped onto the features of training points,
so no output layer can recognize them as novel \citep{liu2020sngp, vanamersfoort2022featurecollapsedeepkernel}.
Therefore the question arose if spectral normalization (SN) can help the other NN-based methods in estimating uncertainty.
We retuned eight NN methods with SN and, as a control, strong weight 
decay (WD) to make sure that the SN effect was not attributable to stronger regularization (Table~\ref{tab:sn_study}). 
SN raises local distance preservation for every 
method (pooled $+0.21$ vs.\ $+0.01$ for strong WD; Appendix~\ref{app:spectral_normalization}). For the four methods 
whose EU is read from a single network's features, SN improves the correlation with the 
posterior excess risk by $+0.10$ to $+0.11$ and with the posterior variance by up to 
$+0.26$. WD recovers at most half of this gain. For NN-Lap and NN-FSP, the gain is mediated 
by distance preservation and grows with depth, since deeper plain networks preserve 
distances worst (Table~\ref{tab:sn_hparams}). For ensembles and MC dropout, whose EU arises 
from disagreement between models, SN leaves EU quality unchanged. SN preserves distances for uncertainty methods that read from a single representation, especially Laplace methods. Its gains grew with depth even though our backbones had at most four residual blocks, suggesting SN might be even more important for larger networks.

\section{Limitations}
\label{sec:limitations}
Our ground-truth targets depend on the chosen prior family, so results may differ for priors outside our benchmark even though method rankings were broadly stable across those tested. Computational constraints limited the BNN priors to small networks, which may favor smoother functions than larger architectures would have produced as exact BNN inference at larger scales remains costly \citep{izmailov2021bayesianneuralnetworkposteriors}. Moreover, the Bregman decomposition covers most losses considered here but not every practically relevant objective, including general weighted losses. Corresponding decompositions are left for future work. Finally, our experiments cover only regression, chosen to permit controlled heteroscedastic noise and complement classification-focused prior work \citep{mukhoti2021pitfalls, postels22a, valdenegro2022deeper, mucsanyi2024}. Extending the benchmark to classification is left for future work.

\section{Conclusion}
\label{sec:conclusion}
We introduced posterior risk as a ground-truth measure of uncertainty capturing a predictor’s expected loss over plausible ground-truth functions. Its method-independent reference posterior retains uncertainty induced by limited data without assuming that an estimator’s posterior is correct. Posterior risk also captures errors arising from misspecification or optimization through the deployment gap, while its Bayes-risk component represents irreducible outcome uncertainty. It therefore provides a common target for evaluating the main sources of AU and EU and enables more fine-grained analysis than estimator-internal or fixed-truth targets. On our semi-synthetic benchmark, SNGP-Het and VLGP showed the strongest overall agreement with posterior-risk targets. Most methods tracked posterior variance more closely than posterior risk. Their AU and EU estimates were also frequently more coupled than their ground-truth counterparts, although the extent depended on the EU definition. Spectral normalization improved uncertainty quality through distance preservation, particularly for Laplace approximation methods. Direct evaluation and proxy tasks also yielded different method rankings, supporting the use of ground-truth targets alongside downstream benchmarks. 

\section*{Acknowledgements}
We acknowledge the financial support of the Gemeinnützige Hertie-Stiftung and the Deutsche
Forschungsgemeinschaft (DFG, German Research Foundation) – 459422098.
%
%
%

\newpage

\bibliography{iclr2027_conference}
\bibliographystyle{iclr2027_conference}


\appendix


\clearpage

\startcontents[appendix]

\section*{Appendix Contents}
\begingroup
\setcounter{tocdepth}{2} 
\printcontents[appendix]{}{1}{}
\endgroup

\newpage

%

\ifcsname lemma\endcsname\else\newtheorem{lemma}[definition]{Lemma}\fi

\newcommand{\losscard}[8]{%
  \begin{tablebox}
  \small
  \renewcommand{\arraystretch}{1.18}%
  \begin{tabularx}{\linewidth}{@{}>{\raggedright\arraybackslash}p{0.30\linewidth}X@{}}
  \textbf{Scoring type} & #1 \\
  \textbf{Outcome space} \(\mathcal Y\) & #2 \\
  \textbf{Action space} \(\mathcal A\), deployed action & #3 \\
  \textbf{Loss} \(\ell(a,y)\) & #4 \\
  \textbf{Representation of} \(Y\) \textbf{in} \(\mathcal D\) & #5 \\
  \textbf{Generator and divergence} & #6 \\
  \textbf{Conditional Bayes act} \(a_f(x)\) & #7 \\
  \textbf{Posterior centroid} \(\bar a_{\mathrm{ref}}(x)\) & #8 \\
  \end{tabularx}
  \end{tablebox}}

\section{Loss-specific instantiations of the posterior-risk decomposition}
\label{app:proofs}

\subsection{Common setup and proof of Equation 4}
\label{app:setup}

\paragraph{Conventions.}
Throughout this appendix we fix an input \(x\) and the observed sample
\(\mathcal S\). As in the main text, \(\mathbb E_F\) denotes expectation over
\(F\sim\Pi_{\mathrm{ref}}(\cdot\mid\mathcal S)\) and \(\mathbb E_{Y\mid F}\)
expectation over \(Y\sim P(\cdot\mid x,F)\); for a fixed generative function
\(f\) we write \(\mathbb E_{Y\mid f}\) for expectation over
\(Y\sim P(\cdot\mid x,f)\). The posterior predictive is
\(P(\cdot\mid x,\mathcal S)=\mathbb E_F\bigl[P(\cdot\mid x,F)\bigr]\)
(\Cref{eq:gtruth-risk}). Within each subsection the deployed action
\(\hat a_{\mathcal S}(x)\) is abbreviated by a loss-specific symbol, stated in
its setting card. As in \Cref{eq:bregman}, \(\phi\) is strictly convex; for
the function-valued cases (CRPS and densities) this holds after identifying
functions that agree almost everywhere. We assume that all displayed
expectations and integrals are finite and that expectations may be exchanged
(Fubini). Wherever the gradient of \(\phi\) is used, it is assumed to exist at
the second argument of \(D_\phi\); boundary cases of the KL divergence and the
CRPS are read through the explicit divergence formulas in the setting cards.

\paragraph{Point and distributional scoring.}
Under \emph{point (value) scoring}, the action is a point
\(a\in\mathcal A\subseteq\mathcal Y\) and the conditional Bayes act is a
functional of the conditional law; for the point loss considered below it is
the conditional mean. Under \emph{distributional scoring}, the action is a
predictive distribution \(a\in\mathcal A=\mathcal P(\mathcal Y)\) scored by a
strictly proper scoring rule, and the conditional Bayes act is the conditional
law itself. \Cref{eq:bregman} covers both cases whenever outcome and action
admit representatives in a common convex set \(\mathcal D\) with
\(\ell(a,Y)=D_\phi(Y\Vert a)\). The continuous log score does not; it is
covered by \Cref{lem:score-form} below.

\paragraph{Compensation identity.}
For any \(\mathcal D\)-valued random variable \(U\) with mean
\(\bar u:=\mathbb E[U]\in\mathcal D\) and any \(a\in\mathcal D\),
\begin{equation}
\mathbb E\bigl[D_\phi(U\Vert a)\bigr]
=
\mathbb E\bigl[D_\phi(U\Vert \bar u)\bigr]
+
D_\phi(\bar u\Vert a).
\label{eq:compensation}
\end{equation}
In the difference of the two expectations the terms \(\phi(U)\) cancel and
\(\mathbb E[U-\bar u]=0\) removes the linear term at \(\bar u\), leaving
\(\phi(\bar u)-\phi(a)-\langle\nabla\phi(a),\bar u-a\rangle=D_\phi(\bar u\Vert a)\).
Because \(\phi\) is strictly convex, \(D_\phi(\bar u\Vert a)\geq 0\) with
equality if and only if \(a=\bar u\), so the mean is the unique minimizer of
\(a\mapsto\mathbb E[D_\phi(U\Vert a)]\) \citep{banerjee2005clustering}. For a
merely convex \(\phi\), the mean would still be a minimizer, but not
necessarily the only one.

\paragraph{Proof of \Cref{eq:bregman}.}
Applying \eqref{eq:compensation} to \(U=Y\sim P(\cdot\mid x,f)\) for fixed
\(f\) shows that \(a_f(x)=\mathbb E_{Y\mid f}[Y]\) is the conditional Bayes act,
that \(\mathbb E_{Y\mid f}\bigl[D_\phi(Y\Vert a_f(x))\bigr]=R_f^\star(x)\), and
that
\[
\mathbb E_{Y\mid f}\bigl[D_\phi\bigl(Y\Vert\hat a_{\mathcal S}(x)\bigr)\bigr]
=
R_f^\star(x)
+
D_\phi\bigl(a_f(x)\Vert\hat a_{\mathcal S}(x)\bigr).
\]
Taking \(\mathbb E_F\) gives the first line of \Cref{eq:bregman},
\(R_{\mathcal S}(\hat a_{\mathcal S};x)
=\mathrm{AU}_{\mathcal S}(x)
+\mathbb E_F\bigl[D_\phi\bigl(a_F(x)\Vert\hat a_{\mathcal S}(x)\bigr)\bigr]\).
Applying \eqref{eq:compensation} once more, now to \(U=a_F(x)\) with mean
\(\bar a_{\mathrm{ref}}(x)\), splits the epistemic term into the generalized
variance and the deployment gap. \qed

\paragraph{The centroid is the Bayes act of the posterior predictive.}
Since
\[
\bar a_{\mathrm{ref}}(x)
=\mathbb E_F\,\mathbb E_{Y\mid F}[Y]
=\mathbb E_{Y\sim P(\cdot\mid x,\mathcal S)}[Y],
\]
the posterior Bregman centroid is the Bayes act of the posterior predictive. For every distributional
score below it is therefore the posterior predictive itself; for the point
loss it is the corresponding functional of it.

\paragraph{Outcomes without a representative in \(\mathcal D\).}
For the log score with continuous outcomes, the natural representative of an
observation is the point mass \(\delta_Y\), which has no Lebesgue density, so
\(\ell(a,Y)=D_\phi(Y\Vert a)\) is unavailable pointwise. The same three-term
decomposition nonetheless holds for every proper scoring rule.

\begin{lemma}[Scoring-rule form of the decomposition]
\label{lem:score-form}
Let \(\ell\) be a proper scoring rule on a convex class \(\mathcal P\) of
distributions that contains \(P_f:=P(\cdot\mid x,f)\) for
\(\Pi_{\mathrm{ref}}\)-almost every \(f\), the posterior predictive
\(\bar P_{\mathrm{ref}}:=P(\cdot\mid x,\mathcal S)\), and the deployed action
\(\hat Q:=\hat a_{\mathcal S}(x)\), and let the action space be
\(\mathcal A=\mathcal P\). Define the generalized entropy and the
score divergence
\[
H_\ell(P):=\mathbb E_{Y\sim P}\bigl[\ell(P,Y)\bigr],
\qquad
d_\ell(P\Vert Q):=\mathbb E_{Y\sim P}\bigl[\ell(Q,Y)\bigr]-H_\ell(P)\;\geq\;0.
\]
Then \(R_f^\star(x)=H_\ell(P_f)\), and
\begin{equation}
R_{\mathcal S}(\hat a_{\mathcal S};x)
=
\underbrace{\mathbb E_F\bigl[H_\ell(P_F)\bigr]}_{\mathrm{AU}_{\mathcal S}(x)}
+
\underbrace{\mathbb E_F\bigl[d_\ell(P_F\Vert\bar P_{\mathrm{ref}})\bigr]}_{\text{generalized variance}}
+
\underbrace{d_\ell(\bar P_{\mathrm{ref}}\Vert\hat Q)}_{\text{deployment gap}},
\label{eq:score-form}
\end{equation}
where \(\bar P_{\mathrm{ref}}\) minimizes
\(Q\mapsto\mathbb E_F\bigl[d_\ell(P_F\Vert Q)\bigr]\).
\end{lemma}

\begin{proof}
Propriety gives
\(R_f^\star(x)=\inf_{Q}\mathbb E_{Y\sim P_f}[\ell(Q,Y)]=H_\ell(P_f)\). Because
\(P\mapsto\mathbb E_{Y\sim P}[\ell(Q,Y)]\) is linear,
\[
R_{\mathcal S}(\hat a_{\mathcal S};x)
=\mathbb E_{Y\sim\bar P_{\mathrm{ref}}}[\ell(\hat Q,Y)]
=H_\ell(\bar P_{\mathrm{ref}})+d_\ell(\bar P_{\mathrm{ref}}\Vert\hat Q),
\]
and, by the same linearity,
\[
H_\ell(\bar P_{\mathrm{ref}})
=\mathbb E_F\,\mathbb E_{Y\sim P_F}[\ell(\bar P_{\mathrm{ref}},Y)]
=\mathbb E_F\bigl[H_\ell(P_F)+d_\ell(P_F\Vert\bar P_{\mathrm{ref}})\bigr].
\]
Finally,
\(\mathbb E_F[d_\ell(P_F\Vert Q)]
=\mathbb E_{Y\sim\bar P_{\mathrm{ref}}}[\ell(Q,Y)]-\mathbb E_F[H_\ell(P_F)]\)
is minimized at \(Q=\bar P_{\mathrm{ref}}\) by propriety.
\end{proof}

The score divergence \(d_\ell\) is the generalized Bregman divergence generated
by the convex function \(-H_\ell\) and the subgradient selected by the scoring
rule \citep{gneiting2007strictly}; when \(-H_\ell\) is differentiable, it
reduces to an ordinary Bregman divergence. When observations and predictive
distributions admit representatives in a common Bregman domain \(\mathcal D\),
as for the CRPS, multiclass log loss and Brier score, identifying each
distribution with its representative gives \(d_\ell=D_\phi\), and
\Cref{lem:score-form} coincides term by term with \Cref{eq:bregman}. For the
continuous log score, \(d_\ell=\mathrm{KL}\) on densities, but the pointwise
loss has no such representation because \(\delta_Y\) has no Lebesgue density;
\Cref{lem:score-form} then writes AU as a generalized entropy rather than as an
expected divergence.

\paragraph{Structure of the following subsections.}
Each subsection has the same four parts: a setting card that fixes every
object entering \Cref{eq:bregman}, followed by the \emph{derivation}, the
resulting \emph{decomposition}, and the form of the \emph{aleatoric term}.
\Cref{tab:app-summary} collects the results.

\subsection{Squared loss}
\label{app:mse}

\losscard
{Point (value) scoring.}
{\(\mathcal Y=\mathbb R\).}
{\(\mathcal A=\mathbb R\); deployed point prediction \(\hat a:=\hat a_{\mathcal S}(x)\).}
{\(\ell(a,y)=(y-a)^2\).}
{\(\mathcal D=\mathbb R\); the outcome \(Y\) itself.}
{\(\phi(u)=u^2\); \ \(D_\phi(u\Vert v)=(u-v)^2\).}
{Conditional mean \(\mu_f(x):=\mathbb E_{Y\mid f}[Y]\).}
{Posterior mean \(m_{\mathrm{ref}}(x):=\mathbb E_F[\mu_F(x)]=\mathbb E[Y\mid x,\mathcal S]\).}

\paragraph{Derivation.}
Write \(\sigma_\varepsilon^2(x,f):=\operatorname{Var}_{Y\mid f}(Y)\) for the
conditional noise variance under \(f\), and
\(\sigma^2_{\mathrm{ref}}(x):=\operatorname{Var}_F\bigl[\mu_F(x)\bigr]\) for the
posterior variance of the conditional mean. For fixed \(f\),
\[
\mathbb E_{Y\mid f}\bigl[(Y-\hat a)^2\bigr]
=
\sigma_\varepsilon^2(x,f)+\bigl(\mu_f(x)-\hat a\bigr)^2,
\]
and averaging over \(F\),
\[
\mathbb E_F\bigl[(\mu_F(x)-\hat a)^2\bigr]
=
\sigma^2_{\mathrm{ref}}(x)+\bigl(m_{\mathrm{ref}}(x)-\hat a\bigr)^2 .
\]

\paragraph{Decomposition.}
\begin{equation}
R_{\mathcal S}(\hat a_{\mathcal S};x)
=
\underbrace{\mathbb E_F\bigl[\sigma_\varepsilon^2(x,F)\bigr]}_{\mathrm{AU}_{\mathcal S}(x)}
+
\underbrace{
\underbrace{\textcolor{bayes}{\sigma^2_{\mathrm{ref}}(x)}}_{\text{generalized variance}}
+
\underbrace{\textcolor{freq}{\bigl(\hat a_{\mathcal S}(x)-m_{\mathrm{ref}}(x)\bigr)^2}}_{\text{deployment gap}}
}_{\mathrm{EU}_{\mathcal S}(\hat a_{\mathcal S};x)} .
\label{eq:mse-decomp}
\end{equation}
The first two terms sum to the posterior predictive variance
\(\operatorname{Var}(Y\mid x,\mathcal S)\) by the law of total variance; the
deployment gap vanishes if and only if \(\hat a_{\mathcal S}(x)=m_{\mathrm{ref}}(x)\).

\paragraph{Aleatoric term.}
If the conditional variance does not depend on \(f\),
\(\sigma_\varepsilon^2(x,f)=\sigma_\varepsilon^2(x)\) (for instance under
additive noise \(Y=f(x)+\varepsilon\) with an \(f\)-independent noise law, where
\(\mu_f(x)=f(x)\)), then \(\mathrm{AU}_{\mathcal S}(x)=\sigma_\varepsilon^2(x)\),
independently of \(\mathcal S\). If instead \(\sigma_\varepsilon^2(x,f)\) varies
over the support of the reference posterior, AU is the posterior mean of the
noise variance and can depend on \(\mathcal S\).

\subsection{Continuous ranked probability score (CRPS)}
\label{app:crps}

\losscard
{Distributional scoring; strictly proper on distributions with finite first moment.}
{\(\mathcal Y=\mathbb R\).}
{\(\mathcal A\): CDFs on \(\mathbb R\) with finite first moment; deployed predictive CDF \(\hat G:=\hat a_{\mathcal S}(x)\).}
{\(\ell(G,y)=\operatorname{CRPS}(G,y)=\int_{\mathbb R}\bigl(G(t)-\mathbf 1\{y\leq t\}\bigr)^2\,dt\).}
{\(\mathcal D\): functions \(G:\mathbb R\to[0,1]\) with \(\int_{\mathbb R}\lvert G(t)-\mathbf 1\{t\geq 0\}\rvert\,dt<\infty\); the step function \(t\mapsto\mathbf 1\{Y\leq t\}\).}
{\(\phi(G)=\int_{\mathbb R}\bigl(G(t)^2-\mathbf 1\{t\geq 0\}\bigr)\,dt\); \ \(D_\phi(u\Vert v)=\int_{\mathbb R}\bigl(u(t)-v(t)\bigr)^2dt\). The subtracted term only makes \(\phi\) finite and cancels in \(D_\phi\).}
{Conditional CDF \(G_f(t):=P(Y\leq t\mid x,f)=\mathbb E_{Y\mid f}\bigl[\mathbf 1\{Y\leq t\}\bigr]\).}
{Posterior predictive CDF \(\bar G_{\mathrm{ref}}(t):=\mathbb E_F[G_F(t)]=P(Y\leq t\mid x,\mathcal S)\).}

\paragraph{Derivation.}
For fixed \(f\) and each \(t\),
\[
\mathbb E_{Y\mid f}\bigl[\mathbf 1\{Y\leq t\}\bigr]=G_f(t),
\qquad
\mathbb E_{Y\mid f}\bigl[(\mathbf 1\{Y\leq t\}-G_f(t))^2\bigr]=G_f(t)\bigl(1-G_f(t)\bigr),
\]
so the cross term vanishes pointwise in \(t\) and
\[
\mathbb E_{Y\mid f}\bigl[\operatorname{CRPS}(\hat G,Y)\bigr]
=
\int_{\mathbb R}G_f(t)\bigl(1-G_f(t)\bigr)\,dt
+
\int_{\mathbb R}\bigl(G_f(t)-\hat G(t)\bigr)^2dt .
\]
Averaging over \(F\) and expanding around \(\bar G_{\mathrm{ref}}\), the cross
term again vanishes pointwise in \(t\) because
\(\mathbb E_F[G_F(t)]=\bar G_{\mathrm{ref}}(t)\). Both steps are instances of
\eqref{eq:compensation} for the squared \(L^2\) distance, which is well defined
on \(\mathcal D\) because differences of its elements lie in
\(L^2(\mathbb R)\), even though CDFs themselves do not.

\paragraph{Decomposition.}
\begin{equation}
\begin{aligned}
R_{\mathcal S}(\hat a_{\mathcal S};x)
={}&
\underbrace{\mathbb E_F\!\int_{\mathbb R}G_F(t)\bigl(1-G_F(t)\bigr)\,dt}_{\mathrm{AU}_{\mathcal S}(x)}
+
\underbrace{\textcolor{bayes}{\mathbb E_F\!\int_{\mathbb R}\bigl(G_F(t)-\bar G_{\mathrm{ref}}(t)\bigr)^2dt}}_{\text{generalized variance}}
\\
&+
\underbrace{\textcolor{freq}{\int_{\mathbb R}\bigl(\bar G_{\mathrm{ref}}(t)-\hat G(t)\bigr)^2dt}}_{\text{deployment gap}} .
\end{aligned}
\label{eq:crps-decomp}
\end{equation}

\paragraph{Aleatoric term.}
The conditional Bayes risk is
\(\int_{\mathbb R}G_f(1-G_f)\,dt=\tfrac12\,\mathbb E\lvert Y-Y'\rvert\) for
\(Y,Y'\stackrel{\mathrm{iid}}{\sim}P(\cdot\mid x,f)\), which is invariant under
shifts of \(P(\cdot\mid x,f)\). Under additive noise
\(Y=\mu_f(x)+\varepsilon\) whose law given \(x\) does not depend on \(f\), AU
therefore equals \(\tfrac12\,\mathbb E\lvert\varepsilon-\varepsilon'\rvert\),
independently of \(\mathcal S\); for Gaussian noise,
\(\mathrm{AU}_{\mathcal S}(x)=\sigma_\varepsilon(x)/\sqrt{\pi}\).

\paragraph{Computation.}
For CDFs \(G_1,G_2\) with independent draws \(X_i,X_i'\sim G_i\),
\(\int_{\mathbb R}(G_1-G_2)^2dt
=\mathbb E\lvert X_1-X_2\rvert-\tfrac12\mathbb E\lvert X_1-X_1'\rvert
-\tfrac12\mathbb E\lvert X_2-X_2'\rvert\). The three terms can therefore be
estimated by Monte Carlo from posterior draws
\(F_m\sim\Pi_{\mathrm{ref}}(\cdot\mid\mathcal S)\) with conditional samples
\(Y_m,Y_m'\sim P(\cdot\mid x,F_m)\), posterior-predictive samples generated from
independent posterior draws, and samples from \(\hat G\). When the conditional
CDFs \(G_{F_m}\) can be evaluated directly, the integrals can instead be
computed by numerical quadrature.

\subsection{Log score for continuous outcomes}
\label{app:nll}

\losscard
{Distributional scoring; strictly proper.}
{\(\mathcal Y=\mathbb R\).}
{\(\mathcal A\): densities with respect to Lebesgue measure \(\lambda\); deployed predictive density \(\hat q:=\hat a_{\mathcal S}(x)\), with \(\hat q>0\) almost everywhere under \(\bar p_{\mathrm{ref}}\).}
{\(\ell(q,y)=-\log q(y)\).}
{None: the representative \(\delta_Y\) has no density. The decomposition follows from \Cref{lem:score-form}.}
{On densities, \(\phi(p)=\int p\log p\,d\lambda\); \ \(D_\phi(p\Vert q)=\mathrm{KL}(p\Vert q)=d_\ell(p\Vert q)\). Generalized entropy: differential entropy \(H(p)=-\int p\log p\,d\lambda\).}
{Conditional density \(p_f(y):=p(y\mid x,f)\).}
{Posterior predictive density \(\bar p_{\mathrm{ref}}(y):=\mathbb E_F[p_F(y)]=p(y\mid x,\mathcal S)\).}

\paragraph{Derivation.}
For fixed \(f\), adding and subtracting \(\log p_f\) gives
\(\mathbb E_{Y\mid f}\bigl[-\log\hat q(Y)\bigr]=H(p_f)+\mathrm{KL}(p_f\Vert\hat q)\).
Averaging over \(F\) and using
\(\mathbb E_F\!\int p_F\log(\bar p_{\mathrm{ref}}/\hat q)\,d\lambda
=\int\bar p_{\mathrm{ref}}\log(\bar p_{\mathrm{ref}}/\hat q)\,d\lambda\),
\[
\mathbb E_F\bigl[\mathrm{KL}(p_F\Vert\hat q)\bigr]
=
\mathbb E_F\bigl[\mathrm{KL}(p_F\Vert\bar p_{\mathrm{ref}})\bigr]
+
\mathrm{KL}(\bar p_{\mathrm{ref}}\Vert\hat q),
\]
which is \eqref{eq:score-form} with \(H_\ell=H\) and \(d_\ell=\mathrm{KL}\).

\paragraph{Decomposition.}
\begin{equation}
R_{\mathcal S}(\hat a_{\mathcal S};x)
=
\underbrace{\mathbb E_F\bigl[H(p_F)\bigr]}_{\mathrm{AU}_{\mathcal S}(x)}
+
\underbrace{\textcolor{bayes}{\mathbb E_F\bigl[\mathrm{KL}(p_F\Vert\bar p_{\mathrm{ref}})\bigr]}}_{\text{generalized variance}}
+
\underbrace{\textcolor{freq}{\mathrm{KL}(\bar p_{\mathrm{ref}}\Vert\hat q)}}_{\text{deployment gap}},
\label{eq:nll-decomp}
\end{equation}
where the generalized variance equals the mutual information
\(I(Y;F\mid x,\mathcal S)=H(\bar p_{\mathrm{ref}})-\mathbb E_F[H(p_F)]\).

\paragraph{Aleatoric term.}
AU is an expected differential entropy. It may be negative and shifts by
\(\log c\) when \(y\) is rescaled by \(c>0\), whereas both epistemic terms are
Kullback--Leibler divergences, non-negative and invariant under invertible
measurable transformations of \(\mathcal Y\). Under additive Gaussian noise
whose variance does not depend on \(f\),
\(H(p_f)=\tfrac12\log\bigl(2\pi e\,\sigma_\varepsilon^2(x)\bigr)\) and AU is
independent of \(\mathcal S\).

\paragraph{Closed form for Gaussian reference posteriors.}
Suppose the observation model is Gaussian with \(f\)-independent noise,
\(Y=\mu_F(x)+\varepsilon\) with
\(\varepsilon\sim\mathcal N\bigl(0,\sigma_\varepsilon^2(x)\bigr)\) independent
of \(F\), and the reference posterior of the conditional mean is Gaussian,
\(\mu_F(x)\sim\mathcal N\bigl(m_{\mathrm{ref}}(x),\sigma^2_{\mathrm{ref}}(x)\bigr)\)
(as for a GP reference prior). Then
\(\bar p_{\mathrm{ref}}=\mathcal N(m_{\mathrm{ref}},v_{\mathrm{ref}})\) with
\(v_{\mathrm{ref}}:=\sigma_\varepsilon^2+\sigma^2_{\mathrm{ref}}\). For a
Gaussian deployed predictive \(\hat q=\mathcal N(\hat\mu,\hat s^2)\),
\[
\underbrace{\tfrac12\log\Bigl(1+\frac{\sigma^2_{\mathrm{ref}}}{\sigma_\varepsilon^2}\Bigr)}_{\text{generalized variance}},
\qquad
\underbrace{\tfrac12\Bigl[\log\frac{\hat s^2}{v_{\mathrm{ref}}}
+\frac{v_{\mathrm{ref}}+(\hat\mu-m_{\mathrm{ref}})^2}{\hat s^2}-1\Bigr]}_{\text{deployment gap}} .
\]
For non-Gaussian reference posteriors, \(\bar p_{\mathrm{ref}}\) is a
non-Gaussian mixture. Terms involving \(\log\bar p_{\mathrm{ref}}\) then require
either an evaluable mixture density or the Monte Carlo estimate
\(\hat p(y)=M^{-1}\sum_{m=1}^M p_{F_m}(y)\) with
\(F_m\sim\Pi_{\mathrm{ref}}(\cdot\mid\mathcal S)\); by Jensen's inequality,
\(\log\hat p\) is biased downward for finite \(M\), but it is consistent as
\(M\to\infty\).

\subsection{Multiclass log loss}
\label{app:ce}

\losscard
{Distributional scoring; strictly proper.}
{\(\mathcal Y=\{1,\dots,K\}\); binary classification is \(K=2\).}
{\(\mathcal A=\Delta^{K-1}\), the probability simplex; deployed class-probability vector \(\hat\pi:=\hat a_{\mathcal S}(x)\) in its relative interior.}
{\(\ell(\pi,y)=-\log\pi_y\).}
{\(\mathcal D=\Delta^{K-1}\); the one-hot vector \(e_Y\).}
{\(\phi(u)=\sum_k u_k\log u_k\) with \(0\log 0:=0\); \ \(D_\phi(u\Vert v)=\sum_k u_k\log(u_k/v_k)=\mathrm{KL}(u\Vert v)\), so that \(D_\phi(e_Y\Vert\pi)=-\log\pi_Y\).}
{Conditional class probabilities \(\pi_f(x):=\mathbb E_{Y\mid f}[e_Y]=\bigl(P(Y=k\mid x,f)\bigr)_{k=1}^K\).}
{Posterior predictive probabilities \(\bar\pi_{\mathrm{ref}}(x):=\mathbb E_F[\pi_F(x)]=\bigl(P(Y=k\mid x,\mathcal S)\bigr)_{k=1}^K\).}

\paragraph{Derivation.}
\Cref{eq:bregman} applies directly. The conditional Bayes risk is the
Shannon entropy of the conditional class probabilities,
\[
\mathbb E_{Y\mid f}\bigl[-\log\pi_{f,Y}(x)\bigr]=H\bigl(\pi_f(x)\bigr).
\]

\paragraph{Decomposition.}
\begin{equation}
R_{\mathcal S}(\hat a_{\mathcal S};x)
=
\underbrace{\mathbb E_F\bigl[H(\pi_F(x))\bigr]}_{\mathrm{AU}_{\mathcal S}(x)}
+
\underbrace{\textcolor{bayes}{\mathbb E_F\bigl[\mathrm{KL}(\pi_F(x)\Vert\bar\pi_{\mathrm{ref}}(x))\bigr]}}_{\text{generalized variance}}
+
\underbrace{\textcolor{freq}{\mathrm{KL}(\bar\pi_{\mathrm{ref}}(x)\Vert\hat\pi)}}_{\text{deployment gap}},
\label{eq:ce-decomp}
\end{equation}
where the generalized variance equals
\(I(Y;F\mid x,\mathcal S)=H(\bar\pi_{\mathrm{ref}})-\mathbb E_F[H(\pi_F)]\).

\paragraph{Aleatoric term.}
Unlike in regression with an \(f\)-independent noise law, \(H(\pi_f(x))\) need
not be constant in \(f\). AU is then a genuine posterior average and can depend
on \(\mathcal S\). It cannot if \(H(\pi_f(x))\) is constant over the support of
the reference prior (\Cref{app:au-invariance}). For example, if the reference
prior supports only labelings that are deterministic at \(x\),
\(\pi_f(x)\in\{e_1,\dots,e_K\}\), then \(\mathrm{AU}_{\mathcal S}(x)=0\) for
every sample and all uncertainty at \(x\) is epistemic, however uncertain the
posterior is about which class is correct. By concavity
of \(H\), \(\mathbb E_F[H(\pi_F)]\leq H(\bar\pi_{\mathrm{ref}})\), and the gap is
exactly the generalized variance:
\(H(\bar\pi_{\mathrm{ref}})=\mathrm{AU}_{\mathcal S}(x)+I(Y;F\mid x,\mathcal S)\).
Evaluating at
\(\hat\pi=\bar\pi_{\mathrm{ref}}\) removes the deployment gap and yields the
information-theoretic decomposition
\(H(\bar\pi_{\mathrm{ref}})=\mathbb E_F[H(\pi_F)]+I(Y;F\mid x,\mathcal S)\), with
the spread taken under the reference posterior rather than an estimator
posterior (cf.\ \Cref{sec:gp-oof}).

\subsection{Brier score}
\label{app:brier}

\losscard
{Distributional scoring; strictly proper.}
{\(\mathcal Y=\{1,\dots,K\}\).}
{\(\mathcal A=\Delta^{K-1}\); deployed class-probability vector \(\hat\pi:=\hat a_{\mathcal S}(x)\).}
{\(\ell(\pi,y)=\lVert\pi-e_y\rVert_2^2\).}
{\(\mathcal D=\Delta^{K-1}\); the one-hot vector \(e_Y\).}
{\(\phi(u)=\lVert u\rVert_2^2\); \ \(D_\phi(u\Vert v)=\lVert u-v\rVert_2^2\).}
{\(\pi_f(x)\), as in \Cref{app:ce}.}
{\(\bar\pi_{\mathrm{ref}}(x)\), as in \Cref{app:ce}.}

\paragraph{Derivation.}
\Cref{eq:bregman} applies directly. The conditional Bayes risk is
\(\mathbb E_{Y\mid f}\bigl[\lVert e_Y-\pi_f(x)\rVert_2^2\bigr]=1-\lVert\pi_f(x)\rVert_2^2\).

\paragraph{Decomposition.}
\begin{equation}
R_{\mathcal S}(\hat a_{\mathcal S};x)
=
\underbrace{\mathbb E_F\bigl[1-\lVert\pi_F(x)\rVert_2^2\bigr]}_{\mathrm{AU}_{\mathcal S}(x)}
+
\underbrace{\textcolor{bayes}{\mathbb E_F\bigl[\lVert\pi_F(x)-\bar\pi_{\mathrm{ref}}(x)\rVert_2^2\bigr]}}_{\text{generalized variance}}
+
\underbrace{\textcolor{freq}{\lVert\bar\pi_{\mathrm{ref}}(x)-\hat\pi\rVert_2^2}}_{\text{deployment gap}} .
\label{eq:brier-decomp}
\end{equation}

\paragraph{Aleatoric term.}
AU is the posterior expected Gini impurity of the conditional class
probabilities. As for the log loss, \(1-\lVert\pi_f(x)\rVert_2^2\) need not be
constant in \(f\), so AU can depend on \(\mathcal S\). It vanishes for every
sample if the reference prior supports only labelings that are deterministic
at \(x\).

\subsection{When is the aleatoric term independent of the reference posterior?}
\label{app:au-invariance}

By \Cref{eq:au-eu},
\(\mathrm{AU}_{\mathcal S}(x)=\mathbb E_F\bigl[R_F^\star(x)\bigr]\). If the
conditional Bayes risk \(R_f^\star(x)\) is constant over the support of the
reference prior, AU takes the same value for every observed sample. For the
three regression losses, this holds under additive noise whose law given \(x\)
does not depend on \(f\), since the variance, the CRPS entropy
\(\tfrac12\mathbb E\lvert Y-Y'\rvert\), and the differential entropy are all
invariant under shifts. The condition is sufficient, not necessary, and it is
not met in two situations: when the reference prior places uncertainty on the
noise law itself, and in classification, where \(R_f^\star(x)\) is a function
of \(\pi_f(x)\). In both situations AU is the posterior expectation of an
unknown irreducible risk and can depend on the observed sample, but it need
not. For instance, if the reference prior supports only deterministic
labelings, \(\pi_f(x)\in\{e_1,\dots,e_K\}\), then \(R_f^\star(x)=0\) for every
\(f\) under both the log loss and the Brier score, so AU vanishes for every
sample and all uncertainty at \(x\) is epistemic.

\begin{table}[t]
\centering
\footnotesize
\renewcommand{\arraystretch}{1.3}
\setlength{\tabcolsep}{4pt}
\caption{Instantiations of \Cref{eq:bregman}. EU is the sum of the last two
columns. For every distributional score the centroid is the posterior
predictive.}
\label{tab:app-summary}
\begin{tabularx}{\linewidth}{@{}l>{\raggedright\arraybackslash}p{1.3cm}XXX@{}}
\toprule
\textbf{Loss} & \textbf{Centroid} & \textbf{AU} & \textbf{\textcolor{bayes}{Generalized variance}} & \textbf{\textcolor{freq}{Deployment gap}} \\
\midrule
Squared
& \(m_{\mathrm{ref}}\)
& \(\mathbb E_F[\sigma_\varepsilon^2(x,F)]\)
& \(\sigma^2_{\mathrm{ref}}(x)\)
& \((\hat a-m_{\mathrm{ref}})^2\) \\
CRPS
& \(\bar G_{\mathrm{ref}}\)
& \(\mathbb E_F\!\int G_F(1-G_F)\,dt\)
& \(\mathbb E_F\!\int(G_F-\bar G_{\mathrm{ref}})^2dt\)
& \(\int(\bar G_{\mathrm{ref}}-\hat G)^2dt\) \\
Log score (cont.)
& \(\bar p_{\mathrm{ref}}\)
& \(\mathbb E_F[H(p_F)]\)
& \(I(Y;F\mid x,\mathcal S)\)
& \(\mathrm{KL}(\bar p_{\mathrm{ref}}\Vert\hat q)\) \\
Multiclass log
& \(\bar\pi_{\mathrm{ref}}\)
& \(\mathbb E_F[H(\pi_F)]\)
& \(I(Y;F\mid x,\mathcal S)\)
& \(\mathrm{KL}(\bar\pi_{\mathrm{ref}}\Vert\hat\pi)\) \\
Brier
& \(\bar\pi_{\mathrm{ref}}\)
& \(\mathbb E_F[1-\lVert\pi_F\rVert_2^2]\)
& \(\mathbb E_F\lVert\pi_F-\bar\pi_{\mathrm{ref}}\rVert_2^2\)
& \(\lVert\bar\pi_{\mathrm{ref}}-\hat\pi\rVert_2^2\) \\
\bottomrule
\end{tabularx}
\end{table}
\newpage

\newcommand{\apttablecaption}[2]{%
  \refstepcounter{table}%
  \begin{center}
  \textbf{Table \thetable:} #1
  \label{#2}
  \end{center}
  \vspace{-0.7em}
}

\section{Semi-synthetic regression tasks and reference posteriors}
\label{app:data}
\raggedbottom

Computing the ground-truth uncertainty of Section~2 requires two objects that a
real dataset does not provide: the noise model $P_{y\mid x,f}$, which defines
aleatoric uncertainty, and the reference posterior $\Pi_{\mathrm{ref}}(\cdot\mid\mathcal S)$
over ground-truth functions, which is one of the central objects for epistemic uncertainty. We therefore
build \emph{semi-synthetic} tasks: real covariates, but the targets are
generated by a process we fully control. This appendix describes how these
tasks and their reference posteriors are produced.

\paragraph{The central idea.}
Every task is produced in three steps. We choose a prior $\Pi_0$ over functions,
draw one function $f^\star \sim \Pi_0$ and add noise to it to obtain training
targets, and then condition the \emph{same} prior $\Pi_0$ on the resulting
training sample $\mathcal S$. Because the prior used for conditioning is exactly
the prior that generated $f^\star$, the reference posterior
$\Pi_{\mathrm{ref}}(\cdot\mid\mathcal S)$ is correctly specified by construction:
seen from the learner's side, $f^\star$ is simply one draw from it. This
holds for \emph{any} choice of prior hyperparameters. The hyperparameters
therefore only decide how realistic and how varied the tasks are. What \emph{can} make it wrong is an
implementation error or a sampler that has not converged, and
Appendix~\ref{app:data:checks} describes how we test for both.

\begin{center}
\fbox{\begin{minipage}{0.95\linewidth}
\textbf{Overview of the generation pipeline.}\\[2pt]
\emph{Once per dataset} (Appendices~\ref{app:data:covariates}--\ref{app:data:sample}):
\begin{enumerate}\setlength\itemsep{0pt}
  \item Load the real covariates, scale them to $[0,1]$, and split the rows into
        an evaluation pool and a pool of candidate training inputs.
  \item For each prior family (GP, HBART, BNN), fix one scale constant so that
        prior draws vary by about one unit over the evaluation pool.
  \item For each of the four noise modes, fix the noise variance function $\sigma_\varepsilon^2(x)$.
  \item Fix the training-set size $n$.
\end{enumerate}
\emph{Once per task instance}, i.e.\ per (dataset, prior family, noise mode,
coverage scenario, seed) (Appendix~\ref{app:data:instance}):
\begin{enumerate}\setlength\itemsep{0pt}\setcounter{enumi}{4}
  \item Select $n$ training inputs according to the coverage scenario.
  \item Draw the ground truth $f^\star \sim \Pi_0$.
  \item Draw noisy targets $y_i = f^\star(x_i) + \varepsilon_i$ at the training inputs.
  \item Condition $\Pi_0$ on $\mathcal S$ and store $D = 1000$ draws from
        $\Pi_{\mathrm{ref}}(\cdot\mid\mathcal S)$ at the evaluation points.
  \item Compute the oracle targets from the stored draws (Appendix~\ref{app:data:targets}).
\end{enumerate}
\end{minipage}}
\end{center}


\subsection{Covariates}
\label{app:data:covariates}

We use 13 tabular regression datasets from UCI and OpenML
(Table~\ref{tab:data:datasets}), with between 4 and 21 covariates and between
908 and 53\,940 rows. Numeric columns are kept as they are; categorical columns
with exactly two levels are coded as $0/1$; categorical columns with more levels
are dropped, because an integer code would impose an ordering that the
covariates do not have. Rows with missing values are removed. Each covariate is
then min-max scaled to $[0,1]$.

The original targets of these datasets are \emph{not} used as targets.

Each dataset is split once, with a fixed seed, into two disjoint parts. The
\emph{evaluation pool} $X_{\mathrm{eval}}$ consists of
$n_{\mathrm{eval}} = \min(1500, \lfloor N/2 \rfloor)$ randomly chosen rows, where
$N$ is the number of rows. All uncertainty targets and all method predictions
are evaluated on this pool. The remaining rows form the \emph{candidate pool},
from which training inputs are drawn. No evaluation point is ever a training
input.

\subsection{Priors over ground-truth functions}
\label{app:data:priors}

We use three prior families $\Pi_0$. They produce different kinds of functions:
functions with correlation structure induced by varying kernels (GP), piecewise
constant functions with axis-aligned jumps (HBART), and functions represented by
a neural network (BNN). Using three families ensures that no method is evaluated
only on functions that happen to match its own inductive bias. Table~\ref{tab:data:priors} summarises all settings.

\paragraph{A common scale.}
Left at their default settings, the three priors produce functions of very
different size: one prior's draws might vary by $\pm 0.3$, another's by $\pm 5$.
To make them comparable, we multiply every prior by a constant $a$, chosen per
dataset and prior family so that a typical draw varies by about one unit across
the evaluation pool. Concretely, we take 100 draws from the prior, compute each
draw's standard deviation over the evaluation pool, and set $a$ so that the median
of these standard deviations is one. All tasks then share the same signal scale,
so a given noise level (Appendix~\ref{app:data:noise}) makes tasks equally noisy
whichever prior generated them. The constant $a$ is fixed once, before any task is
generated, and simply becomes part of the prior. Individual draws are never rescaled.

\paragraph{Gaussian process (GP).}
$f \sim \mathcal{GP}\big(\mu,\, a^2 k\big)$. To introduce variation in the GP tasks we set a combination of kernels individually for each of the datasets.
The resulting kernels differ considerably between datasets
(Table~\ref{tab:data:datasets}): stationary Matérn, RBF, rational-quadratic and
powered-exponential kernels, sums with a linear or periodic component, and
arc-cosine kernels. The mean $\mu$ is zero. The kernel's output variance is the scale
constant $a^2$ (Step 2).

\paragraph{Heteroscedastic Bayesian additive regression trees (HBART).}
A sum of $m = 200$ regression trees
with the standard BART prior~\citep{Chipman_2010}: a node at depth $k$ is
split with probability $0.95\,(1+k)^{-2}$, up to a maximum depth of 6. The split
covariate is chosen uniformly at random and the split value uniformly from a
grid of 255 equally spaced thresholds in $[0,1]$. The leaf values are
independent $\mathcal N(0, a^2/m)$, so that the sum of $m$ leaves has prior
standard deviation $a$. The
``heteroscedastic'' part of the name refers to the likelihood: when conditioning,
each observation is weighted by its own, known noise variance
$\sigma_\varepsilon^2(x_i)$ (Appendix~\ref{app:data:instance}).

\paragraph{Bayesian neural network (BNN).}
A fully connected network with one hidden layer of $H = 50$ $\tanh$ units,
\begin{align*}
  u(x) &= \frac{1}{\sqrt{H}} \sum_{h=1}^{H} v_h \tanh\!\Big(\tfrac{1}{\sqrt d}\, w_h^\top (2x - 1) + c_h\Big), \\
  f(x) &= b + a\Big(u(x) - \tfrac{1}{R}\textstyle\sum_{r=1}^{R} u(x_r^{\mathrm{ref}})\Big),
\end{align*}
with all weights and biases $w_h, c_h, v_h, b$ independent standard normal. The
$1/\sqrt{\text{fan-in}}$ factors are the usual scaling under which the network
prior approaches a GP as the width grows. The second
term is centred on a fixed set of $R = 65$ reference inputs drawn uniformly from
$[0,1]^d$. Without this centring, most of the prior variance of $f$ would be a
constant offset, and the scale $a$ would control the overall level of $f$
rather than how much it varies across inputs. With it, $a$ controls the
variation and the bias $b$ controls the level. The same reference inputs are used
when drawing $f^\star$ and when conditioning. In practice this prior produces smooth
functions over the data.

\begin{figure}[t]
  \centering
  \includegraphics[width=\linewidth]{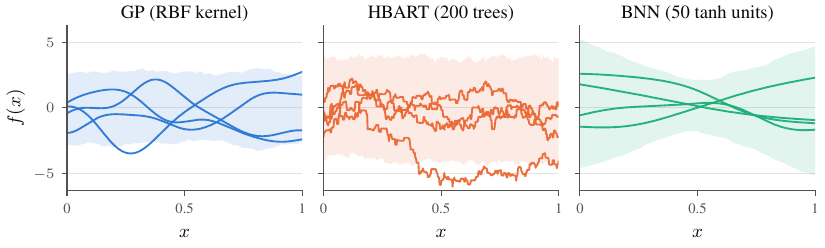}
  \caption{\textbf{Prior draws from the three prior families} on a one-dimensional
  input. Lines show four independent draws $f \sim \Pi_0$; shaded regions show the
  pointwise central 95\% interval over 500 draws. Each prior is scaled with the
  benchmark's rule, so that a typical draw has standard deviation one across the
  inputs. The GP panel uses an RBF kernel for illustration; in the benchmark the
  GP kernel is set separately for each dataset. HBART produces piecewise-constant
  functions with a large per-draw offset; the BNN prior produces smooth functions.}
  \label{fig:prior_draws_1d}
\end{figure}

\subsection{Aleatoric noise}
\label{app:data:noise}

Observations follow a additive Gaussian noise model:
\[
  y = f(x) + \varepsilon, \qquad \varepsilon \sim \mathcal N\big(0, \sigma_\varepsilon^2(x)\big).
\]
The noise variance $\sigma_\varepsilon^2(x)$ is not drawn at random: it is a fixed,
known function of $x$. For a given dataset and noise mode it is identical for all
three prior families, so tasks from different priors differ only in the function
$f$. We write it as
\[
  \log \sigma_\varepsilon^2(x) = c + s\, h(x),
\]
where $h$ is a \emph{shape} function standardised to mean zero and standard
deviation one over the evaluation pool, $s$ controls how strongly the noise
varies, and $c$ sets the overall noise level.

\paragraph{Four noise modes.}
All shapes are functions of a single score $z(x)$, the first principal component
of the column-standardised covariates, itself standardised to mean zero and unit
variance. This gives a direction along which the noise changes that is defined
for any number of covariates, and it makes every shape a closed-form function
of $x$. Before standardisation, the shapes are
\begin{center}
\begin{tabular}{@{}ll@{}}
  \toprule
  Mode & Shape before standardisation \\
  \midrule
  homoscedastic   & $h(x) = 0$ (constant noise) \\
  smooth monotone & $1/\big(1 + e^{-2 z(x)}\big)$ \\
  multimodal      & $\exp\!\big(-\tfrac{(z(x)+1.1)^2}{2\cdot 0.55^2}\big) + \exp\!\big(-\tfrac{(z(x)-1.1)^2}{2\cdot 0.55^2}\big)$ \\
  discontinuous   & $\mathbf 1\{z(x) > 0\}$ \\
  \bottomrule
\end{tabular}
\end{center}
The homoscedastic mode is the baseline. The smooth monotone mode lets the noise
increase steadily along one direction. In the multimodal mode there are two
separate high-noise regions, so the noise is not monotone in any single
direction. The discontinuous mode has a jump in noise level across a hyperplane,
which no smooth noise model can represent exactly.

\begin{figure}[t]
  \centering
  \includegraphics[width=\linewidth]{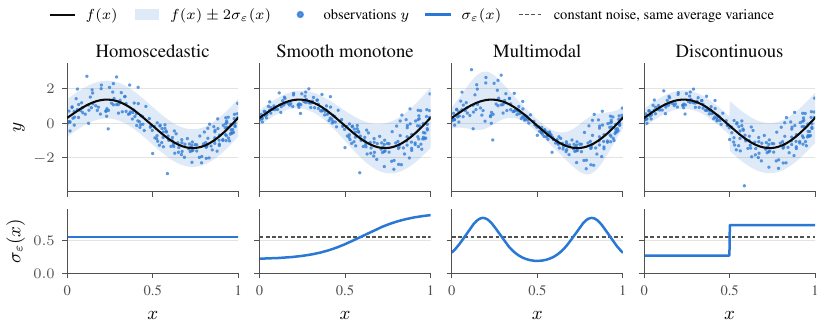}
  \caption{\textbf{The four aleatoric noise modes} on a one-dimensional input.
  Top: a fixed function $f$, the band $f \pm 2\sigma_\varepsilon(x)$ and noisy
  observations, generated from the same underlying random draws in every panel.
  Bottom: the noise standard deviation $\sigma_\varepsilon(x)$; the dashed line is
  the constant noise level with the same average variance. All modes have the same
  average noise variance (here a noise-to-signal ratio of $\rho = 0.3$) and, apart from
  the homoscedastic mode, the same spread of $\log\sigma_\varepsilon^2$. They differ
  only in where the noise is high: increasing along one direction, in two separate
  regions, or with a jump across a hyperplane.}
  \label{fig:noise_modes_1d}
\end{figure}

\paragraph{Strength and level.}
For the three heteroscedastic modes we set $s = 1$, so the log noise variance has
standard deviation one over the evaluation pool; for the homoscedastic mode
$s = 0$. The level $c$ is set so that the average noise variance over the
evaluation pool equals a target noise-to-signal ratio $\rho$:
\[
  \frac{1}{n_{\mathrm{eval}}} \sum_{x \in X_{\mathrm{eval}}} \sigma_\varepsilon^2(x) = \rho
  \quad\Longleftrightarrow\quad
  c = \log\rho - \log\Big(\tfrac{1}{n_{\mathrm{eval}}}\textstyle\sum_{x\in X_{\mathrm{eval}}} e^{s h(x)}\Big).
\]
Since prior draws have variance about one (Appendix~\ref{app:data:priors}), $\rho$
is the ratio of noise variance to signal variance. Each dataset has its own
$\rho$, drawn once, log-uniformly from $[0.2, 0.5]$ (Table~\ref{tab:data:datasets}). The noise therefore accounts for between
roughly $17\%$ and $30\%$ of the total variance of $y$, depending on the dataset.
As an example, on White Wine Quality the noise standard deviation is constant at
$0.47$ in the homoscedastic mode and ranges over $[0.17, 0.78]$ (smooth monotone),
$[0.11, 0.72]$ (multimodal) and $[0.22, 0.60]$ (discontinuous) in the other modes. The noise modes are illustrated in \Cref{fig:noise_modes_1d}

\subsection{Training sample and data coverage}
\label{app:data:sample}

\paragraph{Training-set size.}
Each dataset has a fixed training-set size $n$ (Table~\ref{tab:data:datasets}).
It is drawn once as an integer in $[1500, 5000]$,
limited to at most $60\%$ of the rows, and reduced further where necessary so
that both coverage scenarios below can be realised with the same $n$. The two
scenarios thus differ only in \emph{where} the training inputs lie, not in how
many there are. The resulting sizes range from 336 to 4\,771.

\begin{figure}[t]
  \centering
  \includegraphics[width=\linewidth]{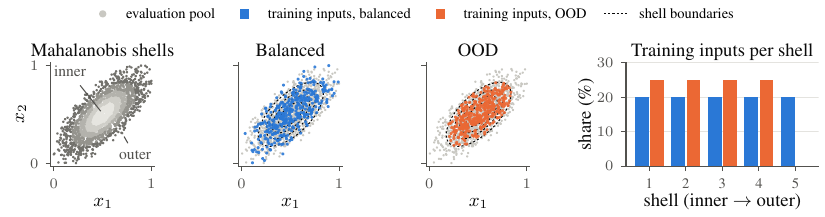}
  \caption{\textbf{The two data-coverage scenarios}, illustrated on a synthetic
  two-dimensional covariate cloud. Left: all rows are split into five shells at the
  quintiles of their Mahalanobis distance to the covariate mean (light: inner, dark:
  outer). Middle: the evaluation pool (grey) with the training inputs of each
  scenario; dashed lines are the shell boundaries. In the balanced scenario the
  training inputs cover all shells equally; in the OOD scenario none come from the
  outermost shell, so about 20\% of the evaluation points lie in a region without
  training data. Right: realised share of the training inputs per shell. Both
  scenarios use the same number of training inputs.}
  \label{fig:coverage_scenarios}
\end{figure}

\paragraph{Coverage scenarios.}
To control where training data is available, we sort all rows of the dataset by
their Mahalanobis distance to the mean of the covariates (using the empirical
covariance, shrunk by $10\%$ towards a scaled identity for numerical
stability) and split them at the quintiles into five \emph{shells}, from the
innermost to the outermost fifth. Training inputs are then drawn uniformly
without replacement from the candidate pool, with a prescribed share from each
shell:
\begin{itemize}\setlength\itemsep{0pt}
  \item \textbf{Balanced:} $20\%$ from every shell. The training inputs cover the
        input space in the same way as the evaluation pool; there is no
        covariate shift.
  \item \textbf{OOD:} $25\%$ from each of the four inner shells and none from the
        outermost shell. The roughly $20\%$ of evaluation points that lie in the
        outermost shell have no training data nearby, so they require
        extrapolation.
\end{itemize}
In the OOD scenario, epistemic uncertainty should rise in the empty outer shell,
while the aleatoric target, which is a fixed function of $x$, is unaffected by
where the training data lies. The coverage scenarios are illustrated in \Cref{fig:coverage_scenarios}.

\subsection{Generating one task instance}
\label{app:data:instance}

A task instance is defined by a dataset, a prior family, a noise mode, a
coverage scenario and a seed. It is generated as follows.
\begin{enumerate}
  \item \textbf{Training inputs.} Select $n$ training inputs
        $x_1,\dots,x_n$ from the candidate pool according to the coverage
        scenario (Appendix~\ref{app:data:sample}).
  \item \textbf{Ground truth.} Draw one function $f^\star \sim \Pi_0$, evaluated
        jointly at the evaluation pool and the training inputs.
  \item \textbf{Noisy targets.} Draw
        $y_i = f^\star(x_i) + \sigma_\varepsilon(x_i)\,\epsilon_i$ with
        $\epsilon_i \sim \mathcal N(0,1)$, and set
        $\mathcal S = \{(x_i, y_i)\}_{i=1}^n$. This is all a method (learner) sees.
  \item \textbf{Reference posterior.} Condition $\Pi_0$ on $\mathcal S$, with the
        noise variances $\sigma_\varepsilon^2(x_i)$ treated as known, and store
        $D = 1000$ draws $f^{(1)},\dots,f^{(D)} \sim \Pi_{\mathrm{ref}}(\cdot\mid\mathcal S)$
        evaluated at every point of the evaluation pool (and at the training
        inputs, for the checks in Appendix~\ref{app:data:checks}).
\end{enumerate}
The reference posterior is a posterior over the function $f$ only. The noise is
known to it, just as the noise model is known in Section~2. This is what makes
the aleatoric target exact.

Each instance draws its own random numbers, derived from the seed and from the
instance's dataset, prior family, noise mode and scenario. Instances are
therefore independent replicates: two noise modes or two scenarios with the same
seed share neither the ground truth nor the training sample.

\paragraph{Conditioning.}
The three prior families require different conditioning procedures.
\begin{itemize}
  \item \textbf{GP: exact.} With known noise variances the likelihood is Gaussian,
        and the posterior is available in closed form~\citep{GP_Rassmussen}.
        With $K$ the prior covariance of the training inputs,
        $\Sigma = \operatorname{diag}\big(\sigma_\varepsilon^2(x_1),\dots,\sigma_\varepsilon^2(x_n)\big)$
        and $k_x$ the prior covariances between $x$ and the training inputs,
        \begin{align*}
          m_{\mathcal S}(x) &= \mu + k_x^\top (K + \Sigma)^{-1}(y - \mu\mathbf 1), \\
          \operatorname{Cov}\big(f(x), f(x') \mid \mathcal S\big) &= a^2 k(x,x') - k_x^\top (K + \Sigma)^{-1} k_{x'}.
        \end{align*}
        The only difference from textbook GP regression is that each observation
        has its own noise variance, so noisy observations automatically count
        less.
  \item \textbf{HBART: MCMC.} We use the standard BART
        sampler~\citep{Chipman_2010} as implemented in
        \texttt{bartz}~\citep{petrillo2025fastbayesianadditiveregression}, with each observation's known
        noise variance in the likelihood. We run 250 independent chains for
        5\,000 warm-up iterations. To help the chains move between very
        different tree configurations, the first $75\%$ of the warm-up uses a
        tempered likelihood whose temperature decreases geometrically from 3 to
        1 in ten stages; the rest of the warm-up and all retained draws use the
        untempered likelihood. Each chain then contributes 4 draws, spaced 25
        iterations apart, giving $D = 1000$.
  \item \textbf{BNN: MCMC.} We sample all network weights with the No-U-Turn
        sampler~\citep{hoffman2011nouturnsampleradaptivelysetting} in NumPyro~\citep{phan2019composableeffectsflexibleaccelerated}: 4
        chains, each with 500 warm-up iterations and 250 retained draws
        ($D = 1000$), target acceptance probability 0.9.
        Each retained weight vector is turned into a function draw by
        evaluating the network at the evaluation pool.
\end{itemize}
The code path that draws $f^\star$ and the code path that conditions use the same
prior definition for each family (the same kernel function, tree prior and
network), so that the generating and the conditioning prior cannot drift apart.

\paragraph{Size of the benchmark.}
The full grid has $13$ datasets $\times$ $3$ prior families $\times$ $4$ noise
modes $\times$ $2$ coverage scenarios $= 312$ conditions. For each condition we
generate 10 seeds, which gives
3\,120 task instances with reference posteriors.

\subsection{Oracle targets}
\label{app:data:targets}

Under squared loss the posterior-risk decomposition of Appendix~\ref{app:mse}
reads, at an evaluation point $x$,
\[
  R_{\mathcal S}(\hat a_{\mathcal S}; x)
  = \underbrace{\sigma_\varepsilon^2(x)}_{\mathrm{AU}_{\mathcal S}(x)}
  + \underbrace{\sigma_f^2(x) + \big(\hat a_{\mathcal S}(x) - m_{\mathcal S}(x)\big)^2}_{\mathrm{EU}_{\mathcal S}(\hat a_{\mathcal S};\, x)}.
\]
Every term can be read off the stored artefacts:
\begin{itemize}
  \item \textbf{Aleatoric uncertainty} $\mathrm{AU}_{\mathcal S}(x) = \sigma_\varepsilon^2(x)$
        is known exactly. Because the noise does not depend on $f$, the
        posterior expectation of the Bayes risk is simply the noise variance.
  \item \textbf{Posterior mean} (the reference Bayes act $\bar a_{\mathrm{ref}}$):
        $m_{\mathcal S}(x) \approx \frac1D \sum_{d=1}^D f^{(d)}(x)$.
  \item \textbf{Posterior variance} (the generalised-variance term):
        $\sigma_f^2(x) \approx \frac1D \sum_{d=1}^D \big(f^{(d)}(x) - m_{\mathcal S}(x)\big)^2$.
        It does not depend on the method.
  \item \textbf{Epistemic uncertainty of a method}: given a method's prediction
        $\hat a_{\mathcal S}(x)$, $\mathrm{EU}_{\mathcal S}(\hat a_{\mathcal S}; x)
        = \sigma_f^2(x) + \big(\hat a_{\mathcal S}(x) - m_{\mathcal S}(x)\big)^2$,
        where the second term is the deployment gap. This target is computed
        separately for every method after it has been trained on $\mathcal S$.
\end{itemize}
Since the realised ground truth $f^\star$ is stored as well, the fixed-truth
excess risk $\big(\hat a_{\mathcal S}(x) - f^\star(x)\big)^2$ of Section~2.3 can
be computed on the same instances for comparison. Because the full set of
posterior draws is stored rather than only its mean and variance, the targets for
other losses (CRPS and log score, Appendices~\ref{app:crps} and \ref{app:nll}) can also be computed
without regenerating any data. With $D = 1000$ draws, the Monte Carlo error of
$\sigma_f^2(x)$ is about $5\%$ of its value at a single point, and this error
averages out over the points and instances that any reported score aggregates.
For the GP, $m_{\mathcal S}$ and $\sigma_f^2$ are also available in closed form.

\subsection{Correctness and convergence checks}
\label{app:data:checks}

We assess the calibration and convergence of the reference posteriors on all $3{,}120$
benchmark instances (Table~\ref{tab:data:checks}). 

\emph{Calibration.} If
$f^\star \sim \Pi_0$ and the draws follow the posterior, the rank of $f^\star(x)$ among
the $D$ draws is uniform \citep{talts2020validatingbayesianinferencealgorithms}. For all three prior families the
pooled rank histograms are flat. The coverage of the central $90\%$ interval is $0.90$ both
pooled (with confidence intervals that resample whole instances, since points within an
instance are dependent) and within the 104 combinations of dataset, noise mode and
coverage scenario (interquartile range $0.89$--$0.91$). Because these checks could also be
passed by draws that ignore the data, we note that the posterior mean tracks the realised
ground truth closely (median correlation with $f^\star$ between $0.94$ and $1.00$).
\emph{Convergence.} The GP posterior is exact. For the BNN, an instance is used only if
the split-$\hat R$ \citep{Vehtari_2021} of its four chains, computed in function
space at every evaluation point, is at most $1.05$ everywhere. Seeds that fail are
replaced by the next seed, for all prior families alike ($5.3\%$ of seeds with balanced
and $8.6\%$ with OOD coverage). In the retained instances, $98\%$ of point-wise $\hat R$
values are below $1.01$, and restricting the BNN tasks to instances with $\hat R \le 1.01$
everywhere leaves the method ranking unchanged (Kendall's $\tau = 0.98$; no method moves
by more than one rank). HBART pools 250 independent chains after a tempered warm-up. On
two of the largest datasets (Protein and kin8nm), a four times longer warm-up changes the
$90\%$ coverage by less than $0.003$ and the posterior standard deviation by at most about
$5\%$. \emph{Precision.}
The Monte Carlo error of the posterior variance, estimated by batch means over chains so
that within-chain dependence is accounted for, has a median of $4$--$6\%$ and a 95th
percentile below $9\%$.

\begin{table}[h]
\centering
\small
\caption{Reference-posterior diagnostics over the benchmark instances (1{,}040 per prior
family). Rank histogram: largest deviation of a bin share from $5\%$ (20 bins).
Coverage: share of evaluation points whose ground truth lies in the central $90\%$
posterior interval; pooled with a $95\%$ bootstrap interval over instances, and the
range over the 104 dataset $\times$ noise mode $\times$ coverage cells. MCSE: relative Monte
Carlo standard error of the posterior variance by batch means (20 batches of whole chains
or chain segments), over all evaluation points.}
\label{tab:data:checks}
\begin{tabular}{l ccc}
\toprule
 & GP & HBART & BNN \\
\midrule
Conditioning & exact & 250 chains & NUTS, 4 chains \\
$90\%$ coverage, pooled & $0.899$ {\scriptsize$[0.898, 0.901]$} & $0.902$ {\scriptsize$[0.901, 0.903]$} & $0.901$ {\scriptsize$[0.898, 0.905]$} \\
$90\%$ coverage, range over cells & $0.880$--$0.932$ & $0.881$--$0.921$ & $0.838$--$0.957$ \\
MCSE of $\sigma_f^2$, median / 95th pct. & $4.4\%$ / $5.7\%$ & $5.6\%$ / $8.6\%$ & $4.7\%$ / $6.1\%$ \\
Point-wise $\hat R$, median / 99th pct. & -- & -- & $1.000$ / $1.013$ \\
Corr.\ $(m_{\mathcal S}, f^\star)$, median & $0.94$ & $0.98$ & $1.00$ \\
\bottomrule
\end{tabular}
\end{table}

\paragraph{Released artefacts.}
For every dataset we release the processed covariates, the evaluation/candidate
split, the scale constant of each prior family and the
noise parameters of each mode. For every task instance we release the training
indices and targets $y$, the noise variances at the training and evaluation
points, the ground truth $f^\star$ at the training and evaluation points, the
$D \times n_{\mathrm{eval}}$ matrix of posterior draws, and the sampler
diagnostics.

\subsection{Summary tables}
\label{app:data:tables}

\begin{table}[H]
\centering
\caption{Datasets. $d$: number of covariates; $N$: rows after cleaning;
$n_{\mathrm{eval}}$: size of the evaluation pool; $n$: training-set size (the
same in both coverage scenarios); $\rho$: noise-to-signal ratio; GP kernel: the
kernel family fitted to the real targets and used by the GP prior
(Mat: Matérn; RQ: rational quadratic; PE: powered exponential; lin: linear).}
\label{tab:data:datasets}
\small
\setlength{\tabcolsep}{4pt}
\begin{tabular}{@{}llrrrrrl@{}}
  \toprule
  Dataset & Source & $d$ & $N$ & $n_{\mathrm{eval}}$ & $n$ & $\rho$ & GP kernel \\
  \midrule
  Abalone            & UCI 1        &  7 &  4\,177 & 1\,500 & 2\,084 & 0.28 & Arc-cosine \\
  Airfoil Self-Noise & UCI 291      &  5 &  1\,503 &    751 &    584 & 0.32 & Mat-$\tfrac12$ + lin \\
  Concrete Strength  & UCI 165      &  8 &  1\,030 &    515 &    348 & 0.23 & RQ + lin \\
  CPU Activity       & OpenML 44978 & 21 &  8\,192 & 1\,500 & 1\,782 & 0.25 & Mat-$\tfrac52$ + PE \\
  Diamonds           & OpenML 44979 &  6 & 53\,940 & 1\,500 & 2\,000 & 0.32 & RBF \\
  Health Insurance   & OpenML 44993 &  4 & 22\,272 & 1\,500 & 3\,279 & 0.42 & Arc-cosine \\
  kin8nm             & OpenML 44980 &  8 &  8\,192 & 1\,500 & 3\,136 & 0.33 & Mat-$\tfrac32$ \\
  Protein Structure  & OpenML 44963 &  9 & 45\,730 & 1\,500 & 3\,223 & 0.36 & RQ + lin \\
  QSAR Fish Toxicity & UCI 504      &  6 &     908 &    454 &    336 & 0.41 & Mat-$\tfrac32$ \\
  Sarcos             & OpenML 44976 & 21 & 48\,933 & 1\,500 & 4\,771 & 0.42 & Mat-$\tfrac52$ \\
  Seoul Bike Sharing & UCI 560      &  9 &  8\,760 & 1\,500 & 1\,945 & 0.21 & Periodic + lin \\
  Space GA           & OpenML 45402 &  6 &  3\,107 & 1\,500 & 1\,260 & 0.21 & PE \\
  White Wine Quality & OpenML 44971 & 11 &  4\,898 & 1\,500 & 2\,668 & 0.23 & RBF \\
  \bottomrule
\end{tabular}
\end{table}

\begin{table}[H]
\centering
\caption{References for the datasets in Table~\ref{tab:data:datasets}.}
\label{tab:data:references}
\small
\setlength{\tabcolsep}{6pt}
\begin{tabular}{@{}lp{0.65\linewidth}@{}}
  \toprule
  Dataset & Reference \\
  \midrule
  Abalone            & \cite{abalone_1} \\
  Airfoil Self-Noise & \citep{airfoil_self-noise_291} \\
  Concrete Strength  & \cite{concrete_compressive_strength_165} \\
  CPU Activity       & \citep{openml_cpu_activity_44978,delve_cpu_activity} \\
  Diamonds           & \cite{openml_diamonds_44979,ggplot2_diamonds} \\
  Health Insurance   & \cite{openml_health_insurance_44993,olson1998spousal_health_insurance} \\
  kin8nm             & \cite{openml_kin8nm_44980,delve_kin_family} \\
  Protein Structure  & \cite{openml_physiochemical_protein_44963,physicochemical_properties_of_protein_tertiary_structure_265} \\
  QSAR Fish Toxicity & \cite{qsar_fish_toxicity_504} \\
  Sarcos             & \cite{openml_sarcos_44976,vijayakumar2000locally_weighted_projection_regression} \\
  Seoul Bike Sharing & \cite{seoul_bike_sharing_demand_560} \\
  Space GA           & \cite{openml_space_ga_45402,pace1997quick_computation_spatial_autoregressive} \\
  White Wine Quality & \cite{openml_white_wine_44971,cortez2009modeling_wine_preferences} \\
  \bottomrule
\end{tabular}
\end{table}

\begin{table}[h]
\centering
\caption{Prior families and how their reference posteriors are computed. $a$ is
the per-dataset scale constant (Appendix~\ref{app:data:priors}). All samplers
return $D = 1000$ posterior draws of $f$ at the evaluation pool.}
\label{tab:data:priors}
\small
\begin{tabular}{@{}p{0.11\linewidth}p{0.42\linewidth}p{0.39\linewidth}@{}}
  \toprule
  Family & Prior over $f$ & Conditioning on $\mathcal S$ (noise known) \\
  \midrule
  GP    & $\mathcal{GP}(\mu, a^2 k)$; kernel family, lengthscales and constant
          mean fitted to the real targets by marginal likelihood
        & Exact closed form; $D$ joint draws from the Gaussian posterior \\
  \addlinespace
  HBART & Sum of 200 trees; split probability $0.95(1+k)^{-2}$ at depth $k$,
          maximum depth 6; uniform split covariate; 255 split thresholds per
          covariate; leaves $\mathcal N(0, a^2/200)$
        & BART MCMC (\texttt{bartz}); 250 chains; 5\,000 warm-up iterations,
          the first $75\%$ with likelihood temperature decreasing from 3 to 1;
          4 draws per chain, thinned by 25 \\
  \addlinespace
  BNN   & One hidden layer of 50 $\tanh$ units; standard normal weights and
          biases with $1/\sqrt{\text{fan-in}}$ scaling; output centred on 65
          fixed reference inputs and multiplied by $a$; output bias
          $\mathcal N(0,1)$
        & NUTS (NumPyro) on all weights; 4 chains $\times$ (500 warm-up $+$ 250
          draws); target acceptance 0.9; double precision \\
  \bottomrule
\end{tabular}
\end{table}

\newpage

\section{Models}
\label{app:benchmarked-models}
\begingroup\sloppy

This appendix specifies the 21 regression methods of the benchmark. All methods
expose the same prediction interface: a predictive mean \(\hat{y}(x)\), an
aleatoric uncertainty estimate, an epistemic uncertainty estimate, and a total
uncertainty estimate, all reported as variances. For every method the total
uncertainty is the sum of its aleatoric and epistemic estimates. Whenever a
method is trained in a standardized target space, predictive means and variances
are transformed back to the original target scale before evaluation.

\paragraph{Hyperparameter tuning.}
Hyperparameters were selected separately for every method and every benchmark
condition, i.e.\ every combination of dataset, noise mode, prior family and
coverage scenario (\(13\times4\times3\times2=312\) conditions). Each selection is
one Bayesian-optimization sweep of 30 trials in Weights \& Biases \citep{wandb}
over the search space listed with the method; entries with a single value are
fixed. A sweep uses the training sample of the condition's first instance (seed
0): a uniformly random 80\% of it trains a candidate configuration, and the
remaining 20\% scores it by the Gaussian negative log-likelihood of the targets
under \(\mathcal{N}\bigl(\hat{y}(x), \hat\sigma^2_{\mathrm{tot}}(x)\bigr)\), where
\(\hat\sigma^2_{\mathrm{tot}}\) is the method's total uncertainty. The winning
configuration is then used for all seeds of that condition. Tuning never sees the
evaluation points or any oracle quantity. Because the objective scores only the
total predictive variance, nothing in the tuning rewards a particular split into
aleatoric and epistemic parts. In total, the benchmark ran 6{,}552 sweeps.

\paragraph{Compute.}
All experiments were run on a GPU cluster. The methods implemented in JAX (GP,
VLGP, DKL, SNGP-Het and all neural-network methods) were fitted on
single NVIDIA GPUs (RTX~2080~Ti, V100 or A100); BLR and the tree-based methods
ran on CPU nodes. Summed over all jobs, the 65{,}457 final fits (3{,}117 per
method) took about 2{,}850 hours of wall-clock time (470 on GPUs, 2{,}380 on CPUs,
dominated by CatBoost-KGB with a median of 24 minutes per fit), and the tuning
sweeps took about 3{,}150 hours (2{,}050 on GPUs, 1{,}100 on CPUs). These are
sums of per-fit and per-sweep wall-clock times; the jobs ran in parallel.

\subsection{Probabilistic Regression Baselines}

\paragraph{Bayesian linear regression (BLR).}
The Bayesian linear regression baseline is a conjugate Gaussian linear model
with an optional intercept term. It places an isotropic Gaussian prior on the
regression weights,
\[
w \sim \mathcal{N}(0,\alpha^{-1}I),
\]
and assumes a homoscedastic Gaussian observation model with fixed noise
standard deviation \(\sigma\),
\[
y \mid x,w \sim \mathcal{N}(x^\top w,\sigma^2).
\]
The intercept, if used, is a constant column of the design matrix with the same
prior as the other weights. Targets are standardized before fitting, so
\(\sigma\) is measured in units of the target standard deviation and one search
range serves all datasets. Given the design matrix \(X\), the posterior over the
weights is available in closed form,
\[
S_N = \left(\alpha I + \beta X^\top X\right)^{-1},
\qquad
m_N = \beta S_N X^\top y,
\qquad
\beta=\sigma^{-2},
\]
up to a small diagonal jitter for numerical stability. The predictive
distribution for a test input \(x\) is then Gaussian with mean
\[
\mu(x)=x^\top m_N
\]
and variance
\[
\sigma^2_{\mathrm{pred}}(x)=\sigma^2 + x^\top S_N x.
\]
The quadratic term \(x^\top S_N x\) is interpreted as epistemic uncertainty, the
fixed noise variance \(\sigma^2\) as aleatoric uncertainty, and the total
predictive uncertainty as their sum \citep{mackay_1992_r7qgh-q6g10}.
\textbf{Hyperparameter search space:}
\(\alpha \in [10^{-4},10]\) (log-uniform), \(\sigma \in [10^{-2},2]\)
(log-uniform), and \texttt{fit\_intercept} \(\in\{\mathrm{True},\mathrm{False}\}\).

\paragraph{Exact Gaussian process (GP).}
Exact GP regression \citep{GP_Rassmussen} with a constant mean function and
the kernel
\[
k(x,x') = \sigma_m^2\, k_{\nu}\bigl(x,x';\ell_1,\dots,\ell_d\bigr) + \sigma_l^2\, x^\top x',
\]
the sum of a Matérn kernel \(k_\nu\) with one lengthscale per input dimension
(automatic relevance determination) and a linear kernel. The same kernel family
is used on every dataset, rather than the kernel that generated the data, so that
the baseline does not inherit knowledge of the oracle. The Gaussian likelihood has
a single, homoscedastic noise variance \(\sigma_n^2\). Inputs and targets are
standardized. The kernel hyperparameters, the constant mean and \(\sigma_n^2\) are
fitted jointly by maximizing the exact log marginal likelihood with AdamW
\citep{loshchilov2019decoupled}, a learning-rate schedule with linear warm-up and
cosine decay to zero, and gradient-norm clipping, implemented in GPJax
\citep{pinder2022gpjax}. The predictive mean is the posterior mean, epistemic
uncertainty is the posterior variance of the latent function, and aleatoric
uncertainty is the learned noise variance \(\sigma_n^2\), which is constant in
\(x\). \textbf{Hyperparameter search space:}
\texttt{matern\_nu} \(\in\{\tfrac12,\tfrac32,\tfrac52\}\),
\texttt{lengthscale} \(\in[0.1,10]\) (log-uniform; initial value of every ARD lengthscale),
\texttt{matern\_variance} \(\in[0.1,5]\) (log-uniform),
\texttt{linear\_variance} \(\in[10^{-3},2]\) (log-uniform),
\texttt{noise\_std} \(\in[0.02,1]\) (log-uniform; initial noise standard deviation),
\texttt{num\_iters} \(\in[150,500]\) (integer),
\texttt{warmup\_steps} \(\in[0,100]\) (integer),
\texttt{peak\_learning\_rate} \(\in[10^{-3},5\cdot10^{-2}]\) (log-uniform),
\texttt{gradient\_clip} \(\in[0.5,2]\) (uniform), and
\texttt{weight\_decay} \(\in[10^{-6},10^{-2}]\) (log-uniform).

\paragraph{Variational latent Gaussian process (VLGP).}
A heteroscedastic sparse variational GP with two latent processes
\citep{lazaro2011variational}: \(f\) for the regression function and
\(g(x)=\log\sigma^2(x)\) for the noise variance,
\[
y \mid f, g \sim \mathcal{N}\bigl(f(x), e^{g(x)}\bigr).
\]
The process \(f\) uses the Matérn-plus-linear kernel and constant mean of the
exact GP. The process \(g\) has its own Matérn kernel, whose linear component is
initialized at a negligible variance of \(10^{-6}\), and a constant prior mean
\(g_0\). Each process has \(M\) inducing inputs, placed at \(k\)-means centres of
the training inputs and shared by both, with a free Gaussian variational
distribution over the inducing values \citep{titsias2009variational,hensman2013gaussian}.
All parameters are fitted by maximizing the heteroscedastic evidence lower bound
of \citet{lazaro2011variational} (as implemented in GPJax) with AdamW, a warm-up
cosine schedule and gradient clipping. Two initialization choices keep the fit
stable. The variational covariances start at the prior covariance, so the
Kullback--Leibler term is zero at the start. The noise anchor \(g_0\), which is
also the initial variational mean of \(g\), is a nearest-neighbour estimate of the
noise level \citep{rice1984bandwidth},
\[
g_0 = \log\Bigl(\tfrac{1}{2n}\textstyle\sum_{i=1}^{n}\bigl(y_i-y_{\mathrm{nn}(i)}\bigr)^2\Bigr),
\]
where \(\mathrm{nn}(i)\) is the training point closest to \(x_i\). Differencing
neighbouring targets largely cancels the smooth signal, so unlike
\(\log\operatorname{Var}(y)\) this estimate is not dominated by the signal variance.
The predictive mean is the variational posterior mean of \(f\). Epistemic
uncertainty is the variational posterior variance of \(f(x)\), and aleatoric
uncertainty is the posterior mean of the noise variance,
\[
\hat\sigma^2_{\mathrm{ale}}(x)=\mathbb{E}_q\bigl[e^{g(x)}\bigr]=\exp\bigl(m_g(x)+\tfrac12 v_g(x)\bigr),
\]
where \(m_g(x)\) and \(v_g(x)\) are the variational mean and variance of \(g(x)\).
Among the GP baselines, VLGP is the only one whose aleatoric estimate varies with
\(x\). \textbf{Hyperparameter search space:}
\texttt{matern\_nu} \(\in\{\tfrac12,\tfrac32,\tfrac52\}\),
\texttt{lengthscale} \(\in[0.1,10]\) (log-uniform),
\texttt{matern\_variance} \(\in[0.1,5]\) (log-uniform),
\texttt{linear\_variance} \(\in[10^{-3},2]\) (log-uniform),
\texttt{noise\_matern\_nu} \(\in\{\tfrac32,\tfrac52\}\),
\texttt{noise\_lengthscale} \(\in[0.5,20]\) (log-uniform),
\texttt{noise\_kernel\_variance} \(\in[0.05,2]\) (log-uniform),
\texttt{n\_inducing} \(\in\{64,128,256\}\),
\texttt{num\_iters} \(\in[1000,4000]\) (integer),
\texttt{warmup\_steps} \(\in[0,200]\) (integer),
\texttt{peak\_learning\_rate} \(\in[2\cdot10^{-4},2\cdot10^{-3}]\) (log-uniform),
\texttt{gradient\_clip} \(\in[0.5,2]\) (uniform), and
\texttt{weight\_decay} \(\in[10^{-6},10^{-2}]\) (log-uniform).

\subsection{Deep Kernel Learning and Distance-Aware Models}

\paragraph{Deep kernel learning (DKL).}
DKL \citep{wilson2015deepkernellearning} combines a neural feature extractor with a
Gaussian-process output layer. A multilayer perceptron with \texttt{n\_layers}
hidden layers of width \texttt{hidden\_dim}, ReLU activations and a linear last
layer maps standardized inputs to features
\(\phi_\theta(x)\in\mathbb{R}^{d_\phi}\). An exact GP with zero mean and an ARD
Matérn-\(\tfrac52\) kernel is placed on these features, with a Gaussian likelihood
with one homoscedastic noise variance. Network weights, kernel hyperparameters
and noise are trained jointly by maximizing the exact log marginal likelihood
with AdamW, a warm-up cosine schedule and gradient clipping. The predictive mean is
the GP posterior mean at \(\phi_\theta(x)\), epistemic uncertainty is the posterior
latent variance, and aleatoric uncertainty is the learned noise variance, which is
constant in \(x\). Targets are standardized, and all reported means and variances
are transformed back to the original target scale.
\textbf{Hyperparameter search space:} \texttt{n\_layers} \(\in\{1,2,3\}\),
\texttt{hidden\_dim} \(\in\{16,32,64,128\}\),
\texttt{feature\_space\_dim} \(\in [2,8]\) (integer),
\texttt{noise\_std} \(\in [0.05,1.0]\) (log-uniform; initial value),
\texttt{base\_kernel\_variance} \(\in [0.1,5.0]\) (log-uniform; initial value),
\texttt{base\_kernel\_lengthscale} \(\in [0.1,5.0]\) (log-uniform; initial value),
\texttt{num\_iters} \(\in [150,500]\) (integer),
\texttt{warmup\_steps} \(\in [0,150]\) (integer),
\texttt{peak\_learning\_rate} \(\in [5\cdot 10^{-4},5\cdot 10^{-2}]\)
(log-uniform), \texttt{end\_learning\_rate}\(=0\),
\texttt{gradient\_clip} \(\in [0.5,2.0]\) (uniform),
\texttt{weight\_decay} \(\in [10^{-6},10^{-2}]\) (log-uniform), and
\texttt{standardize\_inputs}=\texttt{True},
\texttt{standardize\_targets}=\texttt{True}.

\paragraph{Heteroscedastic SNGP (SNGP-Het).}
SNGP \citep{liu2020sngp} uses the shared extractor \(h(x)\) with a random-feature
approximation of a GP output layer. With \(M\) random Fourier features
\(\varphi(h)=\sqrt{2/M}\,\cos(\Omega h+b)\), where the rows of \(\Omega\) are drawn
from \(\mathcal{N}(0,\ell^{-2}I)\) and \(b\) uniformly from \([0,2\pi]\) and both
stay fixed \citep{rahimi2007random}, the predictive mean is
\(\mu(x)=\varphi(h(x))^\top\beta+b_0\). For regression we add a linear
log-variance head on the same features, \(\log\sigma^2(x)=h(x)^\top w_g+b_g\),
initialized at \(w_g=0\) and \(b_g\) equal to the nearest-neighbour noise estimate
described for VLGP. Network and both heads are trained jointly by minimizing the
Gaussian negative log-likelihood plus the ridge penalty
\(\tfrac{\lambda}{2n}\lVert\beta\rVert^2\). After training, the Laplace posterior
over \(\beta\) is computed once, as in SNGP; for a Gaussian likelihood with fixed
\(\sigma^2(x)\) it is exact,
\[
\Sigma^{-1}=\lambda I+\sum_{i=1}^{n}\frac{\varphi_i\varphi_i^\top}{\sigma^2(x_i)},
\qquad \varphi_i=\varphi\bigl(h(x_i)\bigr).
\]
Epistemic uncertainty is \(\varphi(h(x))^\top\Sigma\,\varphi(h(x))\), and aleatoric
uncertainty is the point estimate \(\sigma^2(x)\). Heteroscedastic SNGP has been
published only for classification \citep{fortuin2022deep}; this regression version
is the standard mean--variance construction on SNGP's extractor. Its variance head
is not distance-aware and carries no uncertainty of its own, and the negative
log-likelihood is not reweighted (e.g.\ by the \(\beta\)-NLL of
\citealp{seitzer2022pitfalls}). \textbf{Hyperparameter search space:}
\texttt{width}, \texttt{depth}, \texttt{coeff}, \texttt{dropout},
\texttt{lengthscale\_factor}, \texttt{epochs}, \texttt{batch\_size},
\texttt{learning\_rate} and \texttt{weight\_decay} as for DUE;
\texttt{num\_inducing} (number of random features \(M\))
\(\in\{512,1024,2048\}\); and \texttt{ridge} (\(\lambda\)) \(\in[10^{-3},10]\)
(log-uniform).

\subsection{Tree-Based Models}

\paragraph{NGBoost with Normal-Inverse-Gamma outputs (NGBoost-NIG).}
This model uses NGBoost \citep{ngboost} for evidential regression. The boosting
model predicts the four parameters of a Normal-Inverse-Gamma (NIG) distribution,
\[
\mu(x),\qquad \lambda(x)=e^{\theta_2(x)},\qquad \alpha(x)=e^{\theta_3(x)}+1,\qquad \beta(x)=e^{\theta_4(x)},
\]
with regression trees of depth \texttt{max\_depth} as base learners. It minimizes
the evidential score
\begin{align*}
\mathcal{S}(x_i,y_i)={}&\mathcal{L}_{\mathrm{NIG}}(x_i,y_i)
+\lambda_{\mathrm{evi}}\,\frac1n\sum_{j=1}^{n}\lvert y_j-\mu_j\rvert\,(2\alpha_j+\lambda_j)\\
&+\lambda_{\mathrm{KL}}\,\frac1n\sum_{j=1}^{n}\mathrm{KL}\bigl(\mathrm{NIG}(\mu_j,\lambda_j,\alpha_j,\beta_j)\,\Vert\,\mathrm{NIG}(0,1,2,1)\bigr),
\end{align*}
where \(\mathcal{L}_{\mathrm{NIG}}\) is the NIG marginal negative log-likelihood (the
same expression as for the evidential BNN below, with \(v=\lambda\)), the second term
is the evidential regularizer of \citet{amini2020deepevidentialregression}, and
the third pulls each prediction towards a weak reference NIG. Boosting uses ordinary
gradients: the only Fisher information available for this score is a rank-one
empirical estimate, whose inverse applied to the gradient is not a natural
gradient. Targets are standardized. Uncertainty follows the NIG decomposition,
with aleatoric uncertainty \(\beta(x)/(\alpha(x)-1)\), epistemic uncertainty
\(\beta(x)/\bigl(\lambda(x)(\alpha(x)-1)\bigr)\), and total uncertainty their sum.
\textbf{Hyperparameter search space:}
\texttt{n\_estimators} \(\in [200,800]\) (integer),
\texttt{learning\_rate} \(\in [0.01,0.2]\) (log-uniform),
\texttt{max\_depth} \(\in [2,6]\) (integer),
\texttt{minibatch\_frac} \(\in [0.5,1.0]\) (uniform),
\texttt{col\_sample} \(\in [0.5,1.0]\) (uniform),
\texttt{natural\_gradient}=\texttt{False},
\texttt{use\_svgd}=\texttt{False},
\texttt{evid\_strength} (\(\lambda_{\mathrm{evi}}\)) \(\in [10^{-4},1]\) (log-uniform),
\texttt{kl\_strength} (\(\lambda_{\mathrm{KL}}\)) \(\in [10^{-4},1]\) (log-uniform), and
\texttt{standardize\_y}=\texttt{True}. The space also contains
\texttt{length\_scale} \(\in [0.1,5.0]\), which only affects the SVGD variant
and is therefore inactive.

\paragraph{NGBoost bagging ensemble (NGBoost-Bag).}
The NGBoost bagging baseline is an ensemble of \texttt{n\_regressors} NGBoost
regressors \citep{ngboost} with Gaussian predictive distributions, each trained
under the Gaussian log score on its own resample of the training data. A resample
has \(\lfloor\texttt{sample\_fraction}\cdot n\rfloor\) rows, drawn with or without
replacement according to \texttt{replace} \citep{Breiman1996Bagging}. Each member
predicts a Gaussian location and scale. The benchmark prediction is the average of
the member means. Aleatoric uncertainty is the average member predictive variance,
epistemic uncertainty is the variance across member means, and total uncertainty
is the sum of both terms. Optionally, the target can be standardized before
fitting, with all reported means and variances rescaled to the original target
units. \textbf{Hyperparameter search space:}
\texttt{n\_estimators} \(\in [200,800]\) (integer),
\texttt{learning\_rate} \(\in [0.005,0.1]\) (log-uniform),
\texttt{max\_depth} \(\in [2,6]\) (integer),
\texttt{minibatch\_frac} \(\in [0.5,1.0]\) (uniform),
\texttt{col\_sample} \(\in [0.5,1.0]\) (uniform),
\texttt{natural\_gradient}\(\in\{\mathrm{True},\mathrm{False}\}\),
\texttt{n\_regressors} \(\in [5,25]\) (integer),
\texttt{sample\_fraction} \(\in [0.6,0.95]\) (uniform),
\texttt{replace}\(\in\{\mathrm{True},\mathrm{False}\}\), and
\texttt{standardize\_y}\(\in\{\mathrm{True},\mathrm{False}\}\).

\paragraph{CatBoost posterior sampling (CatBoost-PS).}
This tree ensemble trains \texttt{n\_regressors} CatBoost regressors
\citep{prokhorenkova2019catboostunbiasedboostingcategorical} with the
\texttt{RMSEWithUncertainty} objective, under which each member predicts a mean
and a variance. Each member uses CatBoost's posterior-sampling mode, which trains
it by stochastic gradient Langevin boosting so that it approximates a sample from
the posterior over models \citep{malinin2021uncertainty}, with a Bayesian bootstrap
of strength \texttt{bagging\_temperature} and its own random seed. If
\texttt{bagging\_frac} \(<1\), each member is in addition trained on a bootstrap
sample of \(\lfloor\texttt{bagging\_frac}\cdot n\rfloor\) rows. The benchmark
averages member means for prediction, averages the member-provided variances for
aleatoric uncertainty, and uses the variance across member means for epistemic
uncertainty. When target standardization is enabled, all means and variances are
transformed back to the original target scale.
\textbf{Hyperparameter search space:}
\texttt{iterations} \(\in [100,600]\) (integer),
\texttt{learning\_rate} \(\in [0.01,0.3]\) (log-uniform),
\texttt{depth} \(\in [3,8]\) (integer),
\texttt{l2\_leaf\_reg} \(\in [0.5,10]\) (log-uniform),
\texttt{random\_strength} \(\in [0,2]\) (uniform),
\texttt{bagging\_temperature} \(\in [0,2]\) (uniform),
\texttt{n\_regressors} \(\in [5,20]\) (integer),
\texttt{bagging\_frac} \(\in [0.6,1.0]\) (uniform), and
\texttt{standardize\_y}\(\in\{\mathrm{True},\mathrm{False}\}\).

\paragraph{CatBoost kernel gradient boosting (CatBoost-KGB).}
The CatBoost KGB baseline uses CatBoost's \texttt{sample\_gaussian\_process}
routine, which implements the kernel gradient boosting sampling procedure of
\citet{ustimenko2023gradientboostingperformsgaussian}. In this view, gradient
boosting with symmetric trees is used to approximately sample functions from a
Gaussian-process posterior under a prior of the form
\(f \sim \mathcal{GP}(0,\sigma^2 K+\delta^2 I)\). The wrapper draws
\texttt{n\_regressors} sampled tree functions and forms predictions by averaging
them; epistemic uncertainty is the variance across the sampled functions. The
routine returns function values only and no observation-noise estimate, so the
aleatoric uncertainty is estimated from held-out residuals, as for DEUP: a second
ensemble with the same hyperparameters is trained on a random 90\% of the training
data, and the aleatoric variance is its mean squared error on the remaining 10\%.
This estimate is constant in \(x\). The final ensemble is then trained on all
training data. If target standardization is enabled, predictive means and
variances are transformed back to the original target scale. We use the
implementation available in the CatBoost library
\citep{prokhorenkova2019catboostunbiasedboostingcategorical}.
\textbf{Hyperparameter search space:}
\texttt{posterior\_iterations} \(\in [200,1000]\) (integer),
\texttt{prior\_iterations} \(\in [50,300]\) (integer),
\texttt{learning\_rate} \(\in [0.03,0.3]\) (log-uniform),
\texttt{depth} \(\in [4,8]\) (integer),
\texttt{sigma} \(\in [0.05,0.5]\) (log-uniform),
\texttt{delta} \(\in [0.01,0.5]\) (log-uniform),
\texttt{random\_strength} \(\in [0.05,1.0]\) (log-uniform),
\texttt{eps} \(\in [10^{-5},10^{-3}]\) (log-uniform),
\texttt{n\_regressors} \(\in [5,20]\) (integer), and
\texttt{standardize\_y}\(\in\{\mathrm{True},\mathrm{False}\}\).

\subsection{Bayesian Neural Network Models}

\paragraph{Shared neural network setup.}
All neural-network methods use the same residual tabular network, inspired by
TabResNet \citep{gorishniy2021revisiting}, implemented in JAX/Flax in single
precision. Given an input \(x\), the backbone first forms
\[
h^{(0)}=\operatorname{Dropout}\!\big(\sigma(W_{\mathrm{in}}x+b_{\mathrm{in}})\big),
\]
and then applies \(L\) residual blocks of the form
\[
u^{(\ell)}=\operatorname{Dropout}\!\big(\sigma(W^{(\ell)}_1 h^{(\ell)}+b^{(\ell)}_1)\big),
\qquad
h^{(\ell+1)}=\sigma\!\big(W^{(\ell)}_2 u^{(\ell)}+b^{(\ell)}_2+h^{(\ell)}\big),
\qquad \ell=0,\dots,L-1,
\]
with the ReLU activation \(\sigma\), He initialization and no normalization
layers. The dropout rate is zero except for MC dropout. The final representation
is passed to a linear output head. Width and depth are set by
\texttt{resnet\_width} and \texttt{resnet\_blocks}.

Inputs and targets are standardized during training. If
\(\tilde y=(y-m_y)/s_y\) denotes the standardized target, predictive quantities
are transformed back to the original target scale via
\[
\mu(x)=s_y\tilde\mu(x)+m_y,\qquad
v(x)=s_y^2\tilde v(x),
\]
and analogously for aleatoric, epistemic, and total variances.

Gaussian-output methods (all except EDL and DEUP) predict a mean and a raw scale
parameter,
\[
(\mu(x),\rho(x)),\qquad
s(x)=\operatorname{softplus}(\rho(x))+\varepsilon,\qquad
v(x)=s(x)^2,
\]
with \(\varepsilon=10^{-6}\), and are trained on the Gaussian negative
log-likelihood plus a weak penalty on implausibly large variances,
\[
\ell(y,x)
=
\frac12\left(
\log v(x)+\frac{(y-\mu(x))^2}{v(x)}
\right)
+\lambda_{\mathrm{var}}\,\Bigl[\max\bigl\{0,\ \log v(x)-\log\bigl(20\operatorname{Var}(\tilde y)\bigr)\bigr\}\Bigr]^2,
\]
up to the additive constant \(\frac12\log(2\pi)\), with
\(\lambda_{\mathrm{var}}=10^{-3}\). At prediction time, member variances are
clipped to \([\varepsilon,\,50\operatorname{Var}(\tilde y)]\). The implementation
also supports the natural parameterization of the same Gaussian,
\(\mu=-\eta_1/(2\eta_2)\) and \(v=-1/(2\eta_2)\). It is not used for any reported
result: in it the mean is \(\eta_1\) times the model's own variance, so every
perturbation of the network that the ensemble-, dropout-, Laplace- and SWAG-type
methods use to measure epistemic uncertainty moves the mean in proportion to the
aleatoric estimate, which couples the two estimates.

All networks are trained for \texttt{n\_epochs} epochs of shuffled minibatches with
NAdamW \citep{dozat2015incorporating,loshchilov2019decoupled}: a constant learning
rate \texttt{lr}, first-moment coefficient \(\beta_1=\)\texttt{momentum},
\(\beta_2=0.999\), \(\epsilon=10^{-8}\), decoupled weight decay
\texttt{weight\_decay}, and gradients clipped to a global norm of 1. For ensembles
the clipping is applied per member, and the loss is averaged over the batch but
summed over members, so each member receives exactly the gradient it would receive
if trained alone.

For methods with \(M\) stochastic forward passes, ensemble members or posterior
samples \(\{(\mu_m(x),v_m(x))\}_{m=1}^M\), we report
\[
\hat\mu(x)=\frac1M\sum_{m=1}^M \mu_m(x),\qquad
\hat v_{\mathrm{ale}}(x)=\frac1M\sum_{m=1}^M v_m(x),
\]
\[
\hat v_{\mathrm{epi}}(x)=\operatorname{Var}_m(\mu_m(x)),\qquad
\hat v_{\mathrm{tot}}(x)=\hat v_{\mathrm{ale}}(x)+\hat v_{\mathrm{epi}}(x).
\]
The number of members \(M\) is a different resource for different methods. For
bagging and deep ensembles, a member is a full retraining, so \(M=10\). For the
post-hoc samplers, a member is one forward pass, so \(M\) is set high (128 for MC
dropout and SWAG, 256 for the Laplace variants): the sample variance of \(M\) draws
carries multiplicative Monte Carlo noise that alone caps the achievable rank
correlation, at roughly 0.7 for \(M=10\) and 0.97 for \(M=128\).
\textbf{Shared hyperparameter search space:} for all neural-network methods we
tuned \texttt{resnet\_width}\(\in\{32,64,128,256,512\}\),
\texttt{resnet\_blocks}\(\in\{1,2,3,4\}\),
\texttt{n\_epochs}\(\in [50,200]\) (integer),
\texttt{batch\_size}\(\in\{64,128,256,512\}\),
\texttt{lr}\(\in [10^{-4},5\cdot 10^{-3}]\) (log-uniform),
\texttt{weight\_decay}\(\in [10^{-6},10^{-2}]\) (log-uniform), and
\texttt{momentum}\(\in [0.7,0.95]\) (uniform). The main benchmark tunes no
optimizer-specific parameter. The
optimizer spaces were: Adam with \texttt{adam\_b2}\(\in [0.95,0.999]\)
(uniform) and \texttt{adam\_eps}\(\in [10^{-9},10^{-6}]\) (log-uniform).

\paragraph{BNN bagging (NN-Bag).}
The bagging BNN uses the shared network with \(M\) independently initialized
members, trained on the Gaussian objective above. In contrast to the deep ensemble
below, each member is trained on a fixed subsample of the training data of size
\(\lfloor\texttt{sample\_fraction}\cdot n\rfloor\), drawn with or without replacement
according to \texttt{replace}. The predictive mean is the average of the member
means, aleatoric uncertainty is the average of the member predictive variances,
and epistemic uncertainty is the variance across member means; total uncertainty
is their sum \citep{Breiman1996Bagging}. \textbf{Additional hyperparameter search
space:} \texttt{n\_members}\(=10\),
\texttt{sample\_fraction}\(\in [0.6,0.95]\) (uniform), and
\texttt{replace}\(\in\{\mathrm{True},\mathrm{False}\}\).

\paragraph{Deep ensemble BNN (NN-Ens).}
The deep ensemble uses the shared network with \(M\) independently initialized
members of identical architecture. All members are trained on the full training
set with separate parameters and optimization trajectories, using the Gaussian
objective above. The predictive mean is obtained by averaging member means,
aleatoric uncertainty by averaging member predictive variances, and epistemic
uncertainty by the variance across member means; total uncertainty is the sum of
the aleatoric and epistemic terms
\citep{lakshminarayanan2017simplescalablepredictiveuncertainty}.
\textbf{Additional hyperparameter search space:} \texttt{n\_members}\(=10\).

\paragraph{MC dropout BNN (NN-Drop).}
The MC-dropout BNN uses a single instance of the shared network, trained with
dropout (after the input layer and after the first layer of every residual block)
and the Gaussian objective above. At test time, dropout remains active and
\(M\) stochastic forward passes are drawn with different dropout masks. These
passes are treated like ensemble members: the predictive mean is the average of
the sampled means, aleatoric uncertainty is the average of the sampled predictive
variances, and epistemic uncertainty is the variance across sampled means; total
uncertainty is their sum \citep{galDropout}.
\textbf{Additional hyperparameter search space:} \texttt{n\_members}\(=128\) and
\texttt{dropout}\(\in [0.05,0.35]\) (uniform).

\paragraph{Evidential BNN (NN-EDL).}
The evidential BNN uses the shared network with a four-dimensional evidential
head that predicts the parameters of a Normal-Inverse-Gamma (NIG) distribution
\citep{amini2020deepevidentialregression},
\[
\mu(x),\qquad v(x)>0,\qquad \alpha(x)>1,\qquad \beta(x)>0,
\]
where the constraints are enforced through softplus transformations. The model is
trained with the NIG marginal negative log-likelihood together with an evidential
regularizer,
\[
\mathcal{L}_{\mathrm{EDL}}(x,y)
=
\mathcal{L}_{\mathrm{NIG}}(x,y)
+
\lambda_{\mathrm{evi}}\,
\mathcal{R}_{\mathrm{evi}}(x,y),
\]
where \(\lambda_{\mathrm{evi}}\) corresponds to \texttt{evi\_reg}. The NIG
marginal negative log-likelihood is
\begin{align}
\mathcal{L}_{\mathrm{NIG}}(x,y)
&=
\frac12\bigl(\log \pi-\log v(x)\bigr)
-\alpha(x)\log\!\bigl(2\beta(x)(1+v(x))\bigr)
\\
&\quad
+\Bigl(\alpha(x)+\frac12\Bigr)
\log\!\Bigl(v(x)\bigl(y-\mu(x)\bigr)^2+2\beta(x)(1+v(x))\Bigr)
\\
&\quad
+\log\Gamma\!\bigl(\alpha(x)\bigr)
-\log\Gamma\!\Bigl(\alpha(x)+\frac12\Bigr),
\end{align}
and the evidential regularizer is
\[
\mathcal{R}_{\mathrm{evi}}(x,y)
=
|y-\mu(x)|\,\bigl(2v(x)+\alpha(x)\bigr).
\]
At test time, the predictive mean is \(\mu(x)\), aleatoric uncertainty is
\[
\sigma^2_{\mathrm{ale}}(x)=\frac{\beta(x)}{\alpha(x)-1},
\]
and epistemic uncertainty is
\[
\sigma^2_{\mathrm{epi}}(x)=\frac{\beta(x)}{v(x)(\alpha(x)-1)},
\]
with total uncertainty given by
\[
\sigma^2_{\mathrm{tot}}(x)=
\sigma^2_{\mathrm{ale}}(x)+\sigma^2_{\mathrm{epi}}(x).
\]
\textbf{Additional hyperparameter search space:}
\texttt{evi\_reg}\(\in [10^{-3},3\cdot 10^{-1}]\) (log-uniform).

\paragraph{DEUP BNN (NN-DEUP).}
The DEUP model follows Algorithm~1 of \citet{lahlou2023} for a fixed training set
and consists of two networks with the shared architecture. First, a mean regressor
\(m(x)\) is trained with squared-error loss on a random 80\% of the training data.
The held-out 20\%, \(\mathcal V\), gives a single global aleatoric variance,
\[
\hat\sigma^2_{\mathrm{ale}}
=
\frac{1}{|\mathcal V|}
\sum_{(x_i,y_i)\in\mathcal V}
\bigl(y_i-m(x_i)\bigr)^2 ,
\]
since the benchmark provides no repeated measurements from which an
input-dependent noise level could be estimated. Second, an error predictor
\(e(x)\) is trained under squared loss on the same held-out points to predict the
out-of-sample loss \(e_i=\bigl(y_i-m(x_i)\bigr)^2\), i.e.\ the pointwise risk of
\(m\). At prediction time, the first network provides the predictive mean, and
epistemic uncertainty is the predicted risk minus the aleatoric level,
\(u(x)=e(x)-\hat\sigma^2_{\mathrm{ale}}\), with \(e(x)\) bounded above by
\(50\operatorname{Var}(\tilde y)\). Because \(\hat\sigma^2_{\mathrm{ale}}\) is the
mean squared error, \(u(x)\) is negative wherever the predicted risk is below
average. Rather than flooring these values at zero, which would tie all of them,
\(u\) is shifted by one constant, \(\hat\sigma^2_{\mathrm{epi}}(x)=u(x)-\min\{0,\min_{x'}u(x')\}\),
with the minimum taken over the evaluation points. This preserves the ranking of
\(u\) and changes only its level. Aleatoric uncertainty is the constant
\(\hat\sigma^2_{\mathrm{ale}}\), and total uncertainty is the sum. The two stages
are tuned in sequence: a sweep over the shared search space selects the mean
network (with the error network at its defaults), and a second 30-trial sweep then
selects the error network by its held-out mean squared error in predicting the
loss, with the mean network fixed at the first sweep's winner.
\textbf{Additional hyperparameter search space:} the first-stage regressor uses
the shared search space above. The second-stage error network was tuned over
\texttt{lr\_excess}\(\in [10^{-4},10^{-1}]\) (log-uniform),
\texttt{weight\_decay\_excess}\(\in [10^{-6},10^{-2}]\) (log-uniform),
\texttt{n\_epochs\_excess}\(\in [30,200]\) (integer), and
\texttt{batch\_size\_excess}\(\in\{32,64,128,256\}\).

\paragraph{Laplace BNN (NN-Lap).}
The Laplace BNN is a post-hoc method applied to a point-estimate network
\citep{daxberger_laplaceredux}. We first train the shared Gaussian-output network
on all training data to obtain a MAP solution. Following standard last-layer
Laplace practice \citep{daxberger_laplaceredux,laplax}, we then approximate a
Gaussian posterior over the weights and bias of the mean output only,
\(\omega=(w_\mu,b_\mu)\), while keeping the backbone and the variance output fixed
at their MAP values. With last-layer features \(\phi(x)=[h(x);1]\), the posterior
precision is the full generalized Gauss--Newton matrix, weighted by the MAP noise
variance of each training point,
\[
\Lambda=\tau I+\sum_{i=1}^{n}\frac{\phi(x_i)\phi(x_i)^\top}{v_{\mathrm{MAP}}(x_i)},
\]
which is exact for this linear-Gaussian last layer. The prior precision \(\tau\)
maximizes the Laplace approximation of the log marginal likelihood of the training
data \citep{mackay_1992_r7qgh-q6g10,immer2021scalablemarginallikelihoodestimation} over a logarithmic grid of
100 values in \([10^{-4},10^{4}]\); the posterior is not tempered and no data are
held out. The approximation is constructed with \texttt{laplax} \citep{laplax}. At
prediction time we draw \(M\) samples \(\omega_m\sim\mathcal N(\omega_{\mathrm{MAP}},\Lambda^{-1})\)
and propagate them through the fixed backbone. Epistemic uncertainty is the
variance of the sampled means, a Monte Carlo estimate of
\(\phi(x)^\top\Lambda^{-1}\phi(x)\); aleatoric uncertainty is the MAP variance
output; total uncertainty is their sum. \textbf{Additional hyperparameter search
space:} \texttt{n\_members}\(=256\).

\paragraph{Laplace BNN with spectral normalization (NN-Lap-SN).}
The same method on a spectrally normalized backbone: every dense layer of the
backbone (input layer and both layers of every residual block) is constrained to
\(\lVert W\rVert_2\le\)\texttt{sn\_coeff}, as in the distance-aware feature
extractor above, and the method is tuned under the constraint. The largest
singular value is tracked by persistent power iteration during training and
refined by additional power iterations at prediction time.
\textbf{Additional hyperparameter search space:} \texttt{n\_members}\(=256\),
\texttt{spectral\_norm}\(=\)\texttt{True} and \texttt{sn\_coeff}\(\in[0.5,3]\)
(log-uniform).

\paragraph{Function-space prior Laplace BNN (NN-FSP).}
FSP-Laplace \citep{cinquin2024fsplaplacefunctionspacepriorslaplace} replaces the
isotropic weight prior of the Laplace BNN with a Gaussian-process prior on the
network's mean function. MAP training adds the corresponding function-space
regularizer at a set \(C\) of context points,
\[
n\cdot\overline{\ell}_{\mathrm{batch}}
+\frac12\,\bigl(\mu_\theta(C)-m(C)\bigr)^\top K_{CC}^{-1}\bigl(\mu_\theta(C)-m(C)\bigr),
\]
where \(\overline{\ell}_{\mathrm{batch}}\) is the minibatch average of the Gaussian
objective above, the prior mean is \(m=0\), and the prior kernel on standardized
inputs is
\[
k(x,x')=\sigma_k^2\bigl(k_{3/2}(r)+\tilde x^\top\tilde x'\bigr),\qquad
\tilde x=x/\ell_k,\quad r=\lVert\tilde x-\tilde x'\rVert,
\]
a Matérn-\(\tfrac32\) plus linear kernel with \(\sigma_k=\ell_k=1\). During training,
\(C\) consists of 64 training inputs resampled at every step. After training, the
backbone and the variance output are fixed at their MAP values, and a Gaussian
posterior over the weights and bias of the mean output is constructed with the
low-rank function-space Laplace approximation of
\citet[Algorithm~2]{cinquin2024fsplaplacefunctionspacepriorslaplace}. The prior
term uses 256 context points from a Halton sequence covering the bounding box of
the training inputs, with at most 128 eigendirections of \(K_{CC}\). The data term
is \(\sum_i J(x_i)^\top J(x_i)/v_{\mathrm{MAP}}(x_i)\) over all training points,
where \(J\) is the Jacobian of the mean output with respect to these parameters.
Leading posterior directions are then dropped until the posterior variance at the
context points no longer exceeds the prior variance. The posterior is not tempered
and no data are held out. At
prediction time, \(M\) weight samples give epistemic uncertainty as the variance of
the sampled means, while the MAP variance output provides the aleatoric
uncertainty. \textbf{Additional hyperparameter search space:}
\texttt{n\_members}\(=256\).

\paragraph{Function-space prior Laplace BNN with spectral normalization (NN-FSP-SN).}
FSP-Laplace on the spectrally normalized backbone described for NN-Lap-SN, tuned
under the constraint. \textbf{Additional hyperparameter search space:}
\texttt{n\_members}\(=256\), \texttt{spectral\_norm}\(=\)\texttt{True} and
\texttt{sn\_coeff}\(\in[0.5,3]\) (log-uniform).

\paragraph{SWAG BNN (NN-SWAG).}
The SWAG BNN \citep{maddox2019simplebaselinebayesianuncertainty} first removes a
random calibration split \(\mathcal C\) of relative size
\texttt{calibration\_fraction} from the training data and trains the shared
Gaussian-output network on the rest, as above. Starting from this solution, it
then enters a collection phase in which training continues along a single SGD
trajectory with constant learning rate \texttt{swag\_lr}, momentum
\texttt{swag\_momentum} and no weight decay. After every
\texttt{epoch\_per\_checkpoint} epochs, for \texttt{n\_checkpoints} checkpoints, the
first and second parameter moments \(E[\theta]\) and \(E[\theta^2]\) are updated
online; the pretrained solution enters as the first point. These moments define a
diagonal Gaussian posterior approximation over all network parameters,
\[
\sigma^2_{\mathrm{SWAG}}
=
E[\theta^2]-E[\theta]^2,
\qquad
\theta
\sim
\mathcal{N}\!\left(
\mu_{\mathrm{SWAG}},
\operatorname{diag}(\sigma^2_{\mathrm{SWAG}})
\right),
\]
where \(\mu_{\mathrm{SWAG}}=E[\theta]\). At prediction time, \(M\) networks are
sampled from this posterior and evaluated with deterministic forward passes. The
predictive mean is the average of the sampled means, aleatoric uncertainty is the
average sampled variance output, and epistemic uncertainty is the variance across
sampled means. If \(\mathcal C\) is non-empty, a single global variance scale is
fitted on it,
\[
c
=
\frac{
\frac{1}{|\mathcal C|}
\sum_{(x_i,y_i)\in\mathcal C}
\bigl(y_i-\hat\mu(x_i)\bigr)^2
}{
\frac{1}{|\mathcal C|}
\sum_{(x_i,y_i)\in\mathcal C}
\hat\sigma^2_{\mathrm{tot}}(x_i)
},
\]
clipped to \([10^{-3},10^{3}]\), and both aleatoric and epistemic variances are
multiplied by \(c\) at test time; otherwise \(c=1\). The scale does not change any
ranking. \textbf{Additional hyperparameter search space:}
\texttt{n\_members}\(=128\),
\texttt{n\_checkpoints}\(\in [10,40]\) (integer),
\texttt{epoch\_per\_checkpoint}\(\in [2,12]\) (integer),
\texttt{swag\_lr}\(\in [10^{-3},10^{-1}]\) (log-uniform),
\texttt{swag\_momentum}\(\in\{0.0,0.9,0.95\}\),
\texttt{calibration\_fraction}\(\in [0,0.25]\) (uniform), and
\texttt{grad\_clip\_norm}\(\in [0.5,2.0]\) (uniform), which replaces the default
clipping norm of 1 in both phases.
\par\endgroup
\newpage

\section{Supplementary Results}
\label{app:additional-results}
\subsection{Stability of model rankings across tasks}
\label{app:stability_rankings}

\begin{figure}[H]
    \centering
    \includegraphics[width=\linewidth]{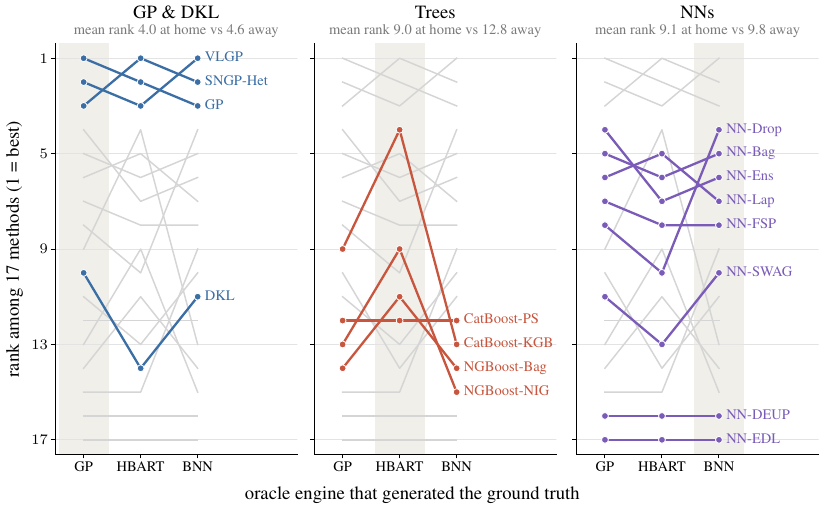}
    \caption{Rank of each method under each oracle engine, by its mean Spearman $\rho$ between EU output and
    posterior excess risk (1 = best of 17). One panel per family. The other methods are shown in grey, and the
    shaded column marks the oracle whose prior matches the family (GP for GP \& DKL, HBART for trees, BNN for
    NNs). BLR has no matching oracle and appears only in grey.}
    \label{fig:home_ranks}
\end{figure}

\begin{figure}[H]
    \centering
    \includegraphics[width=1\linewidth]{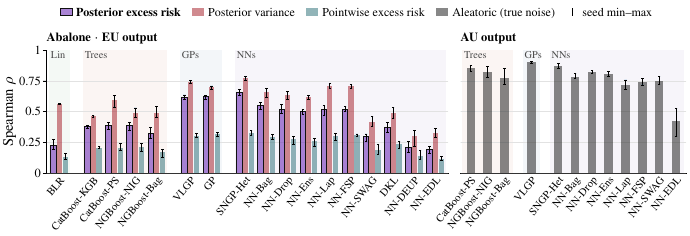}
    \caption{The complementary figure to Figure 2 of the main text, but reduced to only one dataset (abalone)}
    \label{fig:abalone_ranking}
\end{figure}

\begin{figure}[H]
    \centering
    \includegraphics[width=1\linewidth]{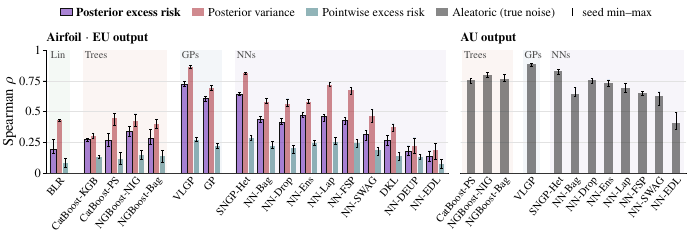}
    \caption{The complementary figure to Figure 2 of the main text, but reduced to only one dataset (airfoil)}
    \label{fig:airfoil_ranking}
\end{figure}

The method ranking depends little on the experimental condition (Table~\ref{tab:kendall_w}): across the levels
of each factor, Kendall's $W$ is at least $0.81$ for the epistemic and $0.93$ for the aleatoric criterion. The
least stable factor is the oracle engine ($W = 0.81$ on ranks), so we check whether methods rank better when the
ground truth is drawn from a prior like their own (Figure~\ref{fig:home_ranks}). This matched-prior advantage is
real: GP-based methods and NNs gain about one rank at their matched oracle (mean rank $4.0$ vs.\ $4.6$ and
$9.1$ vs.\ $9.8$), and trees about four ranks under HBART ($9.0$ vs.\ $12.8$). The top three (GP, VLGP, SNGP-Het) and the bottom two (NN-DEUP,
NN-EDL) are the same under every oracle. 
The rankings of methods on EU defined as posterior risk largely agree between datasets as well (Kendal's W=078). We showcase the results for representative datasets (abalone, airfoil) in \Cref{fig:abalone_ranking} and \Cref{fig:airfoil_ranking}.

\begin{table}[H]
\centering
\caption{Stability of the method rankings across the levels of each experimental factor, measured by Kendall's
coefficient of concordance $W$ (tie-corrected; $W=1$: every level ranks the methods identically, $W=0$: no
agreement). Each level contributes one ranking of the methods, from their mean rank within the conditions of that
level (ranks) or from their mean Spearman $\rho$ over them ($\rho$). Epistemic: EU output against the posterior
excess risk; aleatoric: AU output against the true noise variance, for the methods with a heteroscedastic AU and the
three noise modes whose noise varies.}
\label{tab:kendall_w}
\small
\begin{tabular}{@{}llcccc@{}}
  \toprule
  Criterion & Factor & Levels & Methods & $W$ (ranks) & $W$ ($\rho$) \\
  \midrule
  Epistemic & Oracle engine      & 3 & 19 & 0.81 & 0.92 \\
            & Noise mode         & 4 & 19 & 0.94 & 0.97 \\
            & Covariate coverage & 2 & 19 & 0.98 & 0.99 \\
  \addlinespace
  Aleatoric & Oracle engine      & 3 & 14 & 0.93 & 0.94 \\
            & Noise mode         & 3 & 14 & 0.96 & 0.99 \\
            & Covariate coverage & 2 & 14 & 0.99 & 1.00 \\
  \bottomrule
\end{tabular}
\end{table}


\FloatBarrier

\subsection{Disentanglement}
\label{app:disentanglement}
This subsection reports the full excess-entanglement results underlying \Cref{fig:disentanglement_ood}a. We compare each method's EU--AU coupling with the corresponding ground-truth coupling under posterior excess risk and posterior variance, both overall and across oracle engines and coverage settings. We then examine the pronounced coupling of NIG-based evidential methods and present possible explanations for the additional coupling observed with natural-parameter heads and VLGP.
\begin{table}[h]
\centering
\caption{Excess disentanglement per method with the ground-truth EU defined as the posterior excess risk, $\mathrm{EU}^\star = \mathbb{E}_{\mathrm{post}}\big[(\hat a_{\mathcal S}(x) - f(x))^2\big]$. $\Delta\rho = \rho(\mathrm{EU},\mathrm{AU}) - \rho(\mathrm{EU}^\star,\mathrm{AU}^\star)$ (Spearman, over the evaluation points), averaged over all conditions, and separately per oracle engine and per coverage scenario. $\Delta\rho > 0$: EU and AU outputs more strongly coupled than the truth.}
\label{tab:dis:per}
\small
\setlength{\tabcolsep}{3.5pt}
\begin{tabular}{@{}llcccccccc@{}}
  \toprule
  & & & & & \multicolumn{3}{c}{$\Delta\rho$ by oracle engine} & \multicolumn{2}{c}{$\Delta\rho$ by coverage} \\
  \cmidrule(lr){6-8}\cmidrule(l){9-10}
  & Method & $\rho(\mathrm{EU},\mathrm{AU})$ & $\rho(\mathrm{EU}^\star,\mathrm{AU}^\star)$ & $\Delta\rho$ & GP & HBART & BNN & bal. & OOD \\
  \midrule
  Trees & CatBoost-PS & 0.52 & 0.14 & \textbf{+0.38} & +0.42 & +0.35 & +0.39 & +0.37 & +0.40 \\
   & NGBoost-NIG & 1.00 & 0.29 & \textbf{+0.71} & +0.77 & +0.64 & +0.72 & +0.68 & +0.74 \\
   & NGBoost-Bag & 0.33 & 0.23 & \textbf{+0.10} & +0.13 & +0.03 & +0.14 & +0.07 & +0.12 \\
  \addlinespace
  GPs & VLGP & 0.51 & 0.31 & \textbf{+0.21} & +0.23 & +0.30 & +0.09 & +0.20 & +0.21 \\
  \addlinespace
  NNs & SNGP-Het & 0.46 & 0.29 & \textbf{+0.17} & +0.18 & +0.21 & +0.13 & +0.20 & +0.14 \\
   & NN-Bag & 0.35 & 0.26 & \textbf{+0.09} & +0.09 & +0.16 & +0.02 & +0.07 & +0.11 \\
   & NN-Drop & 0.50 & 0.26 & \textbf{+0.24} & +0.23 & +0.29 & +0.19 & +0.22 & +0.25 \\
   & NN-Ens & 0.36 & 0.26 & \textbf{+0.10} & +0.09 & +0.14 & +0.05 & +0.08 & +0.11 \\
   & NN-Lap & 0.40 & 0.21 & \textbf{+0.20} & +0.18 & +0.22 & +0.20 & +0.24 & +0.16 \\
   & NN-FSP & 0.42 & 0.20 & \textbf{+0.22} & +0.21 & +0.24 & +0.21 & +0.24 & +0.19 \\
   & NN-SWAG & 0.57 & 0.22 & \textbf{+0.34} & +0.31 & +0.44 & +0.29 & +0.29 & +0.39 \\
   & NN-EDL & 0.57 & 0.21 & \textbf{+0.35} & +0.30 & +0.34 & +0.42 & +0.35 & +0.36 \\
  \bottomrule
\end{tabular}
\end{table}
\begin{table}[h]
\centering
\caption{Excess disentanglement per method with the ground-truth EU defined as the posterior variance, $\mathrm{EU}^\star = \mathrm{Var}_{\mathrm{post}}\, f(x)$. $\Delta\rho = \rho(\mathrm{EU},\mathrm{AU}) - \rho(\mathrm{EU}^\star,\mathrm{AU}^\star)$ (Spearman, over the evaluation points), averaged over all conditions, and separately per oracle engine and per coverage scenario. $\Delta\rho > 0$: EU and AU outputs more strongly coupled than the truth.}
\label{tab:dis:pv}
\small
\setlength{\tabcolsep}{3.5pt}
\begin{tabular}{@{}llcccccccc@{}}
  \toprule
  & & & & & \multicolumn{3}{c}{$\Delta\rho$ by oracle engine} & \multicolumn{2}{c}{$\Delta\rho$ by coverage} \\
  \cmidrule(lr){6-8}\cmidrule(l){9-10}
  & Method & $\rho(\mathrm{EU},\mathrm{AU})$ & $\rho(\mathrm{EU}^\star,\mathrm{AU}^\star)$ & $\Delta\rho$ & GP & HBART & BNN & bal. & OOD \\
  \midrule
  Trees & CatBoost-PS & 0.52 & 0.42 & \textbf{+0.10} & +0.19 & +0.03 & +0.09 & +0.04 & +0.16 \\
   & NGBoost-NIG & 1.00 & 0.42 & \textbf{+0.58} & +0.65 & +0.54 & +0.53 & +0.52 & +0.63 \\
   & NGBoost-Bag & 0.33 & 0.42 & \textbf{$-$0.09} & $-$0.04 & $-$0.15 & $-$0.08 & $-$0.15 & $-$0.03 \\
  \addlinespace
  GPs & VLGP & 0.51 & 0.42 & \textbf{+0.09} & +0.16 & +0.04 & +0.07 & +0.06 & +0.12 \\
  \addlinespace
  NNs & SNGP-Het & 0.46 & 0.42 & \textbf{+0.04} & +0.09 & $-$0.04 & +0.06 & +0.05 & +0.03 \\
   & NN-Bag & 0.35 & 0.42 & \textbf{$-$0.07} & $-$0.02 & $-$0.11 & $-$0.10 & $-$0.12 & $-$0.03 \\
   & NN-Drop & 0.50 & 0.42 & \textbf{+0.08} & +0.13 & +0.03 & +0.07 & +0.03 & +0.12 \\
   & NN-Ens & 0.36 & 0.42 & \textbf{$-$0.07} & 0.00 & $-$0.12 & $-$0.08 & $-$0.11 & $-$0.02 \\
   & NN-Lap & 0.40 & 0.42 & \textbf{$-$0.02} & +0.04 & $-$0.07 & $-$0.02 & $-$0.01 & $-$0.02 \\
   & NN-FSP & 0.42 & 0.42 & \textbf{0.00} & +0.06 & $-$0.06 & $-$0.01 & $-$0.01 & +0.01 \\
   & NN-SWAG & 0.57 & 0.42 & \textbf{+0.15} & +0.20 & +0.15 & +0.09 & +0.06 & +0.23 \\
   & NN-EDL & 0.57 & 0.42 & \textbf{+0.15} & +0.17 & +0.06 & +0.21 & +0.10 & +0.19 \\
  \bottomrule
\end{tabular}
\end{table}

\paragraph{Normal--Inverse-Gamma evidential objectives.}
The excess entanglement of NIG-based evidential methods follows from properties previously analyzed by \citet{amini2020deepevidentialregression}, \citet{meinert2023unreasonable}, and \citet{jürgens2024epistemicuncertaintyfaithfullyrepresented}. We summarize these results to interpret the coupling observed in our benchmark. For an NIG distribution with parameters \((\gamma,\nu,\alpha,\beta)\), the commonly used uncertainty estimates are
\begin{equation}
\widehat{\mathrm{AU}}(x)
=
\frac{\beta(x)}{\alpha(x)-1},
\qquad
\widehat{\mathrm{EU}}(x)
=
\frac{\beta(x)}
{\nu(x)[\alpha(x)-1]}
=
\frac{\widehat{\mathrm{AU}}(x)}{\nu(x)}
\label{eq:nig-entanglement}
\end{equation}
Both estimates therefore contain the same input-dependent scale
\(c(x)=\beta(x)/[\alpha(x)-1]\). Unless \(\nu(x)\) varies sufficiently to alter the ranking induced by \(c(x)\), AU and EU inherit a strong positive association.

This coupling is reinforced by the non-identifiability of the NIG decomposition. Marginalizing the NIG distribution produces a Student-\(t\) predictive distribution with location \(\gamma\), \(2\alpha\) degrees of freedom, and scale
\[
s^2=\frac{\beta(1+\nu)}{\alpha\nu}.
\]
For fixed \((\gamma,\alpha,s^2)\), every \(\nu>0\) yields the same predictive distribution when
\(\beta=s^2\alpha\nu/(1+\nu)\). Along this set of equivalent parameterizations,
\[
\widehat{\mathrm{AU}}
=
s^2\frac{\alpha}{\alpha-1}\frac{\nu}{1+\nu},
\qquad
\widehat{\mathrm{EU}}
=
s^2\frac{\alpha}{\alpha-1}\frac{1}{1+\nu},
\]
while
\[
\widehat{\mathrm{AU}}+\widehat{\mathrm{EU}}
=
s^2\frac{\alpha}{\alpha-1}
\]
remains unchanged. The marginal likelihood can therefore identify the predictive distribution without uniquely identifying its allocation into AU and EU. This non-injectivity of the NIG predictive mapping and the resulting non-identifiability of its second-order parameters were established by \citet{jürgens2024epistemicuncertaintyfaithfullyrepresented}.

The evidential regularizer introduces an additional coupling mechanism. The objective proposed by \citet{amini2020deepevidentialregression} contains
\[
R_{\mathrm{evi}}(x,y)
=
\lambda |y-\gamma(x)|[2\nu(x)+\alpha(x)].
\]
At the correct mean under Gaussian noise,
\[
\mathbb E[|Y-\gamma(x)|\mid x]
=
\sigma(x)\sqrt{\frac{2}{\pi}}.
\]
The expected contribution of the regularizer to the gradient with respect to \(\nu\) is therefore
\[
\mathbb E\!\left[
\frac{\partial R_{\mathrm{evi}}}{\partial\nu}
\,\middle|\,x
\right]
=
2\lambda\sigma(x)\sqrt{\frac{2}{\pi}}.
\]
All else fixed, the regularizer thus places stronger pressure toward smaller \(\nu\) in regions with larger aleatoric noise. Because \(\widehat{\mathrm{EU}}=\widehat{\mathrm{AU}}/\nu\), this transfers variation in AU into the epistemic estimate. \citet{meinert2023unreasonable} previously derived and demonstrated this residual-induced leakage under heteroscedastic noise.

These established properties account for the pronounced excess entanglement of NN-EDL and NGBoost-NIG. \Cref{eq:nig-entanglement} does not by itself imply perfect rank correlation, since variation in \(\nu(x)\) can change the ordering of EU relative to AU. The observed correlation of \(\rho=1.00\) for NGBoost-NIG therefore additionally indicates that \(\nu(x)\) does not substantially alter the ranking induced by the shared scale \(c(x)\).

\paragraph{Possible source of excess entanglement: natural-parameter heads.}
The standard and natural-parameter heads represent the same pointwise Gaussian family. Their different coupling cannot therefore be attributed to a different population-optimal conditional distribution. Natural parameterization instead changes the coordinates in which the finite network is optimized and in which variation across ensemble members or stochastic predictions is expressed \citep{immer2023NPLoss}. We consider the following algebraic argument as a post hoc explanation of the observed results, not as a general consequence of natural-parameter training.

For stochastic prediction \(m\), the natural parameters satisfy
\[
v_m(x)=-\frac{1}{2\eta_{2m}(x)},
\qquad
\mu_m(x)
=-\frac{\eta_{1m}(x)}{2\eta_{2m}(x)}
=\eta_{1m}(x)v_m(x).
\]
The resulting estimates are
\[
\widehat{\mathrm{AU}}(x)
=
\mathbb E_m[v_m(x)],
\qquad
\widehat{\mathrm{EU}}(x)
=
\operatorname{Var}_m[\eta_{1m}(x)v_m(x)].
\]
Writing \(\bar\eta_1(x)=\mathbb E_m[\eta_{1m}(x)]\) and
\(\bar v(x)=\mathbb E_m[v_m(x)]\), a first-order expansion gives
\[
\operatorname{Var}_m[\mu_m(x)]
\approx
\bar v(x)^2\operatorname{Var}_m[\eta_{1m}(x)]
+
\bar\eta_1(x)^2\operatorname{Var}_m[v_m(x)]
+
2\bar\eta_1(x)\bar v(x)
\operatorname{Cov}_m[\eta_{1m}(x),v_m(x)].
\]
If the stochastic predictions approximately agree on \(v_m(x)\), the last two terms are small and
\[
\widehat{\mathrm{EU}}(x)
\approx
\widehat{\mathrm{AU}}(x)^2
\operatorname{Var}_m[\eta_{1m}(x)].
\]
Under the additional condition that
\(\operatorname{Var}_m[\eta_{1m}(x)]\) varies more slowly across inputs than
\(\widehat{\mathrm{AU}}(x)\), disagreement in the first natural parameter is amplified in regions with large predicted AU. By contrast, the standard head outputs \(\mu_m(x)\) directly and does not impose this multiplicative dependence on \(v_m(x)\).

The optimization geometry provides a second possible contribution. For Gaussian NLL, the mean gradient in standard coordinates is
\[
\frac{\partial\ell}{\partial\mu}
=
\frac{\mu-y}{v},
\]
whereas the corresponding gradient in natural coordinates is
\[
\frac{\partial\ell}{\partial\eta_1}
=
\mu-y.
\]
Natural coordinates therefore remove the explicit precision weighting from the first-output gradient \citep{immer2023NPLoss}. Combined with
\(\mu=\eta_1v\), this may allow differences between independently trained members to produce greater mean disagreement in high-variance regions.

This explanation is consistent with the increased coupling observed for bagging and deep ensembles, but it is not supported uniformly across stochastic methods. Natural parameterization does not increase coupling for dropout or SWAG, showing that the covariance terms and the source of stochastic variation can offset the proposed amplification. Establishing this mechanism would require decomposing the three terms above from the member-level outputs or intervening on the variance head. We therefore treat it only as a possible explanation of the observed pattern.

\paragraph{Possible source of excess entanglement: posterior uncertainty in the VLGP noise process.}
VLGP represents the conditional mean and heteroscedastic noise through two latent functions. The signal process \(f\) models the conditional mean, while the noise process \(g=\log\sigma^2\) models the input-dependent observation variance. Both functions are inferred jointly using sparse variational GPs under
\[
Y\mid f,g,x\sim\mathcal N\!\left(f(x),\exp(g(x))\right),
\]
as in heteroscedastic GP regression \citep{lazaro2011variational}. Consequently, the model is uncertain not only about the signal function but also about the noise function. However, these two sources of model uncertainty enter the reported AU and EU asymmetrically. Writing
\[
q(f(x))=\mathcal N(m_f(x),v_f(x)),
\qquad
q(g(x))=\mathcal N(m_g(x),v_g(x)),
\]
VLGP reports
\[
\widehat{\mathrm{EU}}(x)=v_f(x),
\qquad
\widehat{\mathrm{AU}}(x)
=
\mathbb E_q[\exp(g(x))]
=
\exp\!\left(m_g(x)+\frac{v_g(x)}{2}\right).
\]
The uncertainty about \(f\) is therefore reported directly as EU, whereas uncertainty about \(g\) changes the reported AU through the posterior expectation of the noise variance. This decomposition is internally coherent because the law of total variance gives
\[
\operatorname{Var}_q(Y\mid x)
=
\mathbb E_q[\exp(g(x))]
+
\operatorname{Var}_q[f(x)].
\]
It nevertheless differs from the oracle used in our benchmark. The oracle AU is computed from the realized noise law, whereas VLGP must infer that law and marginalizes over its method-side uncertainty.

We consider the following argument a post hoc explanation of the observed excess entanglement rather than an established property of VLGP. Its AU estimate factorizes as
\[
\widehat{\mathrm{AU}}(x)
=
\underbrace{\exp(m_g(x))}_{\text{plug-in noise variance}}
\underbrace{\exp(v_g(x)/2)}_{\text{noise-posterior factor}}.
\]
The second factor is the log-normal correction arising from uncertainty about the noise process. At fixed \(m_g(x)\), it increases the reported AU relative to the plug-in estimate \(\exp(m_g(x))\). Since the exponential function is strictly increasing,
\[
\rho_{\mathrm S}\!\left(
\widehat{\mathrm{EU}},\widehat{\mathrm{AU}}
\right)
=
\rho_{\mathrm S}\!\left(
v_f,m_g+\frac{v_g}{2}
\right).
\]
Thus, \(v_g\) can increase EU--AU coupling if it is positively associated with \(v_f\) across inputs. Such an association is plausible in our implementation because the two latent GPs are inferred from the same observations and use the same inducing locations. In regions with limited data coverage, both the signal and the noise function may consequently be weakly constrained, producing large values of both \(v_f\) and \(v_g\). The term \(v_g/2\) would then introduce into AU part of the same coverage-dependent pattern represented by the signal posterior variance.

This argument does not establish that \(v_f\) and \(v_g\) must be positively associated. The two processes use different kernels, the posterior mean \(m_g\) may dominate the AU ranking, and their joint optimization may weaken or reverse the proposed association. Moreover, some correlation between AU and \(v_f\) is appropriate because observations with larger noise constrain the signal function less strongly. The proposed explanation concerns only the additional coupling that may arise because the method-side uncertainty \(v_g\) enters the reported AU.

The hypothesis can be evaluated without retraining by replacing the reported AU with the plug-in estimate
\[
\widehat{\mathrm{AU}}_{\mathrm{plug}}(x)=\exp(m_g(x))
\]
and recomputing excess entanglement. Reporting
\(\rho_{\mathrm S}(v_f,v_g)\) would test the required association directly. A reduction in excess entanglement after removing the factor \(\exp(v_g/2)\) would support the proposed explanation. Without this intervention, uncertainty in the noise process remains a plausible but unverified source of VLGP's excess entanglement.

\begin{table}[H]
\centering
\caption{Excess disentanglement per method with the ground-truth EU defined as the pointwise excess risk, $\mathrm{EU}^\star = (\hat a_{\mathcal S}(x) - f^\star(x))^2$. $\Delta\rho = \rho(\mathrm{EU},\mathrm{AU}) - \rho(\mathrm{EU}^\star,\mathrm{AU}^\star)$ (Spearman, over the evaluation points), averaged over all conditions, and separately per oracle engine and per coverage scenario. $\Delta\rho > 0$: EU and AU outputs more strongly coupled than the truth.}
\label{tab:dis:xr}
\small
\setlength{\tabcolsep}{3.5pt}
\begin{tabular}{@{}llcccccccc@{}}
  \toprule
  & & & & & \multicolumn{3}{c}{$\Delta\rho$ by oracle engine} & \multicolumn{2}{c}{$\Delta\rho$ by coverage} \\
  \cmidrule(lr){6-8}\cmidrule(l){9-10}
  & Method & $\rho(\mathrm{EU},\mathrm{AU})$ & $\rho(\mathrm{EU}^\star,\mathrm{AU}^\star)$ & $\Delta\rho$ & GP & HBART & BNN & bal. & OOD \\
  \midrule
  Trees & CatBoost-PS & 0.52 & 0.05 & \textbf{+0.48} & +0.51 & +0.45 & +0.47 & +0.47 & +0.48 \\
   & NGBoost-NIG & 1.00 & 0.15 & \textbf{+0.85} & +0.90 & +0.82 & +0.82 & +0.84 & +0.85 \\
   & NGBoost-Bag & 0.33 & 0.13 & \textbf{+0.20} & +0.23 & +0.16 & +0.22 & +0.19 & +0.22 \\
  \addlinespace
  GPs & VLGP & 0.51 & 0.13 & \textbf{+0.38} & +0.41 & +0.41 & +0.32 & +0.40 & +0.36 \\
  \addlinespace
  NNs & SNGP-Het & 0.46 & 0.14 & \textbf{+0.32} & +0.34 & +0.31 & +0.31 & +0.38 & +0.27 \\
   & NN-Bag & 0.35 & 0.14 & \textbf{+0.21} & +0.23 & +0.25 & +0.15 & +0.21 & +0.21 \\
   & NN-Drop & 0.50 & 0.13 & \textbf{+0.37} & +0.38 & +0.39 & +0.33 & +0.37 & +0.36 \\
   & NN-Ens & 0.36 & 0.14 & \textbf{+0.22} & +0.24 & +0.23 & +0.17 & +0.22 & +0.22 \\
   & NN-Lap & 0.40 & 0.12 & \textbf{+0.28} & +0.29 & +0.29 & +0.27 & +0.33 & +0.23 \\
   & NN-FSP & 0.42 & 0.12 & \textbf{+0.30} & +0.32 & +0.31 & +0.28 & +0.34 & +0.27 \\
   & NN-SWAG & 0.57 & 0.12 & \textbf{+0.44} & +0.45 & +0.52 & +0.37 & +0.40 & +0.48 \\
   & NN-EDL & 0.57 & 0.13 & \textbf{+0.44} & +0.41 & +0.42 & +0.49 & +0.45 & +0.43 \\
  \bottomrule
\end{tabular}
\end{table}

\subsection{OOD Analysis}
\label{app:OOD}
A common way to evaluate epistemic uncertainty without a ground truth is out-of-distribution (OOD)
detection. We use a
natural covariate shift, white vs.\ red wine, to test how such an OOD score relates to our benchmark
ranking, and whether the answer depends on the kind of shift.

\paragraph{Experimental setup.}
We reuse the benchmark's white-wine instances with balanced covariates and smooth-monotone
heteroscedastic noise, for the three oracle engines (GP, HBART, BNN). Every method is fitted exactly as in the benchmark (same
tuned configuration and training seed) and queried on two sets of 1{,}500 points: held-out white wines, and red wines, drawn from the 1{,}599 rows of the red-wine data (OpenML 44972) \citep{openml_red_wine_44972}, which has the same eleven physicochemical features.

\paragraph{Evaluation.}
The OOD score is the method's EU output. Its AUROC for red (positive) vs.\ held-out white (negative)
is computed per instance and averaged over the 15 instances (three oracle engines times 5 seeds), with $95\%$ intervals from a bootstrap
over instances. We evaluate two versions of the task.
\emph{All points} uses both full sets. \emph{Distance-matched} removes the distance shortcut: for each
instance we fit a Mahalanobis distance to the white training inputs, pool the distances of the white and red query
points, cut them into 20 bins at the pooled quantiles, and keep in every bin the same number of white
and red points (the smaller of the two counts, drawn at random). Both classes then have the same
distance distribution; on average 134 points per class remain, the white wines far from the training
centre and the red wines closest to it. A method whose EU only measures distance from the training
data scores $\approx 0.5$ on this version.

Three reference scores bracket the methods: the Mahalanobis distance itself, the mean distance to the
10 nearest training inputs, and a supervised gradient-boosting classifier trained on red vs.\ white
labels (5-fold cross-validated), which bounds how separable the two sets are.

\paragraph{Analysis.}
We compare the OOD ranking of the 19 methods (mean EU-AUROC) with the benchmark ranking (mean
Spearman $\rho$ between EU and the ground-truth target over all regression runs) using Kendall's
coefficient of concordance $W$ (tie-corrected). $95\%$ intervals come from a bootstrap over the wine instances
with the benchmark ranking held fixed. We repeat the comparison for the three definitions of the EU
target (posterior excess risk, posterior variance, pointwise excess risk).

To check whether the distance-matched red wines are genuinely more uncertain, we evaluate the ground
truth directly on the GP-oracle instances, where the posterior is available in closed form.
We compare the ratio of the median EU on matched red and matched white wines between each
method and the true posterior variance.

\paragraph{Results.}
On all points the task is easy: Mahalanobis distance alone reaches AUROC $0.982$ and the supervised
classifier $0.997$ (Table~\ref{tab:wine_ood_ranking}). The best methods (NN-Lap-SN, GP, VLGP, BLR,
SNGP-Het, NN-FSP-SN; $0.976$--$0.982$) match but do not exceed distance, and the ranking agrees closely
with the benchmark: $W = 0.89$ $[0.85, 0.91]$ for the posterior excess risk, $0.92$ $[0.91, 0.93]$ for
the posterior variance and $0.84$ $[0.81, 0.85]$ for the pointwise excess risk
(Table~\ref{tab:wine_ood_kendall_w}). The methods ranked last are those whose EU is coupled to their AU
(NN-EDL $0.48$, NN-DEUP $0.59$, NGBoost-NIG $0.67$).

After distance matching, all methods fall far below the separability bound ($0.954$). Only NN-Ens
($0.666$ $[0.645, 0.687]$) and NN-Bag ($0.639$ $[0.619, 0.655]$) clearly exceed both distance references
(Mahalanobis $0.537$, 10-NN $0.567$); NN-Lap-SN ($0.580$) and CatBoost-KGB ($0.572$) exceed Mahalanobis
but not 10-NN, and VLGP ($0.534$) and SNGP-Het ($0.547$) do not exceed either. The ranking decouples from
the benchmark ($W = 0.64$ $[0.52, 0.75]$, $0.49$ $[0.37, 0.63]$ and $0.53$ $[0.42, 0.65]$ for the three
targets). The ensembles' advantage holds under all three oracles (NN-Ens $0.689$, $0.645$, $0.664$ for
BNN, GP and HBART), whereas the exact GP drops to $0.468$ under the BNN oracle.

The ground truth confirms that the matched red wines are more uncertain although they are not farther
from the training data: under the GP oracle the true posterior variance is $1.36\times$ higher on
matched red than on matched white wines ($1.22$--$1.60$ across instances) and separates the two with
AUROC $0.643$ ($0.986$ on all points). The ensembles reproduce this shift (median EU ratio $1.49$ for
NN-Ens, $1.30$ for NN-Bag), as do the exact GP ($1.27$) and CatBoost-KGB ($1.24$), whereas VLGP
($1.05$), NN-Lap-SN ($1.04$), SNGP-Het ($0.85$) and NN-FSP-SN ($0.71$) barely change or rate red wine as
less uncertain. At the same time, VLGP, the GP, NN-Lap-SN and SNGP-Het track the true posterior variance
beyond distance more closely than the ensembles (partial Spearman correlation given the Mahalanobis
distance $0.90$, $0.84$, $0.77$ and $0.76$ vs.\ $0.37$ for NN-Ens and $0.43$ for NN-Bag): they reproduce
the posterior's variation within the white-wine data, but not its rise at the red wines. The matched
red wines differ from white wine in feature combinations rather than in distance, most strongly in
lower total sulfur dioxide ($-1.8$ training standard deviations) together with higher pH, fixed acidity
and volatile acidity ($+0.8$ to $+0.9$).

Finally, adding the AU output to the OOD score lowers the AUROC for most NN methods (e.g.\ NN-Lap-SN
$0.982 \to 0.735$, NN-FSP $0.950 \to 0.619$): separating EU from AU matters for OOD detection in practice.

\paragraph{Limitations.}
This is a single dataset shift, so it illustrates rather than establishes how OOD detection relates to
EU quality. The distance-matched set is small (134 points per class).

\begin{table}[H]
\centering
\small
\caption{Wine OOD test: methods trained on white wine (benchmark instances; balanced covariates, smooth-monotone noise, 3 oracle engines $\times$ 5 seeds) score held-out white wine against red wine (1500 each) by their epistemic uncertainty. AUROC, mean over the 15 instances, 95\% bootstrap CI over instances. \emph{Distance-matched}: white and red subsampled to equal Mahalanobis-distance distributions w.r.t.\ the training inputs (134 per class), so distance from the training data alone scores $\approx 0.5$.}
\label{tab:wine_ood_ranking}
\begin{tabular}{r l c l c}
\toprule
 & \multicolumn{2}{c}{All points (far OOD)} & \multicolumn{2}{c}{Distance-matched (near OOD)} \\
\cmidrule(lr){2-3}\cmidrule(lr){4-5}
Rank & Method & AUROC & Method & AUROC \\
\midrule
1 & NN-Lap-SN & 0.982 {\scriptsize [0.979, 0.984]} & NN-Ens & 0.666 {\scriptsize [0.645, 0.687]} \\
2 & GP & 0.979 {\scriptsize [0.977, 0.982]} & NN-Bag & 0.639 {\scriptsize [0.619, 0.655]} \\
3 & VLGP & 0.979 {\scriptsize [0.977, 0.981]} & NN-Lap-SN & 0.580 {\scriptsize [0.557, 0.602]} \\
4 & BLR & 0.979 {\scriptsize [0.976, 0.981]} & CatBoost-KGB & 0.572 {\scriptsize [0.558, 0.588]} \\
5 & SNGP-Het & 0.976 {\scriptsize [0.972, 0.980]} & NN-EDL & 0.569 {\scriptsize [0.483, 0.652]} \\
6 & NN-FSP-SN & 0.976 {\scriptsize [0.968, 0.983]} & GP & 0.566 {\scriptsize [0.528, 0.604]} \\
7 & CatBoost-KGB & 0.969 {\scriptsize [0.966, 0.972]} & NGBoost-NIG & 0.558 {\scriptsize [0.520, 0.595]} \\
8 & NN-Ens & 0.968 {\scriptsize [0.965, 0.972]} & SNGP-Het & 0.547 {\scriptsize [0.507, 0.585]} \\
9 & NN-Bag & 0.960 {\scriptsize [0.955, 0.965]} & CatBoost-PS & 0.540 {\scriptsize [0.479, 0.597]} \\
10 & NN-FSP & 0.950 {\scriptsize [0.936, 0.962]} & NN-FSP-SN & 0.539 {\scriptsize [0.469, 0.604]} \\
11 & NN-Drop & 0.937 {\scriptsize [0.919, 0.954]} & VLGP & 0.534 {\scriptsize [0.517, 0.553]} \\
12 & NN-Lap & 0.933 {\scriptsize [0.918, 0.947]} & NN-Drop & 0.528 {\scriptsize [0.471, 0.590]} \\
13 & NN-SWAG & 0.910 {\scriptsize [0.888, 0.932]} & NN-DEUP$^\ast$ & 0.521 {\scriptsize [0.492, 0.562]} \\
14 & DKL & 0.867 {\scriptsize [0.807, 0.923]} & NGBoost-Bag & 0.519 {\scriptsize [0.478, 0.562]} \\
15 & NGBoost-Bag & 0.862 {\scriptsize [0.825, 0.900]} & NN-FSP & 0.497 {\scriptsize [0.462, 0.537]} \\
16 & CatBoost-PS & 0.795 {\scriptsize [0.687, 0.885]} & NN-SWAG & 0.488 {\scriptsize [0.436, 0.546]} \\
17 & NGBoost-NIG & 0.671 {\scriptsize [0.614, 0.733]} & BLR & 0.485 {\scriptsize [0.455, 0.512]} \\
18 & NN-DEUP$^\ast$ & 0.590 {\scriptsize [0.527, 0.669]} & DKL & 0.469 {\scriptsize [0.394, 0.553]} \\
19 & NN-EDL & 0.482 {\scriptsize [0.332, 0.626]} & NN-Lap & 0.424 {\scriptsize [0.376, 0.474]} \\
\midrule
\multicolumn{5}{l}{\emph{Reference scores}} \\
 & Mahalanobis distance & 0.982 & Mahalanobis distance & 0.537 \\
 & 10-NN distance & 0.981 & 10-NN distance & 0.567 \\
 & Supervised classifier$^\dagger$ & 0.997 & Supervised classifier$^\dagger$ & 0.954 \\
\bottomrule
\multicolumn{5}{l}{\scriptsize $^\dagger$ gradient boosting trained on red vs.\ white labels, 5-fold cross-validated; an upper bound on separability, not an uncertainty method.}
\end{tabular}
\end{table}
\begin{table}[H]
\centering
\small
\caption{Agreement between the wine OOD ranking and the benchmark ranking of the same 19 methods, for three definitions of the epistemic target. Kendall's $W$ (tie-corrected); for two rankings $W = (1 + \rho_S)/2$, so $0.5$ is no association and $1$ identical rankings. Benchmark ranking: mean Spearman $\rho$ between EU and the target over all regression runs (balanced and OOD covariates). 95\% CI: bootstrap over the 15 wine instances, benchmark rankings held fixed. The three benchmark rankings agree among themselves with $W = 0.93$, which keeps the $m = 4$ row high even when the OOD ranking is unrelated to them; the pairwise rows are the informative ones.}
\label{tab:wine_ood_kendall_w}
\begin{tabular}{l c c}
\toprule
Benchmark EU target & All points (far OOD) & Distance-matched (near OOD) \\
\midrule
Posterior risk & 0.89 {\scriptsize [0.85, 0.91]} & 0.64 {\scriptsize [0.52, 0.75]} \\
Posterior variance & 0.92 {\scriptsize [0.91, 0.93]} & 0.49 {\scriptsize [0.37, 0.63]} \\
Pointwise excess risk & 0.84 {\scriptsize [0.81, 0.85]} & 0.53 {\scriptsize [0.42, 0.65]} \\
\midrule
All four rankings ($m=4$) & 0.87 {\scriptsize [0.85, 0.88]} & 0.63 {\scriptsize [0.54, 0.71]} \\
\bottomrule
\end{tabular}
\end{table}

\subsection{Spectral Normalization}
\label{app:spectral_normalization}

In the main benchmark SNGP-Het tracks the posterior excess risk much better than DKL
($\rho = 0.65$ vs.\ $0.38$). Both place a Gaussian-process layer on learned features. Beyond
their output layers, SNGP spectrally normalizes its feature extractor so that distances in input
space are approximately preserved in feature space \citep{liu2020sngp}. This appendix
tests whether spectral normalization (SN) improves the EU of the other
NN methods, and whether any improvement is caused by distance preservation rather
than by the additional regularization SN introduces.

\paragraph{Experimental setup.}
We study the eight NN methods of the benchmark: NN-Lap , NN-FSP
, NN-DEUP, NN-EDL, NN-SWAG, NN-Drop, NN-Ens and NN-Bag. Each is fitted in three arms:
\begin{itemize}
  \item \emph{plain}: the benchmark's own tuned configuration, without SN;
  \item \emph{SN}: every dense layer of the residual backbone is spectrally normalized,
        $W \leftarrow W \cdot \min\!\bigl(1, c/\sigma_{\max}(W)\bigr)$, with $\sigma_{\max}$ estimated
        by power iteration (one step per training update, three refinement steps at inference).
        The bound $c$ is searched log-uniformly in $[0.5, 3]$ and all other hyperparameters are
        re-tuned under the constraint;
  \item \emph{WD}: no SN, but weight decay is forced into $[10^{-2}, 1]$ (the benchmark's range is
        $[10^{-6}, 10^{-2}]$) and all other hyperparameters are re-tuned. This control asks whether a
        conventional regularizer of comparable or greater strength reproduces the effect of SN.
\end{itemize}

Balanced covariates, smooth-monotone heteroscedastic noise, all 13 datasets, the three oracle
engines (GP, HBART, BNN).

\paragraph{Evaluation.}
Let $h(x)$ denote the backbone's output, i.e.\ the features the uncertainty head operates on, and
$\tilde x$ the input standardized with the training mean and standard deviation (the space the
network sees). For each instance we draw $20{,}000$ random pairs $(i, j)$ of evaluation points, the
same pairs for all three arms, and compute input and feature distances
\[
  d^{\mathrm{in}}_{ij} = \lVert \tilde x_i - \tilde x_j \rVert_2, \qquad
  d^{\mathrm{feat}}_{ij} = \lVert h(x_i) - h(x_j) \rVert_2 .
\]
\emph{Local distance preservation} is the Spearman correlation between the two, restricted to the
$25\%$ closest input pairs,
\[
  \mathrm{pres}_{\mathrm{loc}}
  = \rho_S\bigl(d^{\mathrm{in}}_{ij}, d^{\mathrm{feat}}_{ij} \,:\, d^{\mathrm{in}}_{ij} \le q_{0.25}(d^{\mathrm{in}})\bigr).
\]
A value of $1$ means that nearby inputs keep their relative distances in feature space; small
values mean the feature map reorders or collapses neighbourhoods. For ensembles and bagging each member's backbone is scored and the
scores are averaged. SWAG is scored at its SWA mean, MC dropout with dropout switched off.

For each instance and method we form the paired differences
$\Delta = \text{arm} - \text{plain}$ for SN and for WD (Table~\ref{tab:sn_study}). Point estimates
are means over the 195 instances. Confidence intervals
are $95\%$ percentile intervals from a non-parametric bootstrap that resamples the 13
\emph{datasets} with replacement (2{,}000 replicates), since datasets, not instances, are the unit
to which the conclusions should generalize.

\paragraph{Analysis.}
To ask whether SN helps through distance preservation, we regress, per method, the change in
EU quality on the change in preservation, the change in accuracy, the arm, and the change in depth,
pooling the SN and WD contrasts:
\[
  \Delta\rho \;=\; \beta_{\mathrm{pres}}\, \Delta\mathrm{pres}_{\mathrm{loc}}
            + \beta_{\mathrm{fit}}\, \Delta\log\mathrm{NMSE}
            + \beta_{\mathrm{SN}}\, \mathbb{1}[\text{SN}]
            + \beta_{\mathrm{depth}}\, \Delta\mathrm{depth}
            + \alpha_{\mathrm{dataset}} + \varepsilon ,
\]
with dataset fixed effects $\alpha_{\mathrm{dataset}}$, all variables standardized (so coefficients
are in standard deviations), contrasts winsorized at the 1st and 99th percentile within each method,
and $95\%$ intervals from a dataset bootstrap. If SN acts through preservation,
$\beta_{\mathrm{pres}}$ is positive and the SN indicator loses its effect once $\Delta\mathrm{pres}$ is
in the model. If it acts through regularization or fit, $\beta_{\mathrm{fit}}$ carries it instead. A
pooled model over all methods adds method fixed effects.

Because SN changes preservation much more than WD, the arm indicator and $\Delta\mathrm{pres}$ are
correlated in the model above. We therefore also test, within the SN arm alone, whether instances in
which SN raised preservation more also gained more:
$\Delta\rho_{\mathrm{var}} \sim \Delta\mathrm{pres}_{\mathrm{loc}} + \Delta\log\mathrm{NMSE} + \alpha_{\mathrm{dataset}}$,
standardized, with dataset-bootstrap intervals.

For Table~\ref{tab:sn_hparams} we correlate the plain network's tuned depth, width, epochs and
number of gradient steps with its preservation, and its depth with the SN gains, using Spearman's
correlation \emph{within} datasets (dataset means removed from both variables).
Depth is chosen by the tuner rather than set by us, so these are
associations within datasets, not interventions.

\paragraph{Results.}
SN raises local preservation in all eight methods ($+0.15$ to $+0.35$; pooled $+0.21$
$[0.17, 0.25]$), whereas WD leaves it unchanged (pooled $+0.01$ $[-0.01, 0.02]$).

For the four methods whose EU is read from a single network's features (NN-Lap, NN-FSP, NN-DEUP,
NN-EDL), SN improves $\rho_{\mathrm{risk}}$ by $+0.10$ to $+0.11$ and $\rho_{\mathrm{var}}$ by $+0.08$
to $+0.26$. WD yields at most about half of the gain, and only for NN-Lap and NN-FSP
(Table~\ref{tab:sn_study}). For the ensembles and MC dropout, whose EU arises from disagreement
between models or masks, preservation rises without improving EU (NN-Ens $+0.00$, NN-Drop $-0.01$,
NN-Bag $-0.03$ in $\rho_{\mathrm{risk}}$). NN-SWAG deteriorates ($-0.08$ in $\rho_{\mathrm{risk}}$,
$-0.135$ in $\rho_{\mathrm{var}}$), see below.

For posterior-variance tracking, preservation carries the effect for NN-Lap
($\beta_{\mathrm{pres}} = +0.40$ $[0.20, 0.52]$, $\beta_{\mathrm{SN}} = -0.00$ $[-0.18, 0.20]$) and NN-FSP
($+0.51$ $[0.22, 0.78]$, $\beta_{\mathrm{SN}} = -0.08$ $[-0.27, 0.14]$), with a smaller contribution from
accuracy ($\beta_{\mathrm{fit}} = -0.30$ and $-0.26$). Pooled over all methods, the SN advantage in
$\rho_{\mathrm{var}}$ ($+0.11$ $[0.08, 0.15]$ without mediators) vanishes once the mediators are included
($\beta_{\mathrm{SN}} = +0.01$ $[-0.09, 0.09]$) and is carried by preservation
($\beta_{\mathrm{pres}} = +0.15$ $[0.05, 0.25]$), not by accuracy ($\beta_{\mathrm{fit}} = -0.08$
$[-0.17, 0.02]$).

Within the SN arm, a one-standard-deviation larger rise in preservation goes with a $0.44$
$[0.03, 0.69]$ (NN-Lap) and $0.64$ $[0.38, 0.89]$ (NN-FSP) standard-deviation larger gain in
$\rho_{\mathrm{var}}$ at equal change in accuracy. For NN-DEUP and NN-EDL there is no such relation
($+0.09$ and $-0.10$, both not significant): their gains are not explained by preservation.

Within datasets, deeper plain networks preserve distances worse in all eight methods
(Table~\ref{tab:sn_hparams}; pooled $\rho_S = -0.57$), whereas width, epochs and gradient steps show
no consistent relation. SN restores more preservation in deeper networks (pooled $+0.41$; positive
for seven of eight methods) and the gain in $\rho_{\mathrm{var}}$ grows with depth for NN-FSP
($+0.53$), NN-Lap ($+0.38$) and NN-DEUP ($+0.28$). The full benchmark (312 keys) agrees: for NN-FSP the
within-dataset correlation of the SN gain with the plain depth is $+0.36$, and controlling for the
plain network's preservation removes the depth effect ($+0.035$ to $+0.005$ per standard deviation),
consistent with depth acting through preservation. When tuned with SN, networks also select deeper
backbones (e.g.\ $2.6$ vs.\ $1.4$ residual blocks on average for NN-FSP).

\begin{figure}[!htbp]
    \centering
    \includegraphics[width=\linewidth]{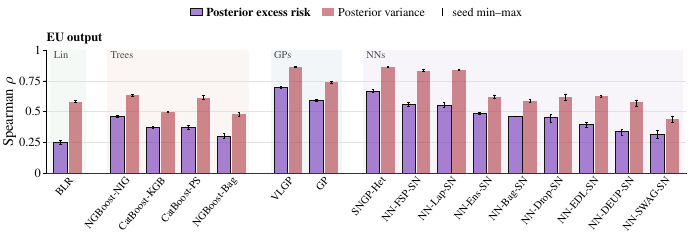}
    \caption{Benchmark results including SN variants on the spectral-normalization study subset: 13 datasets, three oracle engines, and five seeds, with balanced covariates and smooth-monotone heteroscedastic noise.}
    \label{fig:sn_subset_benchmark}
\end{figure}

\paragraph{Benchmark comparison.}
Figure~\ref{fig:sn_subset_benchmark} compares the SN variants with the benchmark methods on the same evaluation subset. Several SN variants approach SNGP-Het (NN-Laplace and FSP-Laplace) in EU quality, substantially narrowing the performance gap observed for their plain counterparts.

\begin{table}[H]
\centering
\small
\setlength{\tabcolsep}{4pt}
\caption{Spectral normalisation (SN) vs.\ strong weight decay (WD), each tuned separately (30-trial Bayes sweep on validation NLL; WD searched in $[10^{-2}, 1]$, 100--1000$\times$ the usual range), as paired changes against the plain network. 13 datasets $\times$ 3 oracles $\times$ 5 seeds (195 instances per method; balanced covariates, smooth-monotone heteroscedastic noise). $\Delta\rho$: change in Spearman correlation of the EU output with the posterior risk and with the posterior variance; $\Delta$pres.: change in local distance preservation of the backbone features; NMSE: ratio of the arm's to the plain network's error (geometric mean). Bold: 95\% CI from bootstrapping datasets excludes 0 (NMSE: 1). }
\label{tab:sn_study}
\begin{tabular}{l cccc cccc}
\toprule
 & \multicolumn{4}{c}{SN $-$ plain} & \multicolumn{4}{c}{WD $-$ plain} \\
\cmidrule(lr){2-5}\cmidrule(lr){6-9}
Method & $\Delta\rho_{\text{risk}}$ & $\Delta\rho_{\text{var}}$ & $\Delta$pres. & NMSE & $\Delta\rho_{\text{risk}}$ & $\Delta\rho_{\text{var}}$ & $\Delta$pres. & NMSE \\
\midrule
NN-Lap & \textbf{+0.103} & \textbf{+0.077} & \textbf{+0.172} & \textbf{0.48} & \textbf{+0.049} & \textbf{+0.033} & \textbf{+0.021} & \textbf{0.70} \\
NN-FSP & \textbf{+0.108} & \textbf{+0.092} & \textbf{+0.155} & \textbf{0.38} & \textbf{+0.046} & \textbf{+0.041} & $-$0.013 & \textbf{0.64} \\
NN-DEUP & \textbf{+0.108} & \textbf{+0.258} & \textbf{+0.345} & \textbf{0.44} & +0.020 & +0.023 & +0.012 & 0.91 \\
NN-EDL & \textbf{+0.109} & \textbf{+0.154} & \textbf{+0.184} & \textbf{0.48} & +0.023 & +0.035 & $-$0.024 & \textbf{0.78} \\
\addlinespace
NN-SWAG & \textbf{$-$0.080} & \textbf{$-$0.135} & \textbf{+0.237} & \textbf{0.62} & $-$0.016 & $-$0.012 & \textbf{+0.041} & \textbf{0.88} \\
NN-Drop & $-$0.006 & $-$0.002 & \textbf{+0.244} & \textbf{0.83} & \textbf{$-$0.020} & \textbf{$-$0.023} & \textbf{+0.045} & \textbf{0.93} \\
\addlinespace
NN-Ens & +0.002 & $-$0.021 & \textbf{+0.179} & \textbf{0.78} & $-$0.001 & \textbf{$-$0.025} & +0.025 & 0.93 \\
NN-Bag & \textbf{$-$0.026} & \textbf{$-$0.065} & \textbf{+0.150} & \textbf{0.75} & \textbf{$-$0.035} & \textbf{$-$0.060} & $-$0.029 & \textbf{0.89} \\
\midrule
All (pooled) & \textbf{+0.040} & \textbf{+0.045} & \textbf{+0.208} & \textbf{0.57} & +0.008 & +0.001 & +0.010 & \textbf{0.83} \\
\bottomrule
\end{tabular}
\end{table}
\begin{table}[H]
\centering
\small
\setlength{\tabcolsep}{4pt}
\caption{Deeper networks preserve distances worse and gain more from SN. Left: Spearman correlation of the plain
network's distance preservation with its tuned training hyperparameters. Right: Spearman correlation of the plain
network's depth with the gain from SN in distance preservation and in $\rho(\mathrm{EU}, \text{posterior variance})$.
Computed across the 195 instances per method (13 datasets $\times$ 3 oracles $\times$ 5 seeds) within datasets
(dataset means removed); pooled: additionally centred per method. Distance preservation: Spearman correlation of
pairwise input and feature distances over the $25\%$ closest input pairs. Bold: $p < 0.05$.}
\label{tab:sn_hparams}
\begin{tabular}{l cccc cc}
\toprule
 & \multicolumn{4}{c}{Plain preservation vs.} & \multicolumn{2}{c}{Depth vs.\ SN gain in} \\
\cmidrule(lr){2-5}\cmidrule(lr){6-7}
Method & Depth & Width & Epochs & Grad.\ steps & Preservation & $\rho_{\text{var}}$ \\
\midrule
NN-Lap  & \textbf{$-$0.70} & \textbf{+0.17} & +0.13 & $-$0.01 & \textbf{+0.63} & \textbf{+0.38} \\
NN-FSP  & \textbf{$-$0.70} & \textbf{+0.51} & +0.09 & +0.10 & \textbf{+0.50} & \textbf{+0.53} \\
NN-DEUP & \textbf{$-$0.59} & \textbf{$-$0.62} & \textbf{$-$0.34} & \textbf{$-$0.41} & \textbf{+0.39} & \textbf{+0.28} \\
NN-EDL  & \textbf{$-$0.44} & \textbf{+0.33} & \textbf{$-$0.37} & \textbf{$-$0.32} & \textbf{+0.37} & +0.07 \\
\addlinespace
NN-SWAG & \textbf{$-$0.36} & +0.08 & \textbf{$-$0.18} & \textbf{$-$0.25} & $-$0.07 & $-$0.02 \\
NN-Drop & \textbf{$-$0.33} & +0.04 & +0.01 & \textbf{$-$0.14} & \textbf{+0.27} & +0.00 \\
\addlinespace
NN-Ens  & \textbf{$-$0.87} & +0.06 & \textbf{$-$0.27} & $-$0.08 & \textbf{+0.55} & \textbf{$-$0.19} \\
NN-Bag  & \textbf{$-$0.76} & \textbf{+0.43} & \textbf{+0.27} & \textbf{+0.36} & \textbf{+0.49} & +0.01 \\
\midrule
Pooled (8 methods) & \textbf{$-$0.57} & \textbf{+0.08} & \textbf{$-$0.12} & \textbf{$-$0.13} & \textbf{+0.41} & \textbf{+0.09} \\
\bottomrule
\end{tabular}
\end{table}

\newpage

\stopcontents[appendix]
\end{document}